%% file: submission.tex
\documentclass[10pt,twocolumn,letterpaper]{article}

\usepackage[pagenumbers]{cvpr} 

\usepackage{graphicx}
\usepackage{amsmath}
\usepackage{amssymb}
\usepackage{booktabs}
\usepackage{makecell}
\usepackage{listings}
\usepackage{xcolor}
\usepackage{tabularx}

\makeatletter
\@namedef{ver@everyshi.sty}{}
\makeatother
\usepackage{pgfplots}

\DeclareMathOperator*{\argmin}{\arg\min}
\usepackage{color, colortbl}
\definecolor{LightCyan}{rgb}{0.9, 0.9, 0.98}

\usepackage{comment}

\newcommand{\mynorm}[1]{ \left\| #1 \right\| }

\usepackage[pagebackref,breaklinks,colorlinks]{hyperref}

\usepackage[capitalize]{cleveref}
\crefname{section}{Sec.}{Secs.}
\Crefname{section}{Section}{Sections}
\Crefname{table}{Table}{Tables}
\crefname{table}{Tab.}{Tabs.}

\def\cvprPaperID{3848} 
\def\confName{CVPR}
\def\confYear{2023}

\begin{document}

\title{DARE-GRAM :  Unsupervised Domain Adaptation Regression by Aligning Inverse Gram Matrices}

\author{Ismail Nejjar\\
EPFL, Switzerland\\
{\tt\small ismail.nejjar@epfl.ch}
\and
Qin Wang\\
ETH Zurich, Switzerland\\
{\tt\small qwang@ethz.ch}
\and
Olga Fink\\
EPFL, Switzerland\\
{\tt\small olga.fink@epfl.ch}
}
\maketitle

\begin{abstract}
Unsupervised Domain Adaptation Regression (DAR) aims to bridge the domain gap between a labeled source dataset and an unlabelled target dataset for regression problems. Recent works mostly focus on learning a deep feature encoder by minimizing the discrepancy between source and target features. 
In this work, we present a different perspective for the DAR problem by analyzing the closed-form ordinary least square~(OLS) solution to the linear regressor in the deep domain adaptation context.  Rather than aligning the original feature embedding space, we propose to align the inverse Gram matrix of the features, which is motivated by its presence in the OLS solution and the Gram matrix's ability to capture the feature correlations. Specifically, we propose a simple yet effective DAR method which leverages the pseudo-inverse low-rank property to align the scale and angle in a selected subspace generated by the pseudo-inverse Gram matrix of the two domains. We evaluate our method on three domain adaptation regression benchmarks. Experimental results demonstrate that our method achieves state-of-the-art performance. Our code is available at \url{https://github.com/ismailnejjar/DARE-GRAM}.

\end{abstract}

\section{Introduction}
\label{sec:intro}

Regression problems, in which models learn to predict continuous variables, are one fundamental paradigm in machine learning. 
Regression problems are omnipresent in many different applications, including computer vision tasks, such as head-pose estimation \cite{Yang_2019_CVPR}, facial landmark detection \cite{li2022towards}, human pose estimation \cite{zheng20213d}, depth estimation~\cite{godard2019digging} and eye-tracking problems\cite{shenoy2021r}, and also widely in industrial applications, such as product quality prediction and condition monitoring \cite{8756463}. Nevertheless, real-world applications are often subject to the environmental conditions under which the data are collected and other influencing factors, hence domain gaps between datasets are inevitable. 

\begin{figure}
    \centering
    \includegraphics[scale=0.69]{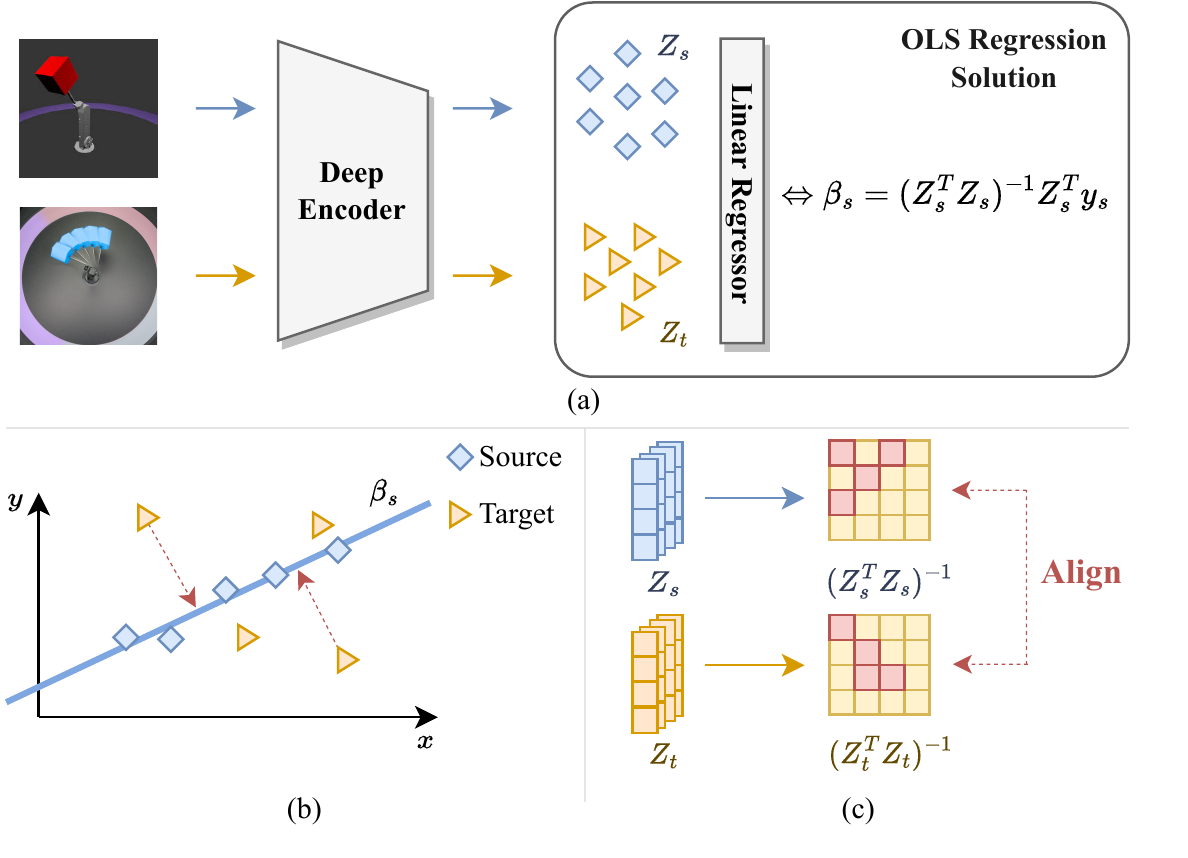}
    \caption{Illustration of the UDA for regression setup and our main motivation. (a) Deep domain adaptation networks commonly use a shared deep feature encoder and a shared linear regression layer. We propose to pay close attention to the linear regressor, where the ordinary least square~(OLS) solution is well-known.   (b) Given a trained source linear regressor $\beta_s$, the target features $Z_t$ may not be calibrated to $\beta_s$.  (c)  Unlike previous adaptation methods which align in the original feature embedding space, we propose to align the inverse Gram matrix of the features $(Z^TZ)^{-1}$, which is motivated by its presence in the OLS solution and the Gram matrix's ability to capture the feature correlations. } 
    \vspace{-0.7cm}
    \label{fig:edge}
\end{figure}

Unsupervised Domain adaptation (UDA) aims to overcome the distributional shift between a labeled source domain and an unlabelled target domain. Many UDA methods have been proposed to alleviate the domain shift problem. One common UDA direction is feature alignment by adversarial learning \cite{long2018conditional} or explicit losses such as maximum mean discrepancy \cite{long2015learning} to learn domain-invariant representations. Input alignment \cite{yang2020fda} and self-training using pseudo-label refinement\cite{lian2019constructing} are also popular UDA directions. While many DA methods have been developed and evaluated for classification and segmentation problems, some are not directly transferable to DA regression~\cite{chen2021representation}.
Pioneer works in Domain Adaptation Regression~(DAR)~\cite{cortes2011domain,mansour2009domain} introduced theoretical analysis for the problem. A few algorithms were proposed to tackle DAR.  For example, importance weighting~\cite{demathelin2021adversarial,yamada2014domain} and feature alignment~\cite{cao2010adaptive, pan2010domain} have shown improved results over learning only from the source. 
Most recent unsupervised DAR methods~\cite{singh2020deep,chen2021representation} use the deep learning framework and focus on learning a shared deep feature extractor by directly minimizing the discrepancy between source and target features. By doing so, it is implicitly assumed that if the feature discrepancy is small, a shared linear regressor can be easily learned from the source supervision. This formulation used by existing works focuses solely on the feature extractor. 

In this work, we propose to look at the DAR problem from a different perspective. In particular, we pay close attention to the linear regressor, which is attached directly after the feature extractor. Motivated by the closed-form ordinary least squares~(OLS) regression solution, we analyze the potential optimal regressor for each domain. We reveal in Section~\ref{sec:Motivation} that even when the discrepancy between source and target features is small, the learning of a shared linear regressor could still be difficult because of the \textit{inverse Gram matrix} term in the OLS solution. 

In light of this, we propose an ordinary least squares inspired deep domain adaptation method for regression called Domain Adaptation Regression by aligning the inverse GRAM matrices (DARE-GRAM). As shown in Figure \ref{fig:edge}, unlike previous methods, which directly align the features, we align the inverse Gram matrix of the features. This is motivated by its presence in the closed-form solution of the ordinary least squares. More specifically, we leverage the low-rank property of the pseudo-inverse to align a selected subspace in scale and angle engendered by the Gram Matrix, which 
represents the intensity and pairwise interactions between different features for the source and target domains. The scale and angle alignment based on the Gram matrix can lead to a better-calibrated regressor with regard to both source and target data. The contributions of this work are as follows:

\begin{itemize}
    \item We offer a new perspective to understand the UDA for regression problems by leveraging the well-known closed-form solutions to the linear regression problem.
    \item Rather than aligning the original feature embedding space, we propose to align the inverse Gram matrix of the features.
    \item Empirical results on three benchmarks validate the superiority of the DARE-GRAM over baseline methods.
\end{itemize}

\section{Related Work}
\label{sec:Related Work}

\noindent\textbf{Unsupervised Domain Adaptation.} The goal of unsupervised domain adaptation (UDA) \cite{patel2015visual} is 
to address the domain-shift problem between a labeled source and an unlabeled target domain.
UDA has been widely studied for classification and segmentation problems \cite{tsai2018learning,vu2019advent} to mitigate the gap between features across different domains. Early works addressed this problem via instance weighting \cite{huang2006correcting,sugiyama2007direct}, feature transformation \cite{pan2010domain}, and feature space alignment \cite{fernando2013unsupervised}. More recently, unsupervised domain adaptation has shown impressive results \cite{10.1145/3400066,na2022contrastive,hoyer2022hrda}. Discrepancy minimization \cite{long2015learning,kang2019contrastive} and domain adversarial learning \cite{ganin2015unsupervised,hoffman2018cycada} have been widely used within UDA methods to mitigate the gap between features across different domains. Moreover, feature regularization-based approaches \cite{chen2019transferability} and domain-specific normalization-based methods \cite{chen2019domain,li2016revisiting} have also demonstrated good performance. While most approaches perform feature alignment in the encoding feature space, some works proposed to carry out alignment in the input space \cite{yang2020fda}.  More recently, self-training has also demonstrated encouraging results by training the network with gradually improved target pseudo-label \cite{Liang_2022_CVPR, Wang_2022_CVPR,zhang2021prototypical,zhang2021efficient}. Existing UDA methods mostly focus on classification and segmentation problems. While some UDA techniques can directly be applied to regression problems, recent works have shown that many do not perform  well in the regression setup~\cite{chen2021representation,Bao_2022_CVPR}. 

\noindent\textbf{Domain Adaptation for Regression.} Domain Adaptation for Regression (DAR) has received relatively little attention in comparison to classification problems. Early theoretical properties for DAR were introduced in \cite{cortes2011domain,mansour2009domain}. Different algorithms were proposed to tackle DAR \cite{redko2020survey}. Unfortunately, most algorithms require access to a labeled target domain and are unsuitable for UDA regression. For instance, Boosting strategies have been explored  \cite{pardoe2010boosting,wang2019transfer} to extend previous classification domain adaptation methods based on AdaBoost~\cite{margineantu1997pruning} to regression tasks. Other instance weighting methods in the shallow regime, \cite{yamada2014domain,yamada2012no,demathelin2021adversarial} have been explored for a different range of applications. 
Some specific vision applications have been explored in the context of UDA \cite{jiang2021regressive,li2021synthetic,ohkawa2022domain,kim2022unified} , such as monocular depth estimation 
 \cite{tonioni2019unsupervised,lo2022learning, Akada_2022_WACV,Bhattacharjee_2022_WACV} or gaze estimation \cite{bao2022generalizing,guo2020domain}. However, these methods aim at improving upon a specific task and not for regression tasks in general. Recent works for UDA regression were proposed  \cite{chen2021representation,singh2020deep,wu2022distribution}. A key finding in RSD \cite{chen2021representation} is that in regression problems, deep neural networks are less robust to feature scaling than classification, and aligning the distributions of deep representations will alter feature scale and impede domain adaptation regression. To tackle this challenge, the authors of \cite{chen2021representation} proposed to match the orthogonal bases of both domains to close domain shifts without altering their feature scale by introducing a new geometrical distance. While RSD-based methods have shown improved results for DAR, matching only the eigenvectors can have some disadvantages, such as more loose numerical error bound \cite{anderson1999lapack} and may not satisfy the more strict conditions for distribution estimation \cite{knowles2013eigenvector}. In contrast to RSD, we propose using the inverse Gram Matrix, which carries the necessary information to align the source and target features while being less sensitive to the batch size.

\noindent\textbf{Gram Matrix and Subspace Alignment.} Distribution alignment approaches have been used for domain adaptation \cite{Wei_2021_CVPR,sun2016deep,fernando2013unsupervised}. Subspace-based domain adaptation has demonstrated good performance in visual domain adaptation \cite{gong2012geodesic,gopalan2011domain}, modeling distribution change by finding the best intermediate subspaces. The methods first independently compute a domain-specific d-dimensional subspace for the source and target data. Then project the source and target data into intermediate ones along the shortest geodesic path connecting the two d-dimensional subspaces on the Grassmann manifold.
Instead of computing a large number of intermediate subspaces, the authors of \cite{fernando2013unsupervised} directly aligns the two subspaces. Furthermore, the authors in \cite{sun2015subspace} proposed to incorporate distribution alignment into subspace adaptation to align the source and target features.
Given their close relation, distribution alignment approaches have also been used for Neural Style Transfer (NST)  \cite{bousmalis2017unsupervised,kalischek2021light}.  Early works on NST \cite{gatys2016image} introduced the Gram Matrix as the statistics of feature maps to extract style-specific attributes. Although the connection between aligning distributions and NST may not be straightforward, it was demonstrated in \cite{li2017demystifying} that the style loss in \cite{gatys2016image} may be expressed as an unbiased empirical estimate of the Maximum Mean Discrepancy (MMD)\cite{gretton2012kernel} with a quadratic kernel.  Unlike previous works in neural style transfer which directly aligns the Gram matrix, we propose to align the inverse Gram matrix as it is presented in the OLS solution. We will show in our method and our ablation study that this is critical for regression problems.

\begin{figure*}
\centering
\includegraphics[width=\textwidth]{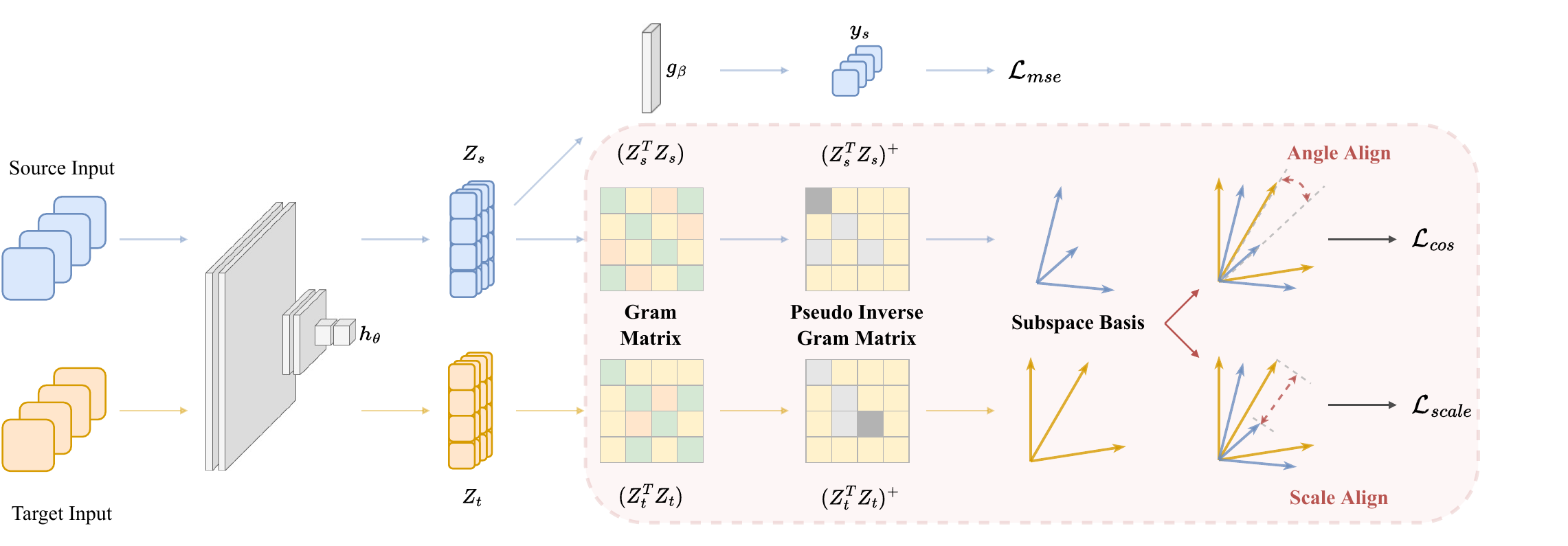}
\caption{An overview of the proposed DARE-GRAM approach for domain adaptive regression problems. Rather than aligning the features $Z$, we align the inverse Gram matrix, which is motivated by the ordinary least square solution. To achieve this, we compute the pseudo-inverse Gram Matrices for source and target features and align their angle and scale. }
\label{fig:tab}
\end{figure*}

\section{Methods}
\label{sec:Methods}


\subsection{Problem Definition}
\label{sec:Problem formulation}
In UDA, we are given labeled samples $\raisebox{2pt}{$\chi$}_s = \{(x_s^i,y_s^i)\}_{i=1}^{N_s}$ from the source domain and unlabeled samples $\raisebox{2pt}{$\chi$}_t = \{(x_t^i)\}_{i=1}^{N_t}$ from the target domain, where $N_s$ and $N_t$ denote the number of samples in $\raisebox{2pt}{$\chi$}_s$ and $\raisebox{2pt}{$\chi$}_t$. In contrast to the discrete labels $\mathcal{Y}$ in classification problems, this work focuses on the regression problem where $\mathcal{Y} \subset \mathbb{R}^{N_r}$ is multidimensional and continuous, and $N_r$ correspond to the number of regression tasks. The discrepancy between $P(\raisebox{2pt}{$\chi$}_s)$ and $P(\raisebox{2pt}{$\chi$}_t)$ is one of the main challenges for UDA. We aim to learn a model $F: x \mapsto y$, which can generalize well on the target domain. Formally, we want to minimize the expected error on the target data:
\begin{equation}
    \argmin_{F} \mathbb{E}_{(x^t,y^t)}\lVert F(x^t),y^t\rVert_2^2,
\end{equation} 
where $y^t$ is not known during the training. 

A source-only baseline can be learned by using the supervision from the source data by minimizing the Mean square error loss (MSE) between the prediction and the ground truth label on the source samples:
\begin{equation}
    \mathcal{L}_{src} = \frac{1}{N_s} \sum_{i=1}^{N_s} \lVert \Tilde{y}_s^i-y_s^i \rVert_2^2,
\end{equation}

\noindent where $\Tilde{y}_s^n=F(x_s^i)$ is the predicted value for the training source image $x_s^i$.  To overcome the distribution gap between the source and target, additional constraints should be given.

\subsection{Motivation}
\label{sec:Motivation}
In deep domain adaptation models, given an input image $x$, a feature encoder $h_\theta$ is used to learn the deep representation $z=h_\theta(x)$ of $p$ dimensions. A linear layer $g_\beta$ is then applied on ${z}$ to make the final prediction:
\begin{equation}
    \tilde{y} = F(x)  = g_\beta(h_\theta(x)) = g_\beta(z) .
\label{eq:zab}
\end{equation}
During training, the feature matrix is $Z = [z^1,...,z^b]$ where $Z \in \mathbb{R}^{b\times p}$ for a batch of $b$ images.  For many adaptation methods\cite{chen2021representation,singh2020deep}, the focus has been on minimizing the distribution difference between source features $Z_s$ and target features $Z_t$. Given the aligned features, it is often assumed that they will then lead to a good performance on the target domain.  However, this formulation focuses solely on the feature extractor $h_\theta$ and does not take the discrimination ability of the final linear layer $g_{\beta}$ into account. Target features aligned with the source domain may not be adapted to the linear layer. This can be especially dangerous for regression problems because it has been demonstrated empirically~\cite{chen2021representation} that in the DA for regression context, the models can be sensitive to feature scale differences.

In this work, we propose to take the linear prediction layer $g_\beta$ into account for the distribution alignment in domain adaptation regression problems. The proposed research is motivated by the question \textit{How to find a feature space, on which a shared linear regressor can easily learn}?

Fortunately, for the linear regression problem, a closed-form solution exists and is well-studied. Given the feature $Z$ and regression ground truth label $Y$, the problem of estimating the parameter ${{\beta}}$ for a linear layer $Y = Z{\beta}$ has 
the ordinary least-squared~(OLS) closed-form solution~\cite{goldberger1964econometric}:
\begin{equation}
    {\hat{\beta}} = (Z^TZ)^{-1}Z^TY
\label{eq:ols_opti}
\end{equation}

\noindent where $(Z^TZ)^{-1} \in \mathbb{R}^{p\times p}$ is the inverse of the Gram Matrix. Entries are then the inner products of the basis functions of the finite-dimensional subspace. $Z^TY \in \mathbb{R}^{p\times N_r}$ projects features to the label space.

The final linear prediction layer $g_\beta$ is shared by the source and target domains. Therefore, the estimated value from the two domains should be similar ${\hat{\beta_s}} \sim {\hat{\beta_t}}$ where
\begin{equation}
\begin{split}
    {\hat{\beta_s}} &= (Z_s^TZ_s)^{-1}\  Z_s^TY_s,  \\
    {\hat{\beta_t}} &= (Z_t^TZ_t)^{-1}\ Z_t^TY_t .
\end{split}\label{eq:zst}
\end{equation}

\begin{figure}
\centering
\begin{subfigure}[t]{0.35\linewidth}
    \includegraphics[width=1.\textwidth]{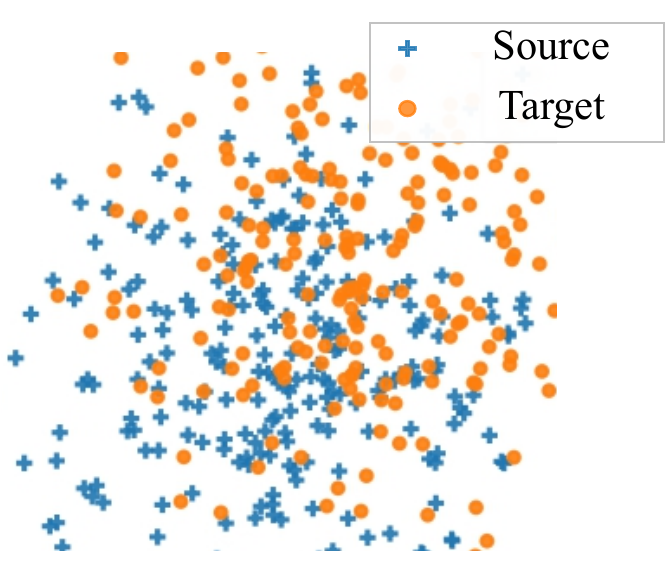}
    \caption{Two distributions}
        \label{fig:toy_distribution}
\end{subfigure}
\begin{subfigure}[t]{0.3\linewidth}
\centering
\includegraphics[width=0.8\textwidth]{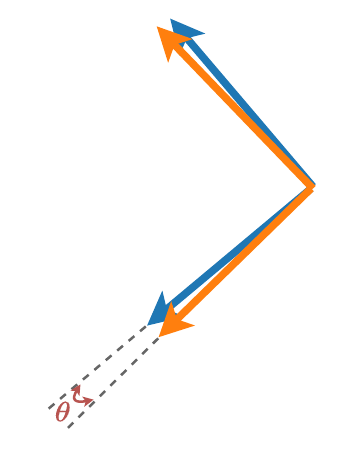}
\caption{$Z$}
\label{fig:toy_rsd_bsp}
\end{subfigure} 
\begin{subfigure}[t]{0.3\linewidth}
\centering
\includegraphics[width=1\textwidth]{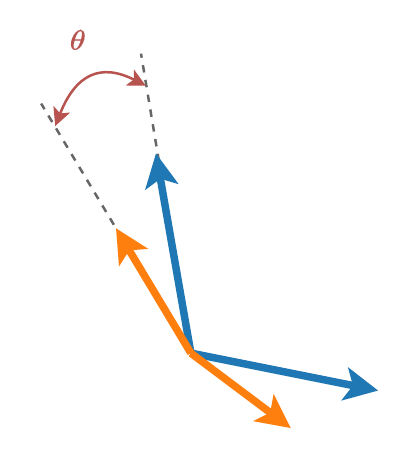}
   \caption{$(Z^TZ)^{-1}$}
    \label{fig:toy_DARE-GRAM}
\end{subfigure}

\caption{Illustration of the impact the inverse Gram operation in the OLS solution. (a) Assume that the features of the source and target follow two Gaussian distributions with slightly different mean and variance. (b) Under the representation subspace distance, the feature subspaces of $Z_s$ and $Z_t$ are well aligned with a very small angle difference. (c) However, because of the inverse Gram operation, the difference in terms of the inverse Gram matrix $(Z^TZ)^{-1}$ can still be large in terms of both angle and scale. }
\label{fig:Toy}
\vspace{-0.3cm}
\end{figure}

Most of the previous DAR works regularize the neural network by minimizing the distance between source and target data in the feature representation subspace of $Z$, i.e. aligning $Z_s$ and $Z_t$. However, because of the inverse operation in Equation~\ref{eq:zst}, even if the distance between $Z_s$ and $Z_t$ is small, the distance in terms of $(Z^TZ)^{-1}$ can be large, as seen in Figure \ref{fig:Toy}. This can further lead to distinct $\hat{\beta_s}$ and $\hat{\beta_t}$ and makes it potentially infeasible to learn a common regressor that performs well for both domains. Given this observation and motivated by the closed-form OLS solution, we propose to focus on the subspace of inverse Gram matrix $(Z^TZ)^{-1}$ for the alignment. 
More specifically, we propose to align the angle between the source and target pseudo-inverse Gram Matrix, formed by a subset of the eigenspace. In addition, we propose to ensure the same scale of $Z$ for the source and target reflected by the Gram matrix $(Z^TZ)$ by minimizing the distance between selected eigenvalues of both domains.

\subsection{Angle Alignment for Gram Matrix Inverse}


The first term in Equation~\ref{eq:zst} concerns the Gram matrix. The Gram matrix is sometimes regarded as a style representation as it calculates the correlations between the different features. It can also be seen as an unbiased empirical estimate of the MMD with a quadratic kernel \cite{gatys2016image}. The inverse operation is also essential because it first relates the variance of the unbiased estimator $\beta$ to the eigenvalues of $(Z^TZ)$. Particular attention must be paid to the small eigenvalues, which have a maximum inflationary effect on the variance of the least squares estimator by significantly destabilizing the estimator
when it approaches zero. Secondly, the ill-conditioned Gram matrix motivates using a low-rank inverse approximation \cite{chung2015optimal}, which allows obtaining regularised basis to be aligned for the source and target domain.

However, such an alignment is non-trivial because the Gram matrix can be non-invertible in deep learning models. During training, the batch size $b$ is generally smaller than the embedding dimension $p$. Given a feature matrix $Z\in \mathbb{R}^{b\times p}$, with $b<p$, the Gram Matrix $(Z^TZ)\in\mathbb{R}^{p\times p}$, has rank $r$ smaller or equal to $b$. Hence the Gram Matrix is not fully ranked and thus not invertible. The Moore-Penrose pseudo-inverse in this case can generalize the concept of matrix inverse when the matrix may not be invertible.


We propose to consider only a selected subspace of the Gram matrix to solve this problem. As not all basis vectors contribute equally, the basis vectors with the highest eigenvalues are the most influential. Therefore, we only consider the most dominant basis vectors in the alignment process.
This step has two main objectives : (i) maximize the mutual information between the two distributions by considering only a selected subset, (ii) avoid numerical instability when not considering degenerate eigenspace.

Concretely, given the singular value decomposition (SVD)~\cite{van1996matrix} of the feature matrix $Z$ defined by $Z = UDV^T$. The Gram matrix $(Z^TZ)$, can be decomposed using the SVD of $Z$ as :
\vspace{-0.1cm}
\begin{equation}
\begin{split}
    (Z^TZ) = (UDV^T)^T(UDV^T) = V \Lambda V^T,\\
    \lambda_k := \Lambda_{k,k} = D_{k,k}^2 \quad \text{for} \quad k = 1,...,p.
    \end{split}
\end{equation}

\noindent where the orthogonal matrix $V \in \mathbb{R}^{p\times p}$ is identical to the matrix in the SVD of $Z$ and $\Lambda \in \mathbb{R}^{p\times p}$ is the diagonal matrix containing the squared eigenvalues of $Z$.

Given the ordered eigenvalues of the Matrix $(Z^TZ)$ $\lambda_1 \geq  ...\geq \lambda_k \geq  ... \geq \lambda_p \geq 0$, the Moore-Penrose pseudo-inverse~\cite{pseudo_inverse} can be derived by discarding the singular values that are below $\lambda_k$ and treating them as zero. The pseudo-inverse of $(Z^TZ)$ can be expressed as:

\begin{equation}
    (Z^TZ)^+ = V^T\Lambda^+V = V^T\left(\begin{array}{ccc|cc} 
    \frac{1}{\lambda_1} &&&&\\
    & \ddots &&&0 \\
    && \frac{1}{\lambda_{k}}\\
    \hline
    &&&  \\
    &0&&& 0 \end{array}\right)V
\end{equation}

The operation is equivalent to removing the dimensions with the largest singular value in the inverse matrix. This is in line with \cite{chen2019transferability} as it has been shown that penalizing high eigenvalues is beneficial in domain adaptation. 

The selection of $k$~(the number of principal components used) can be achieved through a threshold on the cumulative sum of the eigenvalues of $(Z^TZ)$. Since the smaller eigenvalues do not contribute significantly to the cumulative sum, the corresponding principal components may be continued to be dropped as long as the desired threshold limit is not exceeded. Given $\lambda_s$ and $\lambda_t$ respectively the eigenvalues of the matrix $(Z_s^TZ_s)$ and $(Z_t^TZ_t)$, the goal is to find k, s.t.

\begin{equation}
    \frac{\sum_{i=0}^k \lambda_{s,i}}{\sum_{i=0}^p\lambda_{s,i}} > T \quad \text{and} \quad \frac{\sum_{i=0}^k \lambda_{t,i}}{\sum_{i=0}^p\lambda_{t,i}} > T ,
\end{equation}
\vspace{0.1cm}

\noindent where $T$ is a threshold controlling the proportion of explained variance by the first $k$ principal components.
In the following, the pseudo-inverse with respect to $k$ of the Gram matrix for source and target is denoted as $G_s^+ = (Z_s^TZ_s)^+$ and  $G_t^+ = (Z_t^TZ_t)^+$, respectively. 
Following \cite{chen2021representation}, the cosine similarity is used to calculate the angle difference between source and target. Unlike previous methods, the angle calculation directly uses the column space of $G_s^+$ and $G_t^+$, forming a subspace of $\mathbb{R}^{p}$ spanned by the column vectors of $G_s^+$ and $G_t^+$. A direct measurement of the principal angles is defined as follows:
\vspace{-0.1cm}
\begin{equation}
    \cos(\theta_i^{S \leftrightarrow T}) = \frac{G_{s,i}^+ \cdot G_{t,i}^+}{\lVert G_{s,i}^+\lVert \cdot\lVert G_{t,i}^+\lVert}
\end{equation}

\noindent where $i\in[1,p]$, and $G_{i}^+$ represent the $i$th column of the inverse Gram matrix $G^+$.  The cosine similarity between the span of the subspace for both the source and target feature are stored in $M = [\cos(\theta_1^{S \leftrightarrow T}),\dots, \cos(\theta_p^{S \leftrightarrow T}) ]$. 
The loss to align the selected basis from the pseudo-inverse of the Gram matrix can be written as: 
\begin{equation}
    \mathcal{L}_{cos}(Z^S,Z^T) = \mynorm{\mathbb{I} -M}_1^1
    \label{eq:ismai}
\end{equation}
with $\mathbb{I}$ a vector of ones, of shape $p$. 
Minimizing the above term maximizes the cosine similarity between the source and target representation subspace by reducing the angle between the basis of both domains. 
\paragraph{Discussion}




The proposed method is also more robust and stable compared to the direct feature alignment of $Z$ (\eg RSD~\cite{chen2021representation}). An important difference between RSD and DARE-GRAM lies on the choice of subspace for the alignment. RSD relies on the $U$ basis derived from the SVD decomposition of $Z$. However, the vectors $U$ are first, not unique for a matrix with repeated singular values and, secondly, may be numerically unstable since the gradient depends on $\frac{1}{\lambda_i-\lambda_j}$. 
 Morever, A drawback of RSD is that a large batch size $b \geq p$ can result in full space, causing the principal angles(RSD~\cite{chen2021representation}-Eq.2) between two subspace to become zero. In this case, no alignment can be performed by RSD. Our method does not have this drawback.

\subsection{Scale Alignment}
\label{ssec:Scale}
Preserving the source feature scale is critical in domain adaptive regression problems~\cite{chen2021representation}. In addition to the angle alignment presented in the previous section, we propose to explicitly align the scale of the target subspace to the source. 

More specifically, the scale of the matrix $Z$  can be estimated by 
its trace norm $\lVert Z \lVert_1 = \text{Tr}(\sqrt{Z ^TZ }) = \sum_{i=1}^N\sqrt{\lambda_i}$, where the last term is the sum of the singular values of $Z$. 
The scale of $Z$ is therefore the sum of the diagonal elements of the Gram-Matrix. The scale distance between source and target feature is regularized by minimizing the difference between the k-principal eigenvalues: 

\vspace{-0.5cm}
\begin{equation}
    \mathcal{L}_{scale}(Z^S,Z^T) = \lVert\lambda_{s,i={1,\dots,k}}-\lambda_{t,i={1,\dots,k}}\rVert_2 .
\label{eq:scale_alignement}
\end{equation}


Unlike the previous methods~\cite{chen2021representation}, which explicitly avoid aligning the feature scale, the pseudo-inverse Gram columns that form the basis of our subspace are not necessarily orthonormal. Therefore, matching the source and target basis scale is also essential to complete the alignment process. As a note, the eigenvectors from the SVD decomposition are orthonormal, and the length of the vectors is fixed and set to one, as shown in Figure \ref{fig:Toy}(b).



\subsection{Overview}
 
Combining our angle alignment for the inverse gram and scale alignment, the total loss used for the end-to-end training can be written as:

\begin{equation}
\begin{split}
     \mathcal{L}_{total}(Z^S,Z^T) = &\mathcal{L}_{src} + \alpha_{cos}\mathcal{L}_{cos}(Z^S,Z^T) + \\
     &\gamma_{scale}\mathcal{L}_{scale}(Z^S,Z^T),
\end{split}
\end{equation}
\noindent where $\alpha_{cos},\gamma_{scale}$ are hyper-parameters controlling the effect of the angle and scale alignment. An overall of our method is presented in  Figure \ref{fig:tab}.  

\section{Experiments}
\label{sec:Experiments}

\subsection{Experimental setup}
\label{sec:Experimental setup}

We evaluate our proposed method on three domain adaptations for regression benchmark datasets:  dSprites~\cite{dsprites17}, 
MPI3D~\cite{NEURIPS2019_d97d404b} and  Biwi Kinect \cite{fanelli_IJCV}. 

\begin{figure}[h!]
\centering
\begin{subfigure}{0.14\textwidth}
    \includegraphics[width=\textwidth]{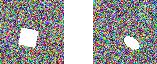}   \caption{Noise}
    \label{fig:noise}
\end{subfigure}
\hspace{1em}%
\begin{subfigure}{0.14\textwidth}
    \includegraphics[width=\textwidth]{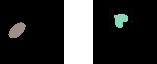}
    \caption{Color}
    \label{fig:Color}
\end{subfigure}
\hspace{1em}%
\begin{subfigure}{0.14\textwidth}
\includegraphics[width=\textwidth]{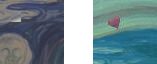}
    \caption{Scream}
    \label{fig:Scream}
\end{subfigure}
\hfill
\vspace{-0.3cm}
\caption{Sample example of different domains in dSprites.}
\label{fig:Dsprites}
\vspace{-0.3cm}
\end{figure}

\noindent\textbf{dSprites} \cite{dsprites17}  is a synthetic 2D dataset generated from five ground truth independent latent factors.
Following common practice~\cite{chen2021representation}, we treat the three variants of the datasets as three different domains. They are generated by adding Color (\textbf{C}) or background noise such as Scream (\textbf{S}) and Noise (\textbf{N}), shown in Figure \ref{fig:Dsprites}. These three domains comprise 737,280 images each. Dsprites can be used as a benchmark for regression domain adaptation, especially if we consider scale, position X, and Y. Similarly to the setup in \cite{chen2021representation}, the orientation factor is excluded from consideration. We evaluate all methods on the three sub-regression tasks  on six adaptation directions: \textbf{C~$\rightarrow$~N}, \textbf{C~$\rightarrow$~S}, \textbf{N~$\rightarrow$~C}, \textbf{N~$\rightarrow$~S}, \textbf{S~$\rightarrow$~C}, and \textbf{S~$\rightarrow$~N}.

\begin{figure}[h]
\centering
\begin{subfigure}{0.14\textwidth}
    \includegraphics[width=\textwidth]{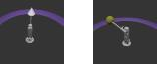}   \caption{Toy}
    \label{fig:TOY}
\end{subfigure}
\hspace{1em}%
\begin{subfigure}{0.14\textwidth}
    \includegraphics[width=\textwidth]{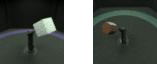}
    \caption{Realistic}
    \label{fig:Realistic}
\end{subfigure}
\hspace{1em}%
\begin{subfigure}{0.14\textwidth}
\includegraphics[width=\textwidth]{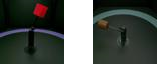}
    \caption{Real}
    \label{fig:REAL}
\end{subfigure}
\hfill
\vspace{-0.3cm}
\caption{Sample example of different domains in MPI3D.}
\label{fig:MPI3D}
\vspace{-0.3cm}
\end{figure}

\noindent\textbf{MPI3D} \cite{NEURIPS2019_d97d404b} is a benchmark dataset that consists of 1,036,800 examples of 3D objects from three different domain : Toy (\textbf{T}), RealistiC (\textbf{RC}) and ReaL (\textbf{RL}), as shown in Figure \ref{fig:MPI3D}. This real-world robotics dataset allows the investigation of the domain gap between real data and simulated ones. This dataset was recorded in a controlled environment, defined by seven factors of variation such as object color, shape, size and position, camera height, background color, and two degrees of freedom of motion of a robotic arm 
The task is to predict these intrinsic factors from the input image. For this paper, we evaluate our method on six transfer tasks: \textbf{RL~$\rightarrow$~RC}, \textbf{RL~$\rightarrow$~T}, \textbf{RC~$\rightarrow$~T},\textbf{RC~$\rightarrow$~RL}, \textbf{T~$\rightarrow$~RL} and \textbf{T~$\rightarrow$~RC}. We only considered the two regression tasks, rotation about a vertical and horizontal axis.
    
\begin{figure}[h]
\centering
\begin{subfigure}{0.22\textwidth}
{\includegraphics[width=0.45\textwidth]{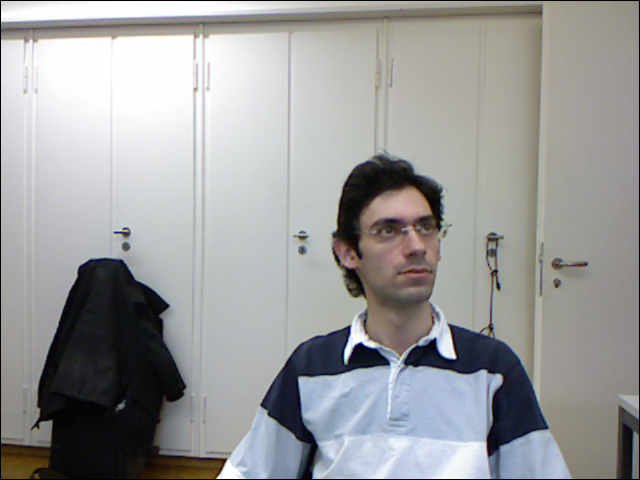}}
{\includegraphics[width=0.45\textwidth]{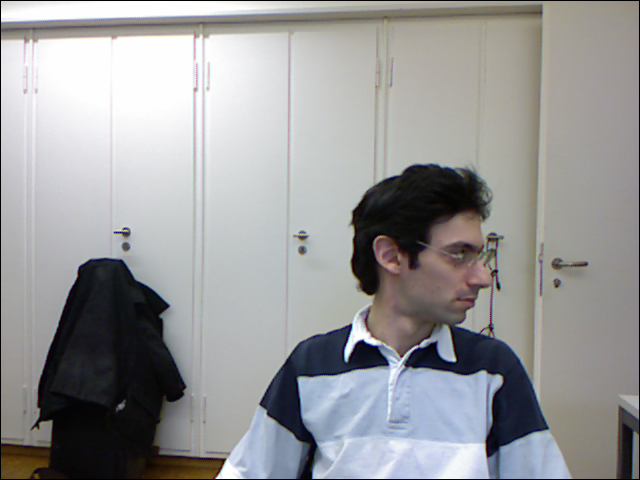}}%
    \caption{Male}
    \label{fig:Male}
\end{subfigure}
\begin{subfigure}{0.22\textwidth}
  {\includegraphics[width=0.45\textwidth]{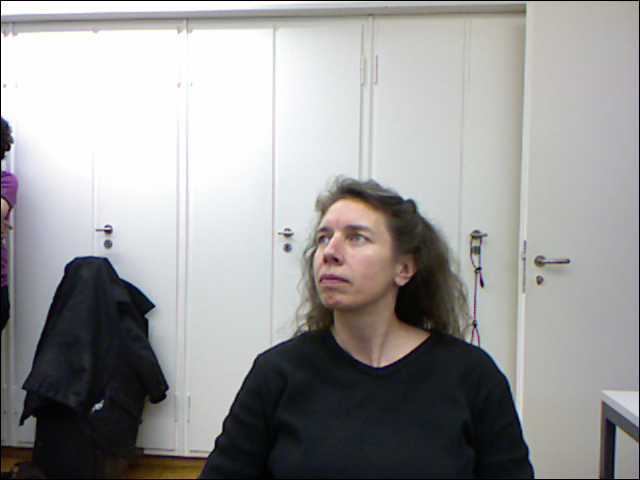}}
{\includegraphics[width=0.45\textwidth]{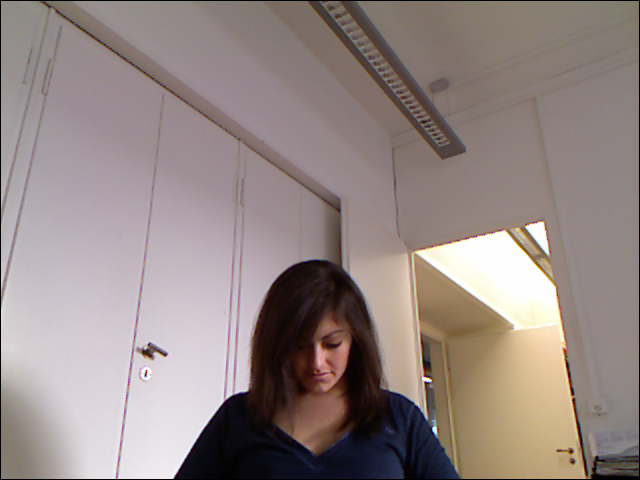}}%
    \caption{Female}
    \label{fig:Female}
\end{subfigure}
\vspace{-0.3cm}
\caption{Sample example of different domains in Biwi kinect}
\label{fig:BIWIKINECT}
\vspace{-0.3cm}
\end{figure}

\noindent\textbf{Biwi kinect} \cite{fanelli_IJCV} is a real-word dataset containing over 15K images of 20 people, 6 Females (\textbf{F}) with 5874 images and 14 Males (\textbf{M}) with 9804 images, recorded with a Microsoft Kinect sensor while turning their heads around freely. The example images are shown in Figure \ref{fig:BIWIKINECT}. The three factors of variations used to evaluate our method are yaw, pitch, and roll angles. 
We evaluate our method on two transfer tasks: \textbf{M~$\rightarrow$~F} and \textbf{F~$\rightarrow$~M}.

\noindent\textbf{Evaluation metrics.} Following previous works~\cite{chen2021representation,Li_2021_CVPR}, Mean Absolute Error (MAE) is used as our evaluation metric across all the regression tasks. Each experiment is repeated three times, and the average results are reported.
\begin{table*}[ht!]
\centering
\small
\setlength{\tabcolsep}{0.05\columnwidth}
\begin{tabular}{lcccccc|c}
\toprule
Method                      & C $\rightarrow$ N & C $\rightarrow$ S & N $\rightarrow$ C & N $\rightarrow$ S & S $\rightarrow$ C & S $\rightarrow$ N  & Avg           \\
\midrule
Resnet-18  \cite{https://doi.org/10.48550/arxiv.1512.03385}                 & 0.94              & 0.90              & 0.16              & 0.65              & 0.08              & 0.26             & 0.498          \\
TCA    \cite{pan2010domain}                   & 0.94               & 0.87              & 0.19              & 0.66              & 0.10              & 0.23             & 0.498 \\
MCD     \cite{saito2018maximum}                & 0.81              & 0.81              & 0.17              & 0.65              & 0.07              & 0.19             & 0.450          \\
JDOT \cite {courty2017joint}  &0.86 & 0.79 & 0.19 & 0.64 & 0.10 & 0.23 & 0.468 \\
AFN     \cite {xu2019larger}                 & 1.00              & 0.96              & 0.16              & 0.62              & 0.08              & 0.32              &0.523          \\
DAN     \cite{long2015learning}                  & 0.70              & 0.77              & 0.12              & 0.50              & 0.06              & 0.11              &0.377          \\
DANN     \cite{ganin2016domain}                   & 0.47              & 0.46              & 0.16              & 0.65              & \textbf{0.05}     & 0.10              &0.315          \\
RSD    \cite{chen2021representation}.                    & 0.31              & 0.31              & 0.12              & 0.53 & 0.07              & 0.08              & 0.237          \\
\textbf{DARE-GRAM (ours)} &\textbf{0.30} & \textbf{0.20} & \textbf{0.11} & \textbf{0.25} & \textbf{0.05} & \textbf{0.07} & \textbf{0.164}\\
\bottomrule
\end{tabular}
\caption{Comparisons with previous works on the dSprites regression tasks. All results are shown in sum of MAE with the ResNet-18.}
\label{tab:results_dsprites}
\end{table*}

\begin{table*}
\centering
\small
\setlength{\tabcolsep}{0.038\columnwidth}
\begin{tabular}{lcccccc|c}
\toprule
Methods                     & RL $\rightarrow$ RC & RL $\rightarrow$ T & RC $\rightarrow$ RL & RC $\rightarrow$ T & T $\rightarrow$ RL & T $\rightarrow$ RC & Avg\\
\midrule
Resnet-18  \cite{https://doi.org/10.48550/arxiv.1512.03385}                 & 0.17                & 0.44               & 0.19                & 0.45               & 0.51                                       & 0.50               & 0.377          \\
TCA  \cite{pan2010domain}                 & 0.17                & 0.42               & 0.19                & 0.42              & 0.50                                       & 0.50               & 0.373          \\
MCD      \cite{saito2018maximum}                   & 0.13                & 0.40               & 0.15                & 0.45               & 0.52                                       & 0.50               & 0.358          \\
JDOT        \cite {courty2017joint}                  & 0.16                & 0.41               & 0.16                & 0.41               & 0.47                                       & 0.47               & 0.353         \\
AFN   \cite {xu2019larger}                     & 0.18                & 0.45               & 0.20                & 0.46               & 0.53                                       & 0.53              & 0.390          \\
DAN     \cite{long2015learning}                    & 0.12                & 0.35               & 0.12                & 0.27               & 0.40                                       & 0.41               & 0.278          \\
DANN    \cite{ganin2016domain}                     & 0.09                & 0.24               & 0.11                & 0.41               & 0.48                                       & 0.37               & 0.283          \\
RSD       \cite{chen2021representation}.                  & \textbf{0.09}                & 0.19               & \textbf{0.08}               & 0.15               & 0.36                                       & 0.36               & 0.205          \\
\textbf{DARE-GRAM (ours)} & \textbf{0.09} & \textbf{0.15} & 0.10 & \textbf{0.14} & \textbf{0.24} & \textbf{0.24} & \textbf{0.160}\\
\bottomrule
\end{tabular}
\caption{Comparisons with related works on the MPI3D regression tasks. All results are shown in sum of MAE with the ResNet-18.}
\label{tab:results_MPI3D}
\end{table*}


 
\noindent\textbf{Implementation Details.} A pre-trained ResNet-18 \cite{https://doi.org/10.48550/arxiv.1512.03385} on ImageNet is used as the backbone for all methods. For all the experiments, the different tasks share the same encoder but a separated single linear regressor with a Sigmoid activation function. The source and target labels were scaled in the range $[0, 1]$ to eliminate the effects of diverse scales in regression values. We use the SGD~\cite{https://doi.org/10.48550/arxiv.1609.04747} optimizer with
a momentum of 0.9. The weight decay is set to $1e^{-3}$ for the loss optimization. The newly added layers are trained with a learning rate ten times that of the pre-trained layers, which is initialized to 	$\eta_0 = 1e^{-2}$. 
We further adopt the same learning rate scheduler $\eta = \eta_0 \cdot(1 + 0.0001 \cdot p)^{-0.75}$ as \cite{ganin2015unsupervised,long2018conditional}, where $p$ is the number of iterations changing from 0 to the maximum number of iterations. The images are resized to 224 × 224 and concatenated into batches of size $b=36$. The number of iterations was set as in \cite{chen2021representation} to 20,000, 10,000, and 1,500 iterations for dSprites, MPI3D, and Biwi Kinect, respectively. These setup choices are identical to RSD~\cite{chen2021representation}. An NVIDIA RTX 3090 GPU was used for all the experiments. 

\noindent\textbf{Compared Methods}  We compare our method with a range of adaptation methods: (i) Domain Adaptation via Transfer Component Analysis (\textbf{TCA})~\cite{pan2010domain}
(ii) Maximum Classifier Discrepancy (\textbf{MCD}) \cite{saito2018maximum}
(iii) Joint Distribution Optimal Transportation for Domain Adaptation (\textbf{JDOT}) \cite {courty2017joint} (iv) Adaptive Feature Norm (\textbf{AFN}) \cite {xu2019larger}
(v) Deep Adaptation Network (\textbf{DAN}) \cite{long2015learning}
(vi) Deep Adaptation Neural Network (\textbf{DANN}) \cite{ganin2016domain} and (vii) Representation Subspace Distance for Domain Adaptation Regression (\textbf{RSD}) \cite{chen2021representation}.


\subsection{Results}

\noindent\textbf{Evaluation on dSprites:} As shown in Table \ref{tab:results_dsprites}, our model achieves the best performance among all competing methods. Specifically, our method outperforms the previous state-of-the-art regression-based method RSD\cite{chen2021representation} by $30.8\%$ in terms of average MSE over all directions. 
On the three difficult adaptation directions \textbf{C~$\rightarrow$~N}, \textbf{C~$\rightarrow$~S}, \textbf{N~$\rightarrow$~S}. \textbf{DARE-GRAM} also improves the performance over the RSD. The improvement is especially significant on the direction \textbf{C~$\rightarrow$~S} and \textbf{N~$\rightarrow$~S}. The improvement is by $33.3\%$ and $52.8\%$, respectively. 
\noindent\textbf{Evaluation on MPI3D:} We further evaluate the effectiveness of our method on this more complex simulation-real data set. As shown in Table \ref{tab:results_MPI3D}, in average, our method outperforms all previous methods. The improvement over the previous state-of-the-art RSD is more than 21.9\%.
The improvement is especially significant in the four hard adaptation directions \textbf{T~$\rightarrow$~RL}, \textbf{T~$\rightarrow$~RC}, \textbf{RL~$\rightarrow$~T}, and \textbf{RC~$\rightarrow$~T}. The performance is comparable with RSD on the \textbf{RC/RL} pair. This might be because the domain gap is relatively small between the pair and the performance~($\approx 0.1$) could be close to saturation.

\noindent\textbf{Evaluation on Biwi Kinect:} Given the much smaller size of the dataset (15,000 images compared to the number of samples in the scale of millions in the other two datasets), the Biwi Kinect regression task is particularly challenging. Additionally, the high imbalance between the two domains and the lack of separation of training and testing sets makes it more difficult and closer to real-world scenarios for DAR. The results reported in Table \ref{tab:results_biwi} demonstrate that our model can also consistently improve over previous methods on both directions on this more challenging task. 

The performance improvement on the three datasets of very different natures demonstrates the effectiveness of our proposed method.

\begin{table}
\centering
\small
\begin{tabular}{lcc|c}
\toprule
Method   & M $\rightarrow$ F & F $\rightarrow$ M & Avg\\
\midrule
Resnet-18 \cite{https://doi.org/10.48550/arxiv.1512.03385} & 0.29 & 0.38 & 0.335\\
TCA \cite{pan2010domain} & 0.31 & 0.39 & 0.350 \\
MCD \cite{saito2018maximum} & 0.31 & 0.37 & 0.340 \\
JDOT \cite {courty2017joint} & 0.29 & 0.39  & 0.340 \\
AFN \cite {xu2019larger} & 0.32 & 0.41 & 0.365 \\
DAN \cite{long2015learning} & 0.28 & 0.37 & 0.325 \\
DANN \cite{ganin2016domain}  & 0.30 & 0.37 & 0.335 \\
RSD \cite{chen2021representation} & 0.26 & 0.30 & 0.280 \\
\textbf{DARE-GRAM (ours)} & \textbf{0.23} & \textbf{0.29} & \textbf{0.260} \\
\bottomrule
\end{tabular}
\caption{Comparisons with previous works on Biwi Kinect.}
\label{tab:results_biwi}
\vspace{-0.5cm}
\end{table}

\subsection{Discussion and Analysis}
To provide more insights on the proposed Unsupervised Domain Adaptation Regression by Aligning Inverse Gram Matrices, we provide a detailed analysis of the different components of the methodology. 

\noindent\textbf{Angle Alignment and Scale Alignment} In the first ablation, we study the impact of our angle alignment on the inverse Gram matrix and our scale alignment on the eigenvalues. The $\textbf{C~$\rightarrow$~S}$ and $\textbf{N~$\rightarrow$~S}$ tasks in dSprites are used for the ablation here. As shown in Table~\ref{tab:loss_ablataion}, both components of the proposed methodology are able to improve over the baseline. Minimizing the angle between the pseudo-inverse of the gram Matrix can reduce the MSE over the source-only baseline by 70\% on the $\textbf{C~$\rightarrow$~S}$ task. We also compare the alternate angle alignment on the Gram matrix and truncated Gram matrix without considering the inverse. The MAE in both cases is significantly worst than our inverse version. This demonstrates the significant impact of the inverse operation~(as in Equation~\ref{eq:zst}) on the regression layer and verifies our motivation for considering the closed-form solution of OLS regression problems. In addition, as discussed in Section~\ref{ssec:Scale}, the scaling constraints provide essential additional supervision on the alignment and further improve the model performance. Both terms are effective in improving the performance. 

\begin{table}
\centering
\small
\begin{tabular}{lcc}
\hline
Method   & C $\rightarrow$ S & N $\rightarrow$ S \\
\hline
Resnet-18 (source only) & 0.90 & 0.65\\
RSD & 0.31 & 0.53\\
Angle Alignment for Gram   & 0.88 & 0.55\\
Angle Alignment for truncated Gram & 0.89 & 0.52 \\
Angle Alignment for Gram  Inverse~(ours) & 0.27 & 0.36\\
Scale Alignment~(ours) & 0.23 & 0.60 \\
\textbf{DARE-GRAM (ours, angle + scale)} & \textbf{0.20} & \textbf{0.25}\\
\hline
\end{tabular}

\caption{Ablation study of different components in our proposed method on C $\rightarrow$ S and N $\rightarrow$ S task from dSprites. All results are shown in sum of MAE.}
\label{tab:loss_ablataion}
\vspace{-0.5cm}
\end{table}

\begin{figure}
\begin{center}
    \scalebox{0.4}{\input{images/ablation/batch_ablation_1.pgf}}
    \caption{Batch size sensitivity on transfer task $\textbf{C~$\rightarrow$~S}$. For RSD, larger batch sizes lead to numerical errors, thus results not shown. Our method DARE-GRAM is able to achieve better performance with large batch sizes because the feature correlations can be better captured by the Gram matrix with larger number of samples.}
    \label{fig:batch}
\end{center}
\vspace{-0.5cm}
\end{figure}
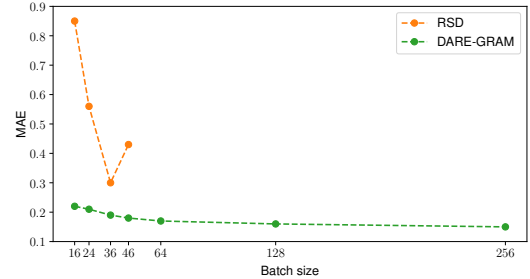

\noindent\textbf{Effect of a Larger Batch Size} Given the direct relationship between the number of samples in a batch to the variance of the estimated $\beta$ in Equation~\ref{eq:zst}, we studied the effect of the different training batch sizes in Figure \ref{fig:batch} (using the same hyperparameters). As the batch size increases, our approach results in lower MAE.  Furthermore, the RSD approach is more sensitive to the batch size and leads to numerical errors for batches bigger than $64$. In this paper, we used only the same batch size of 36 to have a fair comparison with RSD. However, better results can be achieved with our method by further increasing the batch size to 256.

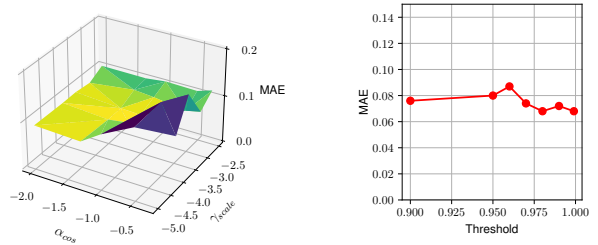
\begin{figure}
\begin{center}
\begin{subfigure}{0.23\textwidth}
    \scalebox{0.45}{\input{images/ablation/sensitivity_s2n.pgf}}
    \caption{Impact of $\alpha_{cos}$ and $\gamma_{scale}$ in log scale on the sum of MAE. }
    \label{fig:alpha_gamma}
\end{subfigure}
\hfill
\begin{subfigure}{0.21\textwidth}
    \scalebox{0.45}{\input{images/ablation/thresold_s2n.pgf}}
    \caption{Impact of Threshold $T$ on the sum of MAE}
    \label{fig:Threshold}
\end{subfigure}
\end{center}
\vspace{-0.4cm}
\caption{Hyperparameter sensitivity of our method on dSprites transfer task $\textbf{S~$\rightarrow$~N}$.}
\label{fig:hyper}
\vspace{-0.5cm}
\end{figure}

\noindent\textbf{Effect of alignment factors:} We conduct additional experiments to evaluate the impact on the performance when using different values of hyperparameter $\alpha_{cos}$, $\gamma_{scale}$ and the threshold $T$. As shown in Figure \ref{fig:hyper}, results confirm that our method is not sensitive to hyperparameters. 

\noindent\textbf{Alignment performance of Z and $({Z^TZ})^{-1}$} To further validate the proposed method, we examined the cosine similarity of the k-principal components of ${Z_s, Z_t}$, as well as ${(Z^T_s Z_s)^{-1}, (Z^T_t Z_t)^{-1}}$, after applying our proposed method and RSD. The results are shown in Table A2 in the supplementary material. Our results demonstrate that aligning $Z$ by RSD can lead to poorly aligned inverse Gram matrix $\boldsymbol({Z^TZ})^{-1}$. In contrast, by aligning $({Z^TZ})^{-1}$, our proposed method leads to a well-aligned $Z$,  providing further empirical support for the effectiveness of our proposed method. 

\section{Conclusion}

In this paper, we have presented a new domain adaptative regression method called DARE-GRAM. We tackled the domain adaptation for regression problems from a different perspective analyzing the ordinary least square solution to the linear regressor in the deep domain adaptation context. Rather than aligning the original feature embedding space, we aligned a selected subspace of the pseudo-inverse Gram matrix, leveraging the pseudo-inverse low-rank property. Finally, two new regularization terms were proposed to align the scale and angle in a selected subspace generated by the Gram matrix of the two domains. Experimental results show that DARE-GRAM achieves significant improvement in three benchmark regression datasets while ensuring the stability and robustness of the training procedure.

\noindent\textbf{Acknowledgments:} This work was supported by
the Swiss National Science Foundation under Grant PP00P2$\_$176878.

{\small
\bibliographystyle{ieee_fullname}
\bibliography{egbib.bib}
}

\end{document}

%% file: images/ablation/batch_ablation_1.pgf
\begingroup%
\makeatletter%
\begin{pgfpicture}%
\pgfpathrectangle{\pgfpointorigin}{\pgfqpoint{6.922597in}{3.818682in}}%
\pgfusepath{use as bounding box, clip}%
\begin{pgfscope}%
\pgfsetbuttcap%
\pgfsetmiterjoin%
\definecolor{currentfill}{rgb}{1.000000,1.000000,1.000000}%
\pgfsetfillcolor{currentfill}%
\pgfsetlinewidth{0.000000pt}%
\definecolor{currentstroke}{rgb}{1.000000,1.000000,1.000000}%
\pgfsetstrokecolor{currentstroke}%
\pgfsetdash{}{0pt}%
\pgfpathmoveto{\pgfqpoint{0.000000in}{0.000000in}}%
\pgfpathlineto{\pgfqpoint{6.922597in}{0.000000in}}%
\pgfpathlineto{\pgfqpoint{6.922597in}{3.818682in}}%
\pgfpathlineto{\pgfqpoint{0.000000in}{3.818682in}}%
\pgfpathlineto{\pgfqpoint{0.000000in}{0.000000in}}%
\pgfpathclose%
\pgfusepath{fill}%
\end{pgfscope}%
\begin{pgfscope}%
\pgfsetbuttcap%
\pgfsetmiterjoin%
\definecolor{currentfill}{rgb}{1.000000,1.000000,1.000000}%
\pgfsetfillcolor{currentfill}%
\pgfsetlinewidth{0.000000pt}%
\definecolor{currentstroke}{rgb}{0.000000,0.000000,0.000000}%
\pgfsetstrokecolor{currentstroke}%
\pgfsetstrokeopacity{0.000000}%
\pgfsetdash{}{0pt}%
\pgfpathmoveto{\pgfqpoint{0.622597in}{0.575369in}}%
\pgfpathlineto{\pgfqpoint{6.822597in}{0.575369in}}%
\pgfpathlineto{\pgfqpoint{6.822597in}{3.655369in}}%
\pgfpathlineto{\pgfqpoint{0.622597in}{3.655369in}}%
\pgfpathlineto{\pgfqpoint{0.622597in}{0.575369in}}%
\pgfpathclose%
\pgfusepath{fill}%
\end{pgfscope}%
\begin{pgfscope}%
\pgfsetbuttcap%
\pgfsetroundjoin%
\definecolor{currentfill}{rgb}{0.000000,0.000000,0.000000}%
\pgfsetfillcolor{currentfill}%
\pgfsetlinewidth{0.803000pt}%
\definecolor{currentstroke}{rgb}{0.000000,0.000000,0.000000}%
\pgfsetstrokecolor{currentstroke}%
\pgfsetdash{}{0pt}%
\pgfsys@defobject{currentmarker}{\pgfqpoint{0.000000in}{-0.048611in}}{\pgfqpoint{0.000000in}{0.000000in}}{%
\pgfpathmoveto{\pgfqpoint{0.000000in}{0.000000in}}%
\pgfpathlineto{\pgfqpoint{0.000000in}{-0.048611in}}%
\pgfusepath{stroke,fill}%
}%
\begin{pgfscope}%
\pgfsys@transformshift{0.904415in}{0.575369in}%
\pgfsys@useobject{currentmarker}{}%
\end{pgfscope}%
\end{pgfscope}%
\begin{pgfscope}%
\definecolor{textcolor}{rgb}{0.000000,0.000000,0.000000}%
\pgfsetstrokecolor{textcolor}%
\pgfsetfillcolor{textcolor}%
\pgftext[x=0.904415in,y=0.478146in,,top]{\color{textcolor}\sffamily\fontsize{12.000000}{14.400000}\selectfont \(\displaystyle {16}\)}%
\end{pgfscope}%
\begin{pgfscope}%
\pgfsetbuttcap%
\pgfsetroundjoin%
\definecolor{currentfill}{rgb}{0.000000,0.000000,0.000000}%
\pgfsetfillcolor{currentfill}%
\pgfsetlinewidth{0.803000pt}%
\definecolor{currentstroke}{rgb}{0.000000,0.000000,0.000000}%
\pgfsetstrokecolor{currentstroke}%
\pgfsetdash{}{0pt}%
\pgfsys@defobject{currentmarker}{\pgfqpoint{0.000000in}{-0.048611in}}{\pgfqpoint{0.000000in}{0.000000in}}{%
\pgfpathmoveto{\pgfqpoint{0.000000in}{0.000000in}}%
\pgfpathlineto{\pgfqpoint{0.000000in}{-0.048611in}}%
\pgfusepath{stroke,fill}%
}%
\begin{pgfscope}%
\pgfsys@transformshift{1.092294in}{0.575369in}%
\pgfsys@useobject{currentmarker}{}%
\end{pgfscope}%
\end{pgfscope}%
\begin{pgfscope}%
\definecolor{textcolor}{rgb}{0.000000,0.000000,0.000000}%
\pgfsetstrokecolor{textcolor}%
\pgfsetfillcolor{textcolor}%
\pgftext[x=1.092294in,y=0.478146in,,top]{\color{textcolor}\sffamily\fontsize{12.000000}{14.400000}\selectfont \(\displaystyle {24}\)}%
\end{pgfscope}%
\begin{pgfscope}%
\pgfsetbuttcap%
\pgfsetroundjoin%
\definecolor{currentfill}{rgb}{0.000000,0.000000,0.000000}%
\pgfsetfillcolor{currentfill}%
\pgfsetlinewidth{0.803000pt}%
\definecolor{currentstroke}{rgb}{0.000000,0.000000,0.000000}%
\pgfsetstrokecolor{currentstroke}%
\pgfsetdash{}{0pt}%
\pgfsys@defobject{currentmarker}{\pgfqpoint{0.000000in}{-0.048611in}}{\pgfqpoint{0.000000in}{0.000000in}}{%
\pgfpathmoveto{\pgfqpoint{0.000000in}{0.000000in}}%
\pgfpathlineto{\pgfqpoint{0.000000in}{-0.048611in}}%
\pgfusepath{stroke,fill}%
}%
\begin{pgfscope}%
\pgfsys@transformshift{1.374112in}{0.575369in}%
\pgfsys@useobject{currentmarker}{}%
\end{pgfscope}%
\end{pgfscope}%
\begin{pgfscope}%
\definecolor{textcolor}{rgb}{0.000000,0.000000,0.000000}%
\pgfsetstrokecolor{textcolor}%
\pgfsetfillcolor{textcolor}%
\pgftext[x=1.374112in,y=0.478146in,,top]{\color{textcolor}\sffamily\fontsize{12.000000}{14.400000}\selectfont \(\displaystyle {36}\)}%
\end{pgfscope}%
\begin{pgfscope}%
\pgfsetbuttcap%
\pgfsetroundjoin%
\definecolor{currentfill}{rgb}{0.000000,0.000000,0.000000}%
\pgfsetfillcolor{currentfill}%
\pgfsetlinewidth{0.803000pt}%
\definecolor{currentstroke}{rgb}{0.000000,0.000000,0.000000}%
\pgfsetstrokecolor{currentstroke}%
\pgfsetdash{}{0pt}%
\pgfsys@defobject{currentmarker}{\pgfqpoint{0.000000in}{-0.048611in}}{\pgfqpoint{0.000000in}{0.000000in}}{%
\pgfpathmoveto{\pgfqpoint{0.000000in}{0.000000in}}%
\pgfpathlineto{\pgfqpoint{0.000000in}{-0.048611in}}%
\pgfusepath{stroke,fill}%
}%
\begin{pgfscope}%
\pgfsys@transformshift{1.608961in}{0.575369in}%
\pgfsys@useobject{currentmarker}{}%
\end{pgfscope}%
\end{pgfscope}%
\begin{pgfscope}%
\definecolor{textcolor}{rgb}{0.000000,0.000000,0.000000}%
\pgfsetstrokecolor{textcolor}%
\pgfsetfillcolor{textcolor}%
\pgftext[x=1.608961in,y=0.478146in,,top]{\color{textcolor}\sffamily\fontsize{12.000000}{14.400000}\selectfont \(\displaystyle {46}\)}%
\end{pgfscope}%
\begin{pgfscope}%
\pgfsetbuttcap%
\pgfsetroundjoin%
\definecolor{currentfill}{rgb}{0.000000,0.000000,0.000000}%
\pgfsetfillcolor{currentfill}%
\pgfsetlinewidth{0.803000pt}%
\definecolor{currentstroke}{rgb}{0.000000,0.000000,0.000000}%
\pgfsetstrokecolor{currentstroke}%
\pgfsetdash{}{0pt}%
\pgfsys@defobject{currentmarker}{\pgfqpoint{0.000000in}{-0.048611in}}{\pgfqpoint{0.000000in}{0.000000in}}{%
\pgfpathmoveto{\pgfqpoint{0.000000in}{0.000000in}}%
\pgfpathlineto{\pgfqpoint{0.000000in}{-0.048611in}}%
\pgfusepath{stroke,fill}%
}%
\begin{pgfscope}%
\pgfsys@transformshift{2.031688in}{0.575369in}%
\pgfsys@useobject{currentmarker}{}%
\end{pgfscope}%
\end{pgfscope}%
\begin{pgfscope}%
\definecolor{textcolor}{rgb}{0.000000,0.000000,0.000000}%
\pgfsetstrokecolor{textcolor}%
\pgfsetfillcolor{textcolor}%
\pgftext[x=2.031688in,y=0.478146in,,top]{\color{textcolor}\sffamily\fontsize{12.000000}{14.400000}\selectfont \(\displaystyle {64}\)}%
\end{pgfscope}%
\begin{pgfscope}%
\pgfsetbuttcap%
\pgfsetroundjoin%
\definecolor{currentfill}{rgb}{0.000000,0.000000,0.000000}%
\pgfsetfillcolor{currentfill}%
\pgfsetlinewidth{0.803000pt}%
\definecolor{currentstroke}{rgb}{0.000000,0.000000,0.000000}%
\pgfsetstrokecolor{currentstroke}%
\pgfsetdash{}{0pt}%
\pgfsys@defobject{currentmarker}{\pgfqpoint{0.000000in}{-0.048611in}}{\pgfqpoint{0.000000in}{0.000000in}}{%
\pgfpathmoveto{\pgfqpoint{0.000000in}{0.000000in}}%
\pgfpathlineto{\pgfqpoint{0.000000in}{-0.048611in}}%
\pgfusepath{stroke,fill}%
}%
\begin{pgfscope}%
\pgfsys@transformshift{3.534718in}{0.575369in}%
\pgfsys@useobject{currentmarker}{}%
\end{pgfscope}%
\end{pgfscope}%
\begin{pgfscope}%
\definecolor{textcolor}{rgb}{0.000000,0.000000,0.000000}%
\pgfsetstrokecolor{textcolor}%
\pgfsetfillcolor{textcolor}%
\pgftext[x=3.534718in,y=0.478146in,,top]{\color{textcolor}\sffamily\fontsize{12.000000}{14.400000}\selectfont \(\displaystyle {128}\)}%
\end{pgfscope}%
\begin{pgfscope}%
\pgfsetbuttcap%
\pgfsetroundjoin%
\definecolor{currentfill}{rgb}{0.000000,0.000000,0.000000}%
\pgfsetfillcolor{currentfill}%
\pgfsetlinewidth{0.803000pt}%
\definecolor{currentstroke}{rgb}{0.000000,0.000000,0.000000}%
\pgfsetstrokecolor{currentstroke}%
\pgfsetdash{}{0pt}%
\pgfsys@defobject{currentmarker}{\pgfqpoint{0.000000in}{-0.048611in}}{\pgfqpoint{0.000000in}{0.000000in}}{%
\pgfpathmoveto{\pgfqpoint{0.000000in}{0.000000in}}%
\pgfpathlineto{\pgfqpoint{0.000000in}{-0.048611in}}%
\pgfusepath{stroke,fill}%
}%
\begin{pgfscope}%
\pgfsys@transformshift{6.540779in}{0.575369in}%
\pgfsys@useobject{currentmarker}{}%
\end{pgfscope}%
\end{pgfscope}%
\begin{pgfscope}%
\definecolor{textcolor}{rgb}{0.000000,0.000000,0.000000}%
\pgfsetstrokecolor{textcolor}%
\pgfsetfillcolor{textcolor}%
\pgftext[x=6.540779in,y=0.478146in,,top]{\color{textcolor}\sffamily\fontsize{12.000000}{14.400000}\selectfont \(\displaystyle {256}\)}%
\end{pgfscope}%
\begin{pgfscope}%
\definecolor{textcolor}{rgb}{0.000000,0.000000,0.000000}%
\pgfsetstrokecolor{textcolor}%
\pgfsetfillcolor{textcolor}%
\pgftext[x=3.722597in,y=0.261295in,,top]{\color{textcolor}\sffamily\fontsize{12.000000}{14.400000}\selectfont Batch size}%
\end{pgfscope}%
\begin{pgfscope}%
\pgfsetbuttcap%
\pgfsetroundjoin%
\definecolor{currentfill}{rgb}{0.000000,0.000000,0.000000}%
\pgfsetfillcolor{currentfill}%
\pgfsetlinewidth{0.803000pt}%
\definecolor{currentstroke}{rgb}{0.000000,0.000000,0.000000}%
\pgfsetstrokecolor{currentstroke}%
\pgfsetdash{}{0pt}%
\pgfsys@defobject{currentmarker}{\pgfqpoint{-0.048611in}{0.000000in}}{\pgfqpoint{-0.000000in}{0.000000in}}{%
\pgfpathmoveto{\pgfqpoint{-0.000000in}{0.000000in}}%
\pgfpathlineto{\pgfqpoint{-0.048611in}{0.000000in}}%
\pgfusepath{stroke,fill}%
}%
\begin{pgfscope}%
\pgfsys@transformshift{0.622597in}{0.575369in}%
\pgfsys@useobject{currentmarker}{}%
\end{pgfscope}%
\end{pgfscope}%
\begin{pgfscope}%
\definecolor{textcolor}{rgb}{0.000000,0.000000,0.000000}%
\pgfsetstrokecolor{textcolor}%
\pgfsetfillcolor{textcolor}%
\pgftext[x=0.316851in, y=0.512055in, left, base]{\color{textcolor}\sffamily\fontsize{12.000000}{14.400000}\selectfont \(\displaystyle {0.1}\)}%
\end{pgfscope}%
\begin{pgfscope}%
\pgfsetbuttcap%
\pgfsetroundjoin%
\definecolor{currentfill}{rgb}{0.000000,0.000000,0.000000}%
\pgfsetfillcolor{currentfill}%
\pgfsetlinewidth{0.803000pt}%
\definecolor{currentstroke}{rgb}{0.000000,0.000000,0.000000}%
\pgfsetstrokecolor{currentstroke}%
\pgfsetdash{}{0pt}%
\pgfsys@defobject{currentmarker}{\pgfqpoint{-0.048611in}{0.000000in}}{\pgfqpoint{-0.000000in}{0.000000in}}{%
\pgfpathmoveto{\pgfqpoint{-0.000000in}{0.000000in}}%
\pgfpathlineto{\pgfqpoint{-0.048611in}{0.000000in}}%
\pgfusepath{stroke,fill}%
}%
\begin{pgfscope}%
\pgfsys@transformshift{0.622597in}{0.960369in}%
\pgfsys@useobject{currentmarker}{}%
\end{pgfscope}%
\end{pgfscope}%
\begin{pgfscope}%
\definecolor{textcolor}{rgb}{0.000000,0.000000,0.000000}%
\pgfsetstrokecolor{textcolor}%
\pgfsetfillcolor{textcolor}%
\pgftext[x=0.316851in, y=0.897055in, left, base]{\color{textcolor}\sffamily\fontsize{12.000000}{14.400000}\selectfont \(\displaystyle {0.2}\)}%
\end{pgfscope}%
\begin{pgfscope}%
\pgfsetbuttcap%
\pgfsetroundjoin%
\definecolor{currentfill}{rgb}{0.000000,0.000000,0.000000}%
\pgfsetfillcolor{currentfill}%
\pgfsetlinewidth{0.803000pt}%
\definecolor{currentstroke}{rgb}{0.000000,0.000000,0.000000}%
\pgfsetstrokecolor{currentstroke}%
\pgfsetdash{}{0pt}%
\pgfsys@defobject{currentmarker}{\pgfqpoint{-0.048611in}{0.000000in}}{\pgfqpoint{-0.000000in}{0.000000in}}{%
\pgfpathmoveto{\pgfqpoint{-0.000000in}{0.000000in}}%
\pgfpathlineto{\pgfqpoint{-0.048611in}{0.000000in}}%
\pgfusepath{stroke,fill}%
}%
\begin{pgfscope}%
\pgfsys@transformshift{0.622597in}{1.345369in}%
\pgfsys@useobject{currentmarker}{}%
\end{pgfscope}%
\end{pgfscope}%
\begin{pgfscope}%
\definecolor{textcolor}{rgb}{0.000000,0.000000,0.000000}%
\pgfsetstrokecolor{textcolor}%
\pgfsetfillcolor{textcolor}%
\pgftext[x=0.316851in, y=1.282055in, left, base]{\color{textcolor}\sffamily\fontsize{12.000000}{14.400000}\selectfont \(\displaystyle {0.3}\)}%
\end{pgfscope}%
\begin{pgfscope}%
\pgfsetbuttcap%
\pgfsetroundjoin%
\definecolor{currentfill}{rgb}{0.000000,0.000000,0.000000}%
\pgfsetfillcolor{currentfill}%
\pgfsetlinewidth{0.803000pt}%
\definecolor{currentstroke}{rgb}{0.000000,0.000000,0.000000}%
\pgfsetstrokecolor{currentstroke}%
\pgfsetdash{}{0pt}%
\pgfsys@defobject{currentmarker}{\pgfqpoint{-0.048611in}{0.000000in}}{\pgfqpoint{-0.000000in}{0.000000in}}{%
\pgfpathmoveto{\pgfqpoint{-0.000000in}{0.000000in}}%
\pgfpathlineto{\pgfqpoint{-0.048611in}{0.000000in}}%
\pgfusepath{stroke,fill}%
}%
\begin{pgfscope}%
\pgfsys@transformshift{0.622597in}{1.730369in}%
\pgfsys@useobject{currentmarker}{}%
\end{pgfscope}%
\end{pgfscope}%
\begin{pgfscope}%
\definecolor{textcolor}{rgb}{0.000000,0.000000,0.000000}%
\pgfsetstrokecolor{textcolor}%
\pgfsetfillcolor{textcolor}%
\pgftext[x=0.316851in, y=1.667055in, left, base]{\color{textcolor}\sffamily\fontsize{12.000000}{14.400000}\selectfont \(\displaystyle {0.4}\)}%
\end{pgfscope}%
\begin{pgfscope}%
\pgfsetbuttcap%
\pgfsetroundjoin%
\definecolor{currentfill}{rgb}{0.000000,0.000000,0.000000}%
\pgfsetfillcolor{currentfill}%
\pgfsetlinewidth{0.803000pt}%
\definecolor{currentstroke}{rgb}{0.000000,0.000000,0.000000}%
\pgfsetstrokecolor{currentstroke}%
\pgfsetdash{}{0pt}%
\pgfsys@defobject{currentmarker}{\pgfqpoint{-0.048611in}{0.000000in}}{\pgfqpoint{-0.000000in}{0.000000in}}{%
\pgfpathmoveto{\pgfqpoint{-0.000000in}{0.000000in}}%
\pgfpathlineto{\pgfqpoint{-0.048611in}{0.000000in}}%
\pgfusepath{stroke,fill}%
}%
\begin{pgfscope}%
\pgfsys@transformshift{0.622597in}{2.115369in}%
\pgfsys@useobject{currentmarker}{}%
\end{pgfscope}%
\end{pgfscope}%
\begin{pgfscope}%
\definecolor{textcolor}{rgb}{0.000000,0.000000,0.000000}%
\pgfsetstrokecolor{textcolor}%
\pgfsetfillcolor{textcolor}%
\pgftext[x=0.316851in, y=2.052055in, left, base]{\color{textcolor}\sffamily\fontsize{12.000000}{14.400000}\selectfont \(\displaystyle {0.5}\)}%
\end{pgfscope}%
\begin{pgfscope}%
\pgfsetbuttcap%
\pgfsetroundjoin%
\definecolor{currentfill}{rgb}{0.000000,0.000000,0.000000}%
\pgfsetfillcolor{currentfill}%
\pgfsetlinewidth{0.803000pt}%
\definecolor{currentstroke}{rgb}{0.000000,0.000000,0.000000}%
\pgfsetstrokecolor{currentstroke}%
\pgfsetdash{}{0pt}%
\pgfsys@defobject{currentmarker}{\pgfqpoint{-0.048611in}{0.000000in}}{\pgfqpoint{-0.000000in}{0.000000in}}{%
\pgfpathmoveto{\pgfqpoint{-0.000000in}{0.000000in}}%
\pgfpathlineto{\pgfqpoint{-0.048611in}{0.000000in}}%
\pgfusepath{stroke,fill}%
}%
\begin{pgfscope}%
\pgfsys@transformshift{0.622597in}{2.500369in}%
\pgfsys@useobject{currentmarker}{}%
\end{pgfscope}%
\end{pgfscope}%
\begin{pgfscope}%
\definecolor{textcolor}{rgb}{0.000000,0.000000,0.000000}%
\pgfsetstrokecolor{textcolor}%
\pgfsetfillcolor{textcolor}%
\pgftext[x=0.316851in, y=2.437055in, left, base]{\color{textcolor}\sffamily\fontsize{12.000000}{14.400000}\selectfont \(\displaystyle {0.6}\)}%
\end{pgfscope}%
\begin{pgfscope}%
\pgfsetbuttcap%
\pgfsetroundjoin%
\definecolor{currentfill}{rgb}{0.000000,0.000000,0.000000}%
\pgfsetfillcolor{currentfill}%
\pgfsetlinewidth{0.803000pt}%
\definecolor{currentstroke}{rgb}{0.000000,0.000000,0.000000}%
\pgfsetstrokecolor{currentstroke}%
\pgfsetdash{}{0pt}%
\pgfsys@defobject{currentmarker}{\pgfqpoint{-0.048611in}{0.000000in}}{\pgfqpoint{-0.000000in}{0.000000in}}{%
\pgfpathmoveto{\pgfqpoint{-0.000000in}{0.000000in}}%
\pgfpathlineto{\pgfqpoint{-0.048611in}{0.000000in}}%
\pgfusepath{stroke,fill}%
}%
\begin{pgfscope}%
\pgfsys@transformshift{0.622597in}{2.885369in}%
\pgfsys@useobject{currentmarker}{}%
\end{pgfscope}%
\end{pgfscope}%
\begin{pgfscope}%
\definecolor{textcolor}{rgb}{0.000000,0.000000,0.000000}%
\pgfsetstrokecolor{textcolor}%
\pgfsetfillcolor{textcolor}%
\pgftext[x=0.316851in, y=2.822055in, left, base]{\color{textcolor}\sffamily\fontsize{12.000000}{14.400000}\selectfont \(\displaystyle {0.7}\)}%
\end{pgfscope}%
\begin{pgfscope}%
\pgfsetbuttcap%
\pgfsetroundjoin%
\definecolor{currentfill}{rgb}{0.000000,0.000000,0.000000}%
\pgfsetfillcolor{currentfill}%
\pgfsetlinewidth{0.803000pt}%
\definecolor{currentstroke}{rgb}{0.000000,0.000000,0.000000}%
\pgfsetstrokecolor{currentstroke}%
\pgfsetdash{}{0pt}%
\pgfsys@defobject{currentmarker}{\pgfqpoint{-0.048611in}{0.000000in}}{\pgfqpoint{-0.000000in}{0.000000in}}{%
\pgfpathmoveto{\pgfqpoint{-0.000000in}{0.000000in}}%
\pgfpathlineto{\pgfqpoint{-0.048611in}{0.000000in}}%
\pgfusepath{stroke,fill}%
}%
\begin{pgfscope}%
\pgfsys@transformshift{0.622597in}{3.270369in}%
\pgfsys@useobject{currentmarker}{}%
\end{pgfscope}%
\end{pgfscope}%
\begin{pgfscope}%
\definecolor{textcolor}{rgb}{0.000000,0.000000,0.000000}%
\pgfsetstrokecolor{textcolor}%
\pgfsetfillcolor{textcolor}%
\pgftext[x=0.316851in, y=3.207055in, left, base]{\color{textcolor}\sffamily\fontsize{12.000000}{14.400000}\selectfont \(\displaystyle {0.8}\)}%
\end{pgfscope}%
\begin{pgfscope}%
\pgfsetbuttcap%
\pgfsetroundjoin%
\definecolor{currentfill}{rgb}{0.000000,0.000000,0.000000}%
\pgfsetfillcolor{currentfill}%
\pgfsetlinewidth{0.803000pt}%
\definecolor{currentstroke}{rgb}{0.000000,0.000000,0.000000}%
\pgfsetstrokecolor{currentstroke}%
\pgfsetdash{}{0pt}%
\pgfsys@defobject{currentmarker}{\pgfqpoint{-0.048611in}{0.000000in}}{\pgfqpoint{-0.000000in}{0.000000in}}{%
\pgfpathmoveto{\pgfqpoint{-0.000000in}{0.000000in}}%
\pgfpathlineto{\pgfqpoint{-0.048611in}{0.000000in}}%
\pgfusepath{stroke,fill}%
}%
\begin{pgfscope}%
\pgfsys@transformshift{0.622597in}{3.655369in}%
\pgfsys@useobject{currentmarker}{}%
\end{pgfscope}%
\end{pgfscope}%
\begin{pgfscope}%
\definecolor{textcolor}{rgb}{0.000000,0.000000,0.000000}%
\pgfsetstrokecolor{textcolor}%
\pgfsetfillcolor{textcolor}%
\pgftext[x=0.316851in, y=3.592055in, left, base]{\color{textcolor}\sffamily\fontsize{12.000000}{14.400000}\selectfont \(\displaystyle {0.9}\)}%
\end{pgfscope}%
\begin{pgfscope}%
\definecolor{textcolor}{rgb}{0.000000,0.000000,0.000000}%
\pgfsetstrokecolor{textcolor}%
\pgfsetfillcolor{textcolor}%
\pgftext[x=0.261295in,y=2.115369in,,bottom,rotate=90.000000]{\color{textcolor}\sffamily\fontsize{12.000000}{14.400000}\selectfont MAE}%
\end{pgfscope}%
\begin{pgfscope}%
\pgfpathrectangle{\pgfqpoint{0.622597in}{0.575369in}}{\pgfqpoint{6.200000in}{3.080000in}}%
\pgfusepath{clip}%
\pgfsetbuttcap%
\pgfsetroundjoin%
\pgfsetlinewidth{1.505625pt}%
\definecolor{currentstroke}{rgb}{1.000000,0.498039,0.054902}%
\pgfsetstrokecolor{currentstroke}%
\pgfsetdash{{5.550000pt}{2.400000pt}}{0.000000pt}%
\pgfpathmoveto{\pgfqpoint{0.904415in}{3.462869in}}%
\pgfpathlineto{\pgfqpoint{1.092294in}{2.346369in}}%
\pgfpathlineto{\pgfqpoint{1.374112in}{1.345369in}}%
\pgfpathlineto{\pgfqpoint{1.608961in}{1.845869in}}%
\pgfusepath{stroke}%
\end{pgfscope}%
\begin{pgfscope}%
\pgfpathrectangle{\pgfqpoint{0.622597in}{0.575369in}}{\pgfqpoint{6.200000in}{3.080000in}}%
\pgfusepath{clip}%
\pgfsetbuttcap%
\pgfsetroundjoin%
\definecolor{currentfill}{rgb}{1.000000,0.498039,0.054902}%
\pgfsetfillcolor{currentfill}%
\pgfsetlinewidth{1.003750pt}%
\definecolor{currentstroke}{rgb}{1.000000,0.498039,0.054902}%
\pgfsetstrokecolor{currentstroke}%
\pgfsetdash{}{0pt}%
\pgfsys@defobject{currentmarker}{\pgfqpoint{-0.041667in}{-0.041667in}}{\pgfqpoint{0.041667in}{0.041667in}}{%
\pgfpathmoveto{\pgfqpoint{0.000000in}{-0.041667in}}%
\pgfpathcurveto{\pgfqpoint{0.011050in}{-0.041667in}}{\pgfqpoint{0.021649in}{-0.037276in}}{\pgfqpoint{0.029463in}{-0.029463in}}%
\pgfpathcurveto{\pgfqpoint{0.037276in}{-0.021649in}}{\pgfqpoint{0.041667in}{-0.011050in}}{\pgfqpoint{0.041667in}{0.000000in}}%
\pgfpathcurveto{\pgfqpoint{0.041667in}{0.011050in}}{\pgfqpoint{0.037276in}{0.021649in}}{\pgfqpoint{0.029463in}{0.029463in}}%
\pgfpathcurveto{\pgfqpoint{0.021649in}{0.037276in}}{\pgfqpoint{0.011050in}{0.041667in}}{\pgfqpoint{0.000000in}{0.041667in}}%
\pgfpathcurveto{\pgfqpoint{-0.011050in}{0.041667in}}{\pgfqpoint{-0.021649in}{0.037276in}}{\pgfqpoint{-0.029463in}{0.029463in}}%
\pgfpathcurveto{\pgfqpoint{-0.037276in}{0.021649in}}{\pgfqpoint{-0.041667in}{0.011050in}}{\pgfqpoint{-0.041667in}{0.000000in}}%
\pgfpathcurveto{\pgfqpoint{-0.041667in}{-0.011050in}}{\pgfqpoint{-0.037276in}{-0.021649in}}{\pgfqpoint{-0.029463in}{-0.029463in}}%
\pgfpathcurveto{\pgfqpoint{-0.021649in}{-0.037276in}}{\pgfqpoint{-0.011050in}{-0.041667in}}{\pgfqpoint{0.000000in}{-0.041667in}}%
\pgfpathlineto{\pgfqpoint{0.000000in}{-0.041667in}}%
\pgfpathclose%
\pgfusepath{stroke,fill}%
}%
\begin{pgfscope}%
\pgfsys@transformshift{0.904415in}{3.462869in}%
\pgfsys@useobject{currentmarker}{}%
\end{pgfscope}%
\begin{pgfscope}%
\pgfsys@transformshift{1.092294in}{2.346369in}%
\pgfsys@useobject{currentmarker}{}%
\end{pgfscope}%
\begin{pgfscope}%
\pgfsys@transformshift{1.374112in}{1.345369in}%
\pgfsys@useobject{currentmarker}{}%
\end{pgfscope}%
\begin{pgfscope}%
\pgfsys@transformshift{1.608961in}{1.845869in}%
\pgfsys@useobject{currentmarker}{}%
\end{pgfscope}%
\end{pgfscope}%
\begin{pgfscope}%
\pgfpathrectangle{\pgfqpoint{0.622597in}{0.575369in}}{\pgfqpoint{6.200000in}{3.080000in}}%
\pgfusepath{clip}%
\pgfsetbuttcap%
\pgfsetroundjoin%
\pgfsetlinewidth{1.505625pt}%
\definecolor{currentstroke}{rgb}{0.172549,0.627451,0.172549}%
\pgfsetstrokecolor{currentstroke}%
\pgfsetdash{{5.550000pt}{2.400000pt}}{0.000000pt}%
\pgfpathmoveto{\pgfqpoint{0.904415in}{1.037369in}}%
\pgfpathlineto{\pgfqpoint{1.092294in}{0.998869in}}%
\pgfpathlineto{\pgfqpoint{1.374112in}{0.921869in}}%
\pgfpathlineto{\pgfqpoint{1.608961in}{0.883369in}}%
\pgfpathlineto{\pgfqpoint{2.031688in}{0.844869in}}%
\pgfpathlineto{\pgfqpoint{3.534718in}{0.806369in}}%
\pgfpathlineto{\pgfqpoint{6.540779in}{0.767869in}}%
\pgfusepath{stroke}%
\end{pgfscope}%
\begin{pgfscope}%
\pgfpathrectangle{\pgfqpoint{0.622597in}{0.575369in}}{\pgfqpoint{6.200000in}{3.080000in}}%
\pgfusepath{clip}%
\pgfsetbuttcap%
\pgfsetroundjoin%
\definecolor{currentfill}{rgb}{0.172549,0.627451,0.172549}%
\pgfsetfillcolor{currentfill}%
\pgfsetlinewidth{1.003750pt}%
\definecolor{currentstroke}{rgb}{0.172549,0.627451,0.172549}%
\pgfsetstrokecolor{currentstroke}%
\pgfsetdash{}{0pt}%
\pgfsys@defobject{currentmarker}{\pgfqpoint{-0.041667in}{-0.041667in}}{\pgfqpoint{0.041667in}{0.041667in}}{%
\pgfpathmoveto{\pgfqpoint{0.000000in}{-0.041667in}}%
\pgfpathcurveto{\pgfqpoint{0.011050in}{-0.041667in}}{\pgfqpoint{0.021649in}{-0.037276in}}{\pgfqpoint{0.029463in}{-0.029463in}}%
\pgfpathcurveto{\pgfqpoint{0.037276in}{-0.021649in}}{\pgfqpoint{0.041667in}{-0.011050in}}{\pgfqpoint{0.041667in}{0.000000in}}%
\pgfpathcurveto{\pgfqpoint{0.041667in}{0.011050in}}{\pgfqpoint{0.037276in}{0.021649in}}{\pgfqpoint{0.029463in}{0.029463in}}%
\pgfpathcurveto{\pgfqpoint{0.021649in}{0.037276in}}{\pgfqpoint{0.011050in}{0.041667in}}{\pgfqpoint{0.000000in}{0.041667in}}%
\pgfpathcurveto{\pgfqpoint{-0.011050in}{0.041667in}}{\pgfqpoint{-0.021649in}{0.037276in}}{\pgfqpoint{-0.029463in}{0.029463in}}%
\pgfpathcurveto{\pgfqpoint{-0.037276in}{0.021649in}}{\pgfqpoint{-0.041667in}{0.011050in}}{\pgfqpoint{-0.041667in}{0.000000in}}%
\pgfpathcurveto{\pgfqpoint{-0.041667in}{-0.011050in}}{\pgfqpoint{-0.037276in}{-0.021649in}}{\pgfqpoint{-0.029463in}{-0.029463in}}%
\pgfpathcurveto{\pgfqpoint{-0.021649in}{-0.037276in}}{\pgfqpoint{-0.011050in}{-0.041667in}}{\pgfqpoint{0.000000in}{-0.041667in}}%
\pgfpathlineto{\pgfqpoint{0.000000in}{-0.041667in}}%
\pgfpathclose%
\pgfusepath{stroke,fill}%
}%
\begin{pgfscope}%
\pgfsys@transformshift{0.904415in}{1.037369in}%
\pgfsys@useobject{currentmarker}{}%
\end{pgfscope}%
\begin{pgfscope}%
\pgfsys@transformshift{1.092294in}{0.998869in}%
\pgfsys@useobject{currentmarker}{}%
\end{pgfscope}%
\begin{pgfscope}%
\pgfsys@transformshift{1.374112in}{0.921869in}%
\pgfsys@useobject{currentmarker}{}%
\end{pgfscope}%
\begin{pgfscope}%
\pgfsys@transformshift{1.608961in}{0.883369in}%
\pgfsys@useobject{currentmarker}{}%
\end{pgfscope}%
\begin{pgfscope}%
\pgfsys@transformshift{2.031688in}{0.844869in}%
\pgfsys@useobject{currentmarker}{}%
\end{pgfscope}%
\begin{pgfscope}%
\pgfsys@transformshift{3.534718in}{0.806369in}%
\pgfsys@useobject{currentmarker}{}%
\end{pgfscope}%
\begin{pgfscope}%
\pgfsys@transformshift{6.540779in}{0.767869in}%
\pgfsys@useobject{currentmarker}{}%
\end{pgfscope}%
\end{pgfscope}%
\begin{pgfscope}%
\pgfsetrectcap%
\pgfsetmiterjoin%
\pgfsetlinewidth{0.803000pt}%
\definecolor{currentstroke}{rgb}{0.000000,0.000000,0.000000}%
\pgfsetstrokecolor{currentstroke}%
\pgfsetdash{}{0pt}%
\pgfpathmoveto{\pgfqpoint{0.622597in}{0.575369in}}%
\pgfpathlineto{\pgfqpoint{0.622597in}{3.655369in}}%
\pgfusepath{stroke}%
\end{pgfscope}%
\begin{pgfscope}%
\pgfsetrectcap%
\pgfsetmiterjoin%
\pgfsetlinewidth{0.803000pt}%
\definecolor{currentstroke}{rgb}{0.000000,0.000000,0.000000}%
\pgfsetstrokecolor{currentstroke}%
\pgfsetdash{}{0pt}%
\pgfpathmoveto{\pgfqpoint{6.822597in}{0.575369in}}%
\pgfpathlineto{\pgfqpoint{6.822597in}{3.655369in}}%
\pgfusepath{stroke}%
\end{pgfscope}%
\begin{pgfscope}%
\pgfsetrectcap%
\pgfsetmiterjoin%
\pgfsetlinewidth{0.803000pt}%
\definecolor{currentstroke}{rgb}{0.000000,0.000000,0.000000}%
\pgfsetstrokecolor{currentstroke}%
\pgfsetdash{}{0pt}%
\pgfpathmoveto{\pgfqpoint{0.622597in}{0.575369in}}%
\pgfpathlineto{\pgfqpoint{6.822597in}{0.575369in}}%
\pgfusepath{stroke}%
\end{pgfscope}%
\begin{pgfscope}%
\pgfsetrectcap%
\pgfsetmiterjoin%
\pgfsetlinewidth{0.803000pt}%
\definecolor{currentstroke}{rgb}{0.000000,0.000000,0.000000}%
\pgfsetstrokecolor{currentstroke}%
\pgfsetdash{}{0pt}%
\pgfpathmoveto{\pgfqpoint{0.622597in}{3.655369in}}%
\pgfpathlineto{\pgfqpoint{6.822597in}{3.655369in}}%
\pgfusepath{stroke}%
\end{pgfscope}%
\begin{pgfscope}%
\pgfsetbuttcap%
\pgfsetmiterjoin%
\definecolor{currentfill}{rgb}{1.000000,1.000000,1.000000}%
\pgfsetfillcolor{currentfill}%
\pgfsetfillopacity{0.800000}%
\pgfsetlinewidth{1.003750pt}%
\definecolor{currentstroke}{rgb}{0.800000,0.800000,0.800000}%
\pgfsetstrokecolor{currentstroke}%
\pgfsetstrokeopacity{0.800000}%
\pgfsetdash{}{0pt}%
\pgfpathmoveto{\pgfqpoint{5.149811in}{3.032778in}}%
\pgfpathlineto{\pgfqpoint{6.705931in}{3.032778in}}%
\pgfpathquadraticcurveto{\pgfqpoint{6.739264in}{3.032778in}}{\pgfqpoint{6.739264in}{3.066111in}}%
\pgfpathlineto{\pgfqpoint{6.739264in}{3.538702in}}%
\pgfpathquadraticcurveto{\pgfqpoint{6.739264in}{3.572035in}}{\pgfqpoint{6.705931in}{3.572035in}}%
\pgfpathlineto{\pgfqpoint{5.149811in}{3.572035in}}%
\pgfpathquadraticcurveto{\pgfqpoint{5.116477in}{3.572035in}}{\pgfqpoint{5.116477in}{3.538702in}}%
\pgfpathlineto{\pgfqpoint{5.116477in}{3.066111in}}%
\pgfpathquadraticcurveto{\pgfqpoint{5.116477in}{3.032778in}}{\pgfqpoint{5.149811in}{3.032778in}}%
\pgfpathlineto{\pgfqpoint{5.149811in}{3.032778in}}%
\pgfpathclose%
\pgfusepath{stroke,fill}%
\end{pgfscope}%
\begin{pgfscope}%
\pgfsetbuttcap%
\pgfsetroundjoin%
\pgfsetlinewidth{1.505625pt}%
\definecolor{currentstroke}{rgb}{1.000000,0.498039,0.054902}%
\pgfsetstrokecolor{currentstroke}%
\pgfsetdash{{5.550000pt}{2.400000pt}}{0.000000pt}%
\pgfpathmoveto{\pgfqpoint{5.183144in}{3.437074in}}%
\pgfpathlineto{\pgfqpoint{5.349811in}{3.437074in}}%
\pgfpathlineto{\pgfqpoint{5.516477in}{3.437074in}}%
\pgfusepath{stroke}%
\end{pgfscope}%
\begin{pgfscope}%
\pgfsetbuttcap%
\pgfsetroundjoin%
\definecolor{currentfill}{rgb}{1.000000,0.498039,0.054902}%
\pgfsetfillcolor{currentfill}%
\pgfsetlinewidth{1.003750pt}%
\definecolor{currentstroke}{rgb}{1.000000,0.498039,0.054902}%
\pgfsetstrokecolor{currentstroke}%
\pgfsetdash{}{0pt}%
\pgfsys@defobject{currentmarker}{\pgfqpoint{-0.041667in}{-0.041667in}}{\pgfqpoint{0.041667in}{0.041667in}}{%
\pgfpathmoveto{\pgfqpoint{0.000000in}{-0.041667in}}%
\pgfpathcurveto{\pgfqpoint{0.011050in}{-0.041667in}}{\pgfqpoint{0.021649in}{-0.037276in}}{\pgfqpoint{0.029463in}{-0.029463in}}%
\pgfpathcurveto{\pgfqpoint{0.037276in}{-0.021649in}}{\pgfqpoint{0.041667in}{-0.011050in}}{\pgfqpoint{0.041667in}{0.000000in}}%
\pgfpathcurveto{\pgfqpoint{0.041667in}{0.011050in}}{\pgfqpoint{0.037276in}{0.021649in}}{\pgfqpoint{0.029463in}{0.029463in}}%
\pgfpathcurveto{\pgfqpoint{0.021649in}{0.037276in}}{\pgfqpoint{0.011050in}{0.041667in}}{\pgfqpoint{0.000000in}{0.041667in}}%
\pgfpathcurveto{\pgfqpoint{-0.011050in}{0.041667in}}{\pgfqpoint{-0.021649in}{0.037276in}}{\pgfqpoint{-0.029463in}{0.029463in}}%
\pgfpathcurveto{\pgfqpoint{-0.037276in}{0.021649in}}{\pgfqpoint{-0.041667in}{0.011050in}}{\pgfqpoint{-0.041667in}{0.000000in}}%
\pgfpathcurveto{\pgfqpoint{-0.041667in}{-0.011050in}}{\pgfqpoint{-0.037276in}{-0.021649in}}{\pgfqpoint{-0.029463in}{-0.029463in}}%
\pgfpathcurveto{\pgfqpoint{-0.021649in}{-0.037276in}}{\pgfqpoint{-0.011050in}{-0.041667in}}{\pgfqpoint{0.000000in}{-0.041667in}}%
\pgfpathlineto{\pgfqpoint{0.000000in}{-0.041667in}}%
\pgfpathclose%
\pgfusepath{stroke,fill}%
}%
\begin{pgfscope}%
\pgfsys@transformshift{5.349811in}{3.437074in}%
\pgfsys@useobject{currentmarker}{}%
\end{pgfscope}%
\end{pgfscope}%
\begin{pgfscope}%
\definecolor{textcolor}{rgb}{0.000000,0.000000,0.000000}%
\pgfsetstrokecolor{textcolor}%
\pgfsetfillcolor{textcolor}%
\pgftext[x=5.649811in,y=3.378741in,left,base]{\color{textcolor}\sffamily\fontsize{12.000000}{14.400000}\selectfont RSD}%
\end{pgfscope}%
\begin{pgfscope}%
\pgfsetbuttcap%
\pgfsetroundjoin%
\pgfsetlinewidth{1.505625pt}%
\definecolor{currentstroke}{rgb}{0.172549,0.627451,0.172549}%
\pgfsetstrokecolor{currentstroke}%
\pgfsetdash{{5.550000pt}{2.400000pt}}{0.000000pt}%
\pgfpathmoveto{\pgfqpoint{5.183144in}{3.192446in}}%
\pgfpathlineto{\pgfqpoint{5.349811in}{3.192446in}}%
\pgfpathlineto{\pgfqpoint{5.516477in}{3.192446in}}%
\pgfusepath{stroke}%
\end{pgfscope}%
\begin{pgfscope}%
\pgfsetbuttcap%
\pgfsetroundjoin%
\definecolor{currentfill}{rgb}{0.172549,0.627451,0.172549}%
\pgfsetfillcolor{currentfill}%
\pgfsetlinewidth{1.003750pt}%
\definecolor{currentstroke}{rgb}{0.172549,0.627451,0.172549}%
\pgfsetstrokecolor{currentstroke}%
\pgfsetdash{}{0pt}%
\pgfsys@defobject{currentmarker}{\pgfqpoint{-0.041667in}{-0.041667in}}{\pgfqpoint{0.041667in}{0.041667in}}{%
\pgfpathmoveto{\pgfqpoint{0.000000in}{-0.041667in}}%
\pgfpathcurveto{\pgfqpoint{0.011050in}{-0.041667in}}{\pgfqpoint{0.021649in}{-0.037276in}}{\pgfqpoint{0.029463in}{-0.029463in}}%
\pgfpathcurveto{\pgfqpoint{0.037276in}{-0.021649in}}{\pgfqpoint{0.041667in}{-0.011050in}}{\pgfqpoint{0.041667in}{0.000000in}}%
\pgfpathcurveto{\pgfqpoint{0.041667in}{0.011050in}}{\pgfqpoint{0.037276in}{0.021649in}}{\pgfqpoint{0.029463in}{0.029463in}}%
\pgfpathcurveto{\pgfqpoint{0.021649in}{0.037276in}}{\pgfqpoint{0.011050in}{0.041667in}}{\pgfqpoint{0.000000in}{0.041667in}}%
\pgfpathcurveto{\pgfqpoint{-0.011050in}{0.041667in}}{\pgfqpoint{-0.021649in}{0.037276in}}{\pgfqpoint{-0.029463in}{0.029463in}}%
\pgfpathcurveto{\pgfqpoint{-0.037276in}{0.021649in}}{\pgfqpoint{-0.041667in}{0.011050in}}{\pgfqpoint{-0.041667in}{0.000000in}}%
\pgfpathcurveto{\pgfqpoint{-0.041667in}{-0.011050in}}{\pgfqpoint{-0.037276in}{-0.021649in}}{\pgfqpoint{-0.029463in}{-0.029463in}}%
\pgfpathcurveto{\pgfqpoint{-0.021649in}{-0.037276in}}{\pgfqpoint{-0.011050in}{-0.041667in}}{\pgfqpoint{0.000000in}{-0.041667in}}%
\pgfpathlineto{\pgfqpoint{0.000000in}{-0.041667in}}%
\pgfpathclose%
\pgfusepath{stroke,fill}%
}%
\begin{pgfscope}%
\pgfsys@transformshift{5.349811in}{3.192446in}%
\pgfsys@useobject{currentmarker}{}%
\end{pgfscope}%
\end{pgfscope}%
\begin{pgfscope}%
\definecolor{textcolor}{rgb}{0.000000,0.000000,0.000000}%
\pgfsetstrokecolor{textcolor}%
\pgfsetfillcolor{textcolor}%
\pgftext[x=5.649811in,y=3.134112in,left,base]{\color{textcolor}\sffamily\fontsize{12.000000}{14.400000}\selectfont DARE-GRAM}%
\end{pgfscope}%
\end{pgfpicture}%
\makeatother%
\endgroup%

%% file: images/ablation/sensitivity_s2n.pgf
\begingroup%
\makeatletter%
\begin{pgfpicture}%
\pgfpathrectangle{\pgfpointorigin}{\pgfqpoint{3.461917in}{3.085925in}}%
\pgfusepath{use as bounding box, clip}%
\begin{pgfscope}%
\pgfsetbuttcap%
\pgfsetmiterjoin%
\definecolor{currentfill}{rgb}{1.000000,1.000000,1.000000}%
\pgfsetfillcolor{currentfill}%
\pgfsetlinewidth{0.000000pt}%
\definecolor{currentstroke}{rgb}{1.000000,1.000000,1.000000}%
\pgfsetstrokecolor{currentstroke}%
\pgfsetdash{}{0pt}%
\pgfpathmoveto{\pgfqpoint{0.000000in}{0.000000in}}%
\pgfpathlineto{\pgfqpoint{3.461917in}{0.000000in}}%
\pgfpathlineto{\pgfqpoint{3.461917in}{3.085925in}}%
\pgfpathlineto{\pgfqpoint{0.000000in}{3.085925in}}%
\pgfpathlineto{\pgfqpoint{0.000000in}{0.000000in}}%
\pgfpathclose%
\pgfusepath{fill}%
\end{pgfscope}%
\begin{pgfscope}%
\pgfsetbuttcap%
\pgfsetmiterjoin%
\definecolor{currentfill}{rgb}{1.000000,1.000000,1.000000}%
\pgfsetfillcolor{currentfill}%
\pgfsetlinewidth{0.000000pt}%
\definecolor{currentstroke}{rgb}{0.000000,0.000000,0.000000}%
\pgfsetstrokecolor{currentstroke}%
\pgfsetstrokeopacity{0.000000}%
\pgfsetdash{}{0pt}%
\pgfpathmoveto{\pgfqpoint{0.100000in}{0.285925in}}%
\pgfpathlineto{\pgfqpoint{2.800000in}{0.285925in}}%
\pgfpathlineto{\pgfqpoint{2.800000in}{2.985925in}}%
\pgfpathlineto{\pgfqpoint{0.100000in}{2.985925in}}%
\pgfpathlineto{\pgfqpoint{0.100000in}{0.285925in}}%
\pgfpathclose%
\pgfusepath{fill}%
\end{pgfscope}%
\begin{pgfscope}%
\pgfsetbuttcap%
\pgfsetmiterjoin%
\definecolor{currentfill}{rgb}{0.950000,0.950000,0.950000}%
\pgfsetfillcolor{currentfill}%
\pgfsetfillopacity{0.500000}%
\pgfsetlinewidth{1.003750pt}%
\definecolor{currentstroke}{rgb}{0.950000,0.950000,0.950000}%
\pgfsetstrokecolor{currentstroke}%
\pgfsetstrokeopacity{0.500000}%
\pgfsetdash{}{0pt}%
\pgfpathmoveto{\pgfqpoint{0.303869in}{0.951659in}}%
\pgfpathlineto{\pgfqpoint{1.195497in}{1.699039in}}%
\pgfpathlineto{\pgfqpoint{1.183102in}{2.776895in}}%
\pgfpathlineto{\pgfqpoint{0.248805in}{2.095088in}}%
\pgfusepath{stroke,fill}%
\end{pgfscope}%
\begin{pgfscope}%
\pgfsetbuttcap%
\pgfsetmiterjoin%
\definecolor{currentfill}{rgb}{0.900000,0.900000,0.900000}%
\pgfsetfillcolor{currentfill}%
\pgfsetfillopacity{0.500000}%
\pgfsetlinewidth{1.003750pt}%
\definecolor{currentstroke}{rgb}{0.900000,0.900000,0.900000}%
\pgfsetstrokecolor{currentstroke}%
\pgfsetstrokeopacity{0.500000}%
\pgfsetdash{}{0pt}%
\pgfpathmoveto{\pgfqpoint{1.195497in}{1.699039in}}%
\pgfpathlineto{\pgfqpoint{2.626242in}{1.283178in}}%
\pgfpathlineto{\pgfqpoint{2.677300in}{2.398160in}}%
\pgfpathlineto{\pgfqpoint{1.183102in}{2.776895in}}%
\pgfusepath{stroke,fill}%
\end{pgfscope}%
\begin{pgfscope}%
\pgfsetbuttcap%
\pgfsetmiterjoin%
\definecolor{currentfill}{rgb}{0.925000,0.925000,0.925000}%
\pgfsetfillcolor{currentfill}%
\pgfsetfillopacity{0.500000}%
\pgfsetlinewidth{1.003750pt}%
\definecolor{currentstroke}{rgb}{0.925000,0.925000,0.925000}%
\pgfsetstrokecolor{currentstroke}%
\pgfsetstrokeopacity{0.500000}%
\pgfsetdash{}{0pt}%
\pgfpathmoveto{\pgfqpoint{0.303869in}{0.951659in}}%
\pgfpathlineto{\pgfqpoint{1.820528in}{0.456317in}}%
\pgfpathlineto{\pgfqpoint{2.626242in}{1.283178in}}%
\pgfpathlineto{\pgfqpoint{1.195497in}{1.699039in}}%
\pgfusepath{stroke,fill}%
\end{pgfscope}%
\begin{pgfscope}%
\pgfsetrectcap%
\pgfsetroundjoin%
\pgfsetlinewidth{0.803000pt}%
\definecolor{currentstroke}{rgb}{0.000000,0.000000,0.000000}%
\pgfsetstrokecolor{currentstroke}%
\pgfsetdash{}{0pt}%
\pgfpathmoveto{\pgfqpoint{0.303869in}{0.951659in}}%
\pgfpathlineto{\pgfqpoint{1.820528in}{0.456317in}}%
\pgfusepath{stroke}%
\end{pgfscope}%
\begin{pgfscope}%
\definecolor{textcolor}{rgb}{0.000000,0.000000,0.000000}%
\pgfsetstrokecolor{textcolor}%
\pgfsetfillcolor{textcolor}%
\pgftext[x=0.670960in, y=0.217875in, left, base,rotate=341.912962]{\color{textcolor}\sffamily\fontsize{12.000000}{14.400000}\selectfont \(\displaystyle \alpha_{cos}\)}%
\end{pgfscope}%
\begin{pgfscope}%
\pgfsetbuttcap%
\pgfsetroundjoin%
\pgfsetlinewidth{0.803000pt}%
\definecolor{currentstroke}{rgb}{0.690196,0.690196,0.690196}%
\pgfsetstrokecolor{currentstroke}%
\pgfsetdash{}{0pt}%
\pgfpathmoveto{\pgfqpoint{0.395725in}{0.921658in}}%
\pgfpathlineto{\pgfqpoint{1.282510in}{1.673748in}}%
\pgfpathlineto{\pgfqpoint{1.273795in}{2.753907in}}%
\pgfusepath{stroke}%
\end{pgfscope}%
\begin{pgfscope}%
\pgfsetbuttcap%
\pgfsetroundjoin%
\pgfsetlinewidth{0.803000pt}%
\definecolor{currentstroke}{rgb}{0.690196,0.690196,0.690196}%
\pgfsetstrokecolor{currentstroke}%
\pgfsetdash{}{0pt}%
\pgfpathmoveto{\pgfqpoint{0.772671in}{0.798548in}}%
\pgfpathlineto{\pgfqpoint{1.639093in}{1.570103in}}%
\pgfpathlineto{\pgfqpoint{1.645697in}{2.659641in}}%
\pgfusepath{stroke}%
\end{pgfscope}%
\begin{pgfscope}%
\pgfsetbuttcap%
\pgfsetroundjoin%
\pgfsetlinewidth{0.803000pt}%
\definecolor{currentstroke}{rgb}{0.690196,0.690196,0.690196}%
\pgfsetstrokecolor{currentstroke}%
\pgfsetdash{}{0pt}%
\pgfpathmoveto{\pgfqpoint{1.159868in}{0.672089in}}%
\pgfpathlineto{\pgfqpoint{2.004558in}{1.463877in}}%
\pgfpathlineto{\pgfqpoint{2.027270in}{2.562923in}}%
\pgfusepath{stroke}%
\end{pgfscope}%
\begin{pgfscope}%
\pgfsetbuttcap%
\pgfsetroundjoin%
\pgfsetlinewidth{0.803000pt}%
\definecolor{currentstroke}{rgb}{0.690196,0.690196,0.690196}%
\pgfsetstrokecolor{currentstroke}%
\pgfsetdash{}{0pt}%
\pgfpathmoveto{\pgfqpoint{1.557739in}{0.542144in}}%
\pgfpathlineto{\pgfqpoint{2.379243in}{1.354971in}}%
\pgfpathlineto{\pgfqpoint{2.418896in}{2.463657in}}%
\pgfusepath{stroke}%
\end{pgfscope}%
\begin{pgfscope}%
\pgfsetrectcap%
\pgfsetroundjoin%
\pgfsetlinewidth{0.803000pt}%
\definecolor{currentstroke}{rgb}{0.000000,0.000000,0.000000}%
\pgfsetstrokecolor{currentstroke}%
\pgfsetdash{}{0pt}%
\pgfpathmoveto{\pgfqpoint{0.403448in}{0.928207in}}%
\pgfpathlineto{\pgfqpoint{0.380248in}{0.908532in}}%
\pgfusepath{stroke}%
\end{pgfscope}%
\begin{pgfscope}%
\definecolor{textcolor}{rgb}{0.000000,0.000000,0.000000}%
\pgfsetstrokecolor{textcolor}%
\pgfsetfillcolor{textcolor}%
\pgftext[x=0.291012in,y=0.703154in,,top]{\color{textcolor}\sffamily\fontsize{10.000000}{12.000000}\selectfont \(\displaystyle {\ensuremath{-}2.0}\)}%
\end{pgfscope}%
\begin{pgfscope}%
\pgfsetrectcap%
\pgfsetroundjoin%
\pgfsetlinewidth{0.803000pt}%
\definecolor{currentstroke}{rgb}{0.000000,0.000000,0.000000}%
\pgfsetstrokecolor{currentstroke}%
\pgfsetdash{}{0pt}%
\pgfpathmoveto{\pgfqpoint{0.780224in}{0.805274in}}%
\pgfpathlineto{\pgfqpoint{0.757532in}{0.785066in}}%
\pgfusepath{stroke}%
\end{pgfscope}%
\begin{pgfscope}%
\definecolor{textcolor}{rgb}{0.000000,0.000000,0.000000}%
\pgfsetstrokecolor{textcolor}%
\pgfsetfillcolor{textcolor}%
\pgftext[x=0.668492in,y=0.576462in,,top]{\color{textcolor}\sffamily\fontsize{10.000000}{12.000000}\selectfont \(\displaystyle {\ensuremath{-}1.5}\)}%
\end{pgfscope}%
\begin{pgfscope}%
\pgfsetrectcap%
\pgfsetroundjoin%
\pgfsetlinewidth{0.803000pt}%
\definecolor{currentstroke}{rgb}{0.000000,0.000000,0.000000}%
\pgfsetstrokecolor{currentstroke}%
\pgfsetdash{}{0pt}%
\pgfpathmoveto{\pgfqpoint{1.167239in}{0.678999in}}%
\pgfpathlineto{\pgfqpoint{1.145092in}{0.658239in}}%
\pgfusepath{stroke}%
\end{pgfscope}%
\begin{pgfscope}%
\definecolor{textcolor}{rgb}{0.000000,0.000000,0.000000}%
\pgfsetstrokecolor{textcolor}%
\pgfsetfillcolor{textcolor}%
\pgftext[x=1.056291in,y=0.446305in,,top]{\color{textcolor}\sffamily\fontsize{10.000000}{12.000000}\selectfont \(\displaystyle {\ensuremath{-}1.0}\)}%
\end{pgfscope}%
\begin{pgfscope}%
\pgfsetrectcap%
\pgfsetroundjoin%
\pgfsetlinewidth{0.803000pt}%
\definecolor{currentstroke}{rgb}{0.000000,0.000000,0.000000}%
\pgfsetstrokecolor{currentstroke}%
\pgfsetdash{}{0pt}%
\pgfpathmoveto{\pgfqpoint{1.564917in}{0.549246in}}%
\pgfpathlineto{\pgfqpoint{1.543352in}{0.527909in}}%
\pgfusepath{stroke}%
\end{pgfscope}%
\begin{pgfscope}%
\definecolor{textcolor}{rgb}{0.000000,0.000000,0.000000}%
\pgfsetstrokecolor{textcolor}%
\pgfsetfillcolor{textcolor}%
\pgftext[x=1.454839in,y=0.312542in,,top]{\color{textcolor}\sffamily\fontsize{10.000000}{12.000000}\selectfont \(\displaystyle {\ensuremath{-}0.5}\)}%
\end{pgfscope}%
\begin{pgfscope}%
\pgfsetrectcap%
\pgfsetroundjoin%
\pgfsetlinewidth{0.803000pt}%
\definecolor{currentstroke}{rgb}{0.000000,0.000000,0.000000}%
\pgfsetstrokecolor{currentstroke}%
\pgfsetdash{}{0pt}%
\pgfpathmoveto{\pgfqpoint{2.626242in}{1.283178in}}%
\pgfpathlineto{\pgfqpoint{1.820528in}{0.456317in}}%
\pgfusepath{stroke}%
\end{pgfscope}%
\begin{pgfscope}%
\definecolor{textcolor}{rgb}{0.000000,0.000000,0.000000}%
\pgfsetstrokecolor{textcolor}%
\pgfsetfillcolor{textcolor}%
\pgftext[x=2.526509in, y=0.341123in, left, base,rotate=45.742112]{\color{textcolor}\sffamily\fontsize{12.000000}{14.400000}\selectfont \(\displaystyle \gamma_{scale}\)}%
\end{pgfscope}%
\begin{pgfscope}%
\pgfsetbuttcap%
\pgfsetroundjoin%
\pgfsetlinewidth{0.803000pt}%
\definecolor{currentstroke}{rgb}{0.690196,0.690196,0.690196}%
\pgfsetstrokecolor{currentstroke}%
\pgfsetdash{}{0pt}%
\pgfpathmoveto{\pgfqpoint{0.313414in}{2.142236in}}%
\pgfpathlineto{\pgfqpoint{0.365309in}{1.003159in}}%
\pgfpathlineto{\pgfqpoint{1.876277in}{0.513530in}}%
\pgfusepath{stroke}%
\end{pgfscope}%
\begin{pgfscope}%
\pgfsetbuttcap%
\pgfsetroundjoin%
\pgfsetlinewidth{0.803000pt}%
\definecolor{currentstroke}{rgb}{0.690196,0.690196,0.690196}%
\pgfsetstrokecolor{currentstroke}%
\pgfsetdash{}{0pt}%
\pgfpathmoveto{\pgfqpoint{0.488358in}{2.269902in}}%
\pgfpathlineto{\pgfqpoint{0.531836in}{1.142745in}}%
\pgfpathlineto{\pgfqpoint{2.027207in}{0.668421in}}%
\pgfusepath{stroke}%
\end{pgfscope}%
\begin{pgfscope}%
\pgfsetbuttcap%
\pgfsetroundjoin%
\pgfsetlinewidth{0.803000pt}%
\definecolor{currentstroke}{rgb}{0.690196,0.690196,0.690196}%
\pgfsetstrokecolor{currentstroke}%
\pgfsetdash{}{0pt}%
\pgfpathmoveto{\pgfqpoint{0.657775in}{2.393535in}}%
\pgfpathlineto{\pgfqpoint{0.693328in}{1.278111in}}%
\pgfpathlineto{\pgfqpoint{2.173336in}{0.818385in}}%
\pgfusepath{stroke}%
\end{pgfscope}%
\begin{pgfscope}%
\pgfsetbuttcap%
\pgfsetroundjoin%
\pgfsetlinewidth{0.803000pt}%
\definecolor{currentstroke}{rgb}{0.690196,0.690196,0.690196}%
\pgfsetstrokecolor{currentstroke}%
\pgfsetdash{}{0pt}%
\pgfpathmoveto{\pgfqpoint{0.821926in}{2.513325in}}%
\pgfpathlineto{\pgfqpoint{0.850012in}{1.409446in}}%
\pgfpathlineto{\pgfqpoint{2.314890in}{0.963655in}}%
\pgfusepath{stroke}%
\end{pgfscope}%
\begin{pgfscope}%
\pgfsetbuttcap%
\pgfsetroundjoin%
\pgfsetlinewidth{0.803000pt}%
\definecolor{currentstroke}{rgb}{0.690196,0.690196,0.690196}%
\pgfsetstrokecolor{currentstroke}%
\pgfsetdash{}{0pt}%
\pgfpathmoveto{\pgfqpoint{0.981050in}{2.629446in}}%
\pgfpathlineto{\pgfqpoint{1.002097in}{1.536928in}}%
\pgfpathlineto{\pgfqpoint{2.452081in}{1.104446in}}%
\pgfusepath{stroke}%
\end{pgfscope}%
\begin{pgfscope}%
\pgfsetbuttcap%
\pgfsetroundjoin%
\pgfsetlinewidth{0.803000pt}%
\definecolor{currentstroke}{rgb}{0.690196,0.690196,0.690196}%
\pgfsetstrokecolor{currentstroke}%
\pgfsetdash{}{0pt}%
\pgfpathmoveto{\pgfqpoint{1.135375in}{2.742066in}}%
\pgfpathlineto{\pgfqpoint{1.149785in}{1.660723in}}%
\pgfpathlineto{\pgfqpoint{2.585108in}{1.240964in}}%
\pgfusepath{stroke}%
\end{pgfscope}%
\begin{pgfscope}%
\pgfsetrectcap%
\pgfsetroundjoin%
\pgfsetlinewidth{0.803000pt}%
\definecolor{currentstroke}{rgb}{0.000000,0.000000,0.000000}%
\pgfsetstrokecolor{currentstroke}%
\pgfsetdash{}{0pt}%
\pgfpathmoveto{\pgfqpoint{1.863544in}{0.517656in}}%
\pgfpathlineto{\pgfqpoint{1.901776in}{0.505267in}}%
\pgfusepath{stroke}%
\end{pgfscope}%
\begin{pgfscope}%
\definecolor{textcolor}{rgb}{0.000000,0.000000,0.000000}%
\pgfsetstrokecolor{textcolor}%
\pgfsetfillcolor{textcolor}%
\pgftext[x=2.055215in,y=0.327054in,,top]{\color{textcolor}\sffamily\fontsize{10.000000}{12.000000}\selectfont \(\displaystyle {\ensuremath{-}5.0}\)}%
\end{pgfscope}%
\begin{pgfscope}%
\pgfsetrectcap%
\pgfsetroundjoin%
\pgfsetlinewidth{0.803000pt}%
\definecolor{currentstroke}{rgb}{0.000000,0.000000,0.000000}%
\pgfsetstrokecolor{currentstroke}%
\pgfsetdash{}{0pt}%
\pgfpathmoveto{\pgfqpoint{2.014616in}{0.672415in}}%
\pgfpathlineto{\pgfqpoint{2.052421in}{0.660423in}}%
\pgfusepath{stroke}%
\end{pgfscope}%
\begin{pgfscope}%
\definecolor{textcolor}{rgb}{0.000000,0.000000,0.000000}%
\pgfsetstrokecolor{textcolor}%
\pgfsetfillcolor{textcolor}%
\pgftext[x=2.203372in,y=0.485086in,,top]{\color{textcolor}\sffamily\fontsize{10.000000}{12.000000}\selectfont \(\displaystyle {\ensuremath{-}4.5}\)}%
\end{pgfscope}%
\begin{pgfscope}%
\pgfsetrectcap%
\pgfsetroundjoin%
\pgfsetlinewidth{0.803000pt}%
\definecolor{currentstroke}{rgb}{0.000000,0.000000,0.000000}%
\pgfsetstrokecolor{currentstroke}%
\pgfsetdash{}{0pt}%
\pgfpathmoveto{\pgfqpoint{2.160884in}{0.822253in}}%
\pgfpathlineto{\pgfqpoint{2.198271in}{0.810640in}}%
\pgfusepath{stroke}%
\end{pgfscope}%
\begin{pgfscope}%
\definecolor{textcolor}{rgb}{0.000000,0.000000,0.000000}%
\pgfsetstrokecolor{textcolor}%
\pgfsetfillcolor{textcolor}%
\pgftext[x=2.346814in,y=0.638087in,,top]{\color{textcolor}\sffamily\fontsize{10.000000}{12.000000}\selectfont \(\displaystyle {\ensuremath{-}4.0}\)}%
\end{pgfscope}%
\begin{pgfscope}%
\pgfsetrectcap%
\pgfsetroundjoin%
\pgfsetlinewidth{0.803000pt}%
\definecolor{currentstroke}{rgb}{0.000000,0.000000,0.000000}%
\pgfsetstrokecolor{currentstroke}%
\pgfsetdash{}{0pt}%
\pgfpathmoveto{\pgfqpoint{2.302575in}{0.967402in}}%
\pgfpathlineto{\pgfqpoint{2.339551in}{0.956150in}}%
\pgfusepath{stroke}%
\end{pgfscope}%
\begin{pgfscope}%
\definecolor{textcolor}{rgb}{0.000000,0.000000,0.000000}%
\pgfsetstrokecolor{textcolor}%
\pgfsetfillcolor{textcolor}%
\pgftext[x=2.485761in,y=0.786294in,,top]{\color{textcolor}\sffamily\fontsize{10.000000}{12.000000}\selectfont \(\displaystyle {\ensuremath{-}3.5}\)}%
\end{pgfscope}%
\begin{pgfscope}%
\pgfsetrectcap%
\pgfsetroundjoin%
\pgfsetlinewidth{0.803000pt}%
\definecolor{currentstroke}{rgb}{0.000000,0.000000,0.000000}%
\pgfsetstrokecolor{currentstroke}%
\pgfsetdash{}{0pt}%
\pgfpathmoveto{\pgfqpoint{2.439900in}{1.108079in}}%
\pgfpathlineto{\pgfqpoint{2.476473in}{1.097171in}}%
\pgfusepath{stroke}%
\end{pgfscope}%
\begin{pgfscope}%
\definecolor{textcolor}{rgb}{0.000000,0.000000,0.000000}%
\pgfsetstrokecolor{textcolor}%
\pgfsetfillcolor{textcolor}%
\pgftext[x=2.620422in,y=0.929930in,,top]{\color{textcolor}\sffamily\fontsize{10.000000}{12.000000}\selectfont \(\displaystyle {\ensuremath{-}3.0}\)}%
\end{pgfscope}%
\begin{pgfscope}%
\pgfsetrectcap%
\pgfsetroundjoin%
\pgfsetlinewidth{0.803000pt}%
\definecolor{currentstroke}{rgb}{0.000000,0.000000,0.000000}%
\pgfsetstrokecolor{currentstroke}%
\pgfsetdash{}{0pt}%
\pgfpathmoveto{\pgfqpoint{2.573059in}{1.244488in}}%
\pgfpathlineto{\pgfqpoint{2.609234in}{1.233908in}}%
\pgfusepath{stroke}%
\end{pgfscope}%
\begin{pgfscope}%
\definecolor{textcolor}{rgb}{0.000000,0.000000,0.000000}%
\pgfsetstrokecolor{textcolor}%
\pgfsetfillcolor{textcolor}%
\pgftext[x=2.750991in,y=1.069202in,,top]{\color{textcolor}\sffamily\fontsize{10.000000}{12.000000}\selectfont \(\displaystyle {\ensuremath{-}2.5}\)}%
\end{pgfscope}%
\begin{pgfscope}%
\pgfsetrectcap%
\pgfsetroundjoin%
\pgfsetlinewidth{0.803000pt}%
\definecolor{currentstroke}{rgb}{0.000000,0.000000,0.000000}%
\pgfsetstrokecolor{currentstroke}%
\pgfsetdash{}{0pt}%
\pgfpathmoveto{\pgfqpoint{2.626242in}{1.283178in}}%
\pgfpathlineto{\pgfqpoint{2.677300in}{2.398160in}}%
\pgfusepath{stroke}%
\end{pgfscope}%
\begin{pgfscope}%
\definecolor{textcolor}{rgb}{0.000000,0.000000,0.000000}%
\pgfsetstrokecolor{textcolor}%
\pgfsetfillcolor{textcolor}%
\pgftext[x=3.210617in,y=1.903401in,,]{\color{textcolor}\sffamily\fontsize{10.000000}{12.000000}\selectfont MAE}%
\end{pgfscope}%
\begin{pgfscope}%
\pgfsetbuttcap%
\pgfsetroundjoin%
\pgfsetlinewidth{0.803000pt}%
\definecolor{currentstroke}{rgb}{0.690196,0.690196,0.690196}%
\pgfsetstrokecolor{currentstroke}%
\pgfsetdash{}{0pt}%
\pgfpathmoveto{\pgfqpoint{2.627220in}{1.304540in}}%
\pgfpathlineto{\pgfqpoint{1.195259in}{1.719732in}}%
\pgfpathlineto{\pgfqpoint{0.302815in}{0.973529in}}%
\pgfusepath{stroke}%
\end{pgfscope}%
\begin{pgfscope}%
\pgfsetbuttcap%
\pgfsetroundjoin%
\pgfsetlinewidth{0.803000pt}%
\definecolor{currentstroke}{rgb}{0.690196,0.690196,0.690196}%
\pgfsetstrokecolor{currentstroke}%
\pgfsetdash{}{0pt}%
\pgfpathmoveto{\pgfqpoint{2.651212in}{1.828455in}}%
\pgfpathlineto{\pgfqpoint{1.189429in}{2.226729in}}%
\pgfpathlineto{\pgfqpoint{0.276963in}{1.510366in}}%
\pgfusepath{stroke}%
\end{pgfscope}%
\begin{pgfscope}%
\pgfsetbuttcap%
\pgfsetroundjoin%
\pgfsetlinewidth{0.803000pt}%
\definecolor{currentstroke}{rgb}{0.690196,0.690196,0.690196}%
\pgfsetstrokecolor{currentstroke}%
\pgfsetdash{}{0pt}%
\pgfpathmoveto{\pgfqpoint{2.676234in}{2.374882in}}%
\pgfpathlineto{\pgfqpoint{1.183361in}{2.754439in}}%
\pgfpathlineto{\pgfqpoint{0.249957in}{2.071176in}}%
\pgfusepath{stroke}%
\end{pgfscope}%
\begin{pgfscope}%
\pgfsetrectcap%
\pgfsetroundjoin%
\pgfsetlinewidth{0.803000pt}%
\definecolor{currentstroke}{rgb}{0.000000,0.000000,0.000000}%
\pgfsetstrokecolor{currentstroke}%
\pgfsetdash{}{0pt}%
\pgfpathmoveto{\pgfqpoint{2.615202in}{1.308024in}}%
\pgfpathlineto{\pgfqpoint{2.651286in}{1.297562in}}%
\pgfusepath{stroke}%
\end{pgfscope}%
\begin{pgfscope}%
\definecolor{textcolor}{rgb}{0.000000,0.000000,0.000000}%
\pgfsetstrokecolor{textcolor}%
\pgfsetfillcolor{textcolor}%
\pgftext[x=2.880606in,y=1.340623in,,top]{\color{textcolor}\sffamily\fontsize{10.000000}{12.000000}\selectfont \(\displaystyle {0.0}\)}%
\end{pgfscope}%
\begin{pgfscope}%
\pgfsetrectcap%
\pgfsetroundjoin%
\pgfsetlinewidth{0.803000pt}%
\definecolor{currentstroke}{rgb}{0.000000,0.000000,0.000000}%
\pgfsetstrokecolor{currentstroke}%
\pgfsetdash{}{0pt}%
\pgfpathmoveto{\pgfqpoint{2.638931in}{1.831801in}}%
\pgfpathlineto{\pgfqpoint{2.675803in}{1.821755in}}%
\pgfusepath{stroke}%
\end{pgfscope}%
\begin{pgfscope}%
\definecolor{textcolor}{rgb}{0.000000,0.000000,0.000000}%
\pgfsetstrokecolor{textcolor}%
\pgfsetfillcolor{textcolor}%
\pgftext[x=2.909804in,y=1.863100in,,top]{\color{textcolor}\sffamily\fontsize{10.000000}{12.000000}\selectfont \(\displaystyle {0.1}\)}%
\end{pgfscope}%
\begin{pgfscope}%
\pgfsetrectcap%
\pgfsetroundjoin%
\pgfsetlinewidth{0.803000pt}%
\definecolor{currentstroke}{rgb}{0.000000,0.000000,0.000000}%
\pgfsetstrokecolor{currentstroke}%
\pgfsetdash{}{0pt}%
\pgfpathmoveto{\pgfqpoint{2.663679in}{2.378074in}}%
\pgfpathlineto{\pgfqpoint{2.701375in}{2.368490in}}%
\pgfusepath{stroke}%
\end{pgfscope}%
\begin{pgfscope}%
\definecolor{textcolor}{rgb}{0.000000,0.000000,0.000000}%
\pgfsetstrokecolor{textcolor}%
\pgfsetfillcolor{textcolor}%
\pgftext[x=2.940251in,y=2.407930in,,top]{\color{textcolor}\sffamily\fontsize{10.000000}{12.000000}\selectfont \(\displaystyle {0.2}\)}%
\end{pgfscope}%
\begin{pgfscope}%
\pgfpathrectangle{\pgfqpoint{0.100000in}{0.285925in}}{\pgfqpoint{2.700000in}{2.700000in}}%
\pgfusepath{clip}%
\pgfsetbuttcap%
\pgfsetroundjoin%
\definecolor{currentfill}{rgb}{0.458674,0.816363,0.329727}%
\pgfsetfillcolor{currentfill}%
\pgfsetlinewidth{0.000000pt}%
\definecolor{currentstroke}{rgb}{0.000000,0.000000,0.000000}%
\pgfsetstrokecolor{currentstroke}%
\pgfsetdash{}{0pt}%
\pgfpathmoveto{\pgfqpoint{1.224889in}{2.175878in}}%
\pgfpathlineto{\pgfqpoint{1.082823in}{1.978301in}}%
\pgfpathlineto{\pgfqpoint{1.600004in}{1.896108in}}%
\pgfpathlineto{\pgfqpoint{1.224889in}{2.175878in}}%
\pgfpathclose%
\pgfusepath{fill}%
\end{pgfscope}%
\begin{pgfscope}%
\pgfpathrectangle{\pgfqpoint{0.100000in}{0.285925in}}{\pgfqpoint{2.700000in}{2.700000in}}%
\pgfusepath{clip}%
\pgfsetbuttcap%
\pgfsetroundjoin%
\definecolor{currentfill}{rgb}{0.259857,0.745492,0.444467}%
\pgfsetfillcolor{currentfill}%
\pgfsetlinewidth{0.000000pt}%
\definecolor{currentstroke}{rgb}{0.000000,0.000000,0.000000}%
\pgfsetstrokecolor{currentstroke}%
\pgfsetdash{}{0pt}%
\pgfpathmoveto{\pgfqpoint{1.600004in}{1.896108in}}%
\pgfpathlineto{\pgfqpoint{1.738515in}{2.029921in}}%
\pgfpathlineto{\pgfqpoint{1.224889in}{2.175878in}}%
\pgfpathlineto{\pgfqpoint{1.600004in}{1.896108in}}%
\pgfpathclose%
\pgfusepath{fill}%
\end{pgfscope}%
\begin{pgfscope}%
\pgfpathrectangle{\pgfqpoint{0.100000in}{0.285925in}}{\pgfqpoint{2.700000in}{2.700000in}}%
\pgfusepath{clip}%
\pgfsetbuttcap%
\pgfsetroundjoin%
\definecolor{currentfill}{rgb}{0.220124,0.725509,0.466226}%
\pgfsetfillcolor{currentfill}%
\pgfsetlinewidth{0.000000pt}%
\definecolor{currentstroke}{rgb}{0.000000,0.000000,0.000000}%
\pgfsetstrokecolor{currentstroke}%
\pgfsetdash{}{0pt}%
\pgfpathmoveto{\pgfqpoint{1.738515in}{2.029921in}}%
\pgfpathlineto{\pgfqpoint{1.600004in}{1.896108in}}%
\pgfpathlineto{\pgfqpoint{1.965856in}{1.994265in}}%
\pgfpathlineto{\pgfqpoint{1.738515in}{2.029921in}}%
\pgfpathclose%
\pgfusepath{fill}%
\end{pgfscope}%
\begin{pgfscope}%
\pgfpathrectangle{\pgfqpoint{0.100000in}{0.285925in}}{\pgfqpoint{2.700000in}{2.700000in}}%
\pgfusepath{clip}%
\pgfsetbuttcap%
\pgfsetroundjoin%
\definecolor{currentfill}{rgb}{0.886271,0.892374,0.095374}%
\pgfsetfillcolor{currentfill}%
\pgfsetlinewidth{0.000000pt}%
\definecolor{currentstroke}{rgb}{0.000000,0.000000,0.000000}%
\pgfsetstrokecolor{currentstroke}%
\pgfsetdash{}{0pt}%
\pgfpathmoveto{\pgfqpoint{1.443802in}{1.682409in}}%
\pgfpathlineto{\pgfqpoint{1.082823in}{1.978301in}}%
\pgfpathlineto{\pgfqpoint{0.920970in}{1.870121in}}%
\pgfpathlineto{\pgfqpoint{1.443802in}{1.682409in}}%
\pgfpathclose%
\pgfusepath{fill}%
\end{pgfscope}%
\begin{pgfscope}%
\pgfpathrectangle{\pgfqpoint{0.100000in}{0.285925in}}{\pgfqpoint{2.700000in}{2.700000in}}%
\pgfusepath{clip}%
\pgfsetbuttcap%
\pgfsetroundjoin%
\definecolor{currentfill}{rgb}{0.762373,0.876424,0.137064}%
\pgfsetfillcolor{currentfill}%
\pgfsetlinewidth{0.000000pt}%
\definecolor{currentstroke}{rgb}{0.000000,0.000000,0.000000}%
\pgfsetstrokecolor{currentstroke}%
\pgfsetdash{}{0pt}%
\pgfpathmoveto{\pgfqpoint{1.443802in}{1.682409in}}%
\pgfpathlineto{\pgfqpoint{1.600004in}{1.896108in}}%
\pgfpathlineto{\pgfqpoint{1.082823in}{1.978301in}}%
\pgfpathlineto{\pgfqpoint{1.443802in}{1.682409in}}%
\pgfpathclose%
\pgfusepath{fill}%
\end{pgfscope}%
\begin{pgfscope}%
\pgfpathrectangle{\pgfqpoint{0.100000in}{0.285925in}}{\pgfqpoint{2.700000in}{2.700000in}}%
\pgfusepath{clip}%
\pgfsetbuttcap%
\pgfsetroundjoin%
\definecolor{currentfill}{rgb}{0.296479,0.761561,0.424223}%
\pgfsetfillcolor{currentfill}%
\pgfsetlinewidth{0.000000pt}%
\definecolor{currentstroke}{rgb}{0.000000,0.000000,0.000000}%
\pgfsetstrokecolor{currentstroke}%
\pgfsetdash{}{0pt}%
\pgfpathmoveto{\pgfqpoint{1.600004in}{1.896108in}}%
\pgfpathlineto{\pgfqpoint{1.828674in}{1.810663in}}%
\pgfpathlineto{\pgfqpoint{1.965856in}{1.994265in}}%
\pgfpathlineto{\pgfqpoint{1.600004in}{1.896108in}}%
\pgfpathclose%
\pgfusepath{fill}%
\end{pgfscope}%
\begin{pgfscope}%
\pgfpathrectangle{\pgfqpoint{0.100000in}{0.285925in}}{\pgfqpoint{2.700000in}{2.700000in}}%
\pgfusepath{clip}%
\pgfsetbuttcap%
\pgfsetroundjoin%
\definecolor{currentfill}{rgb}{0.344074,0.780029,0.397381}%
\pgfsetfillcolor{currentfill}%
\pgfsetlinewidth{0.000000pt}%
\definecolor{currentstroke}{rgb}{0.000000,0.000000,0.000000}%
\pgfsetstrokecolor{currentstroke}%
\pgfsetdash{}{0pt}%
\pgfpathmoveto{\pgfqpoint{1.965856in}{1.994265in}}%
\pgfpathlineto{\pgfqpoint{2.197156in}{1.653149in}}%
\pgfpathlineto{\pgfqpoint{2.332683in}{1.883264in}}%
\pgfpathlineto{\pgfqpoint{1.965856in}{1.994265in}}%
\pgfpathclose%
\pgfusepath{fill}%
\end{pgfscope}%
\begin{pgfscope}%
\pgfpathrectangle{\pgfqpoint{0.100000in}{0.285925in}}{\pgfqpoint{2.700000in}{2.700000in}}%
\pgfusepath{clip}%
\pgfsetbuttcap%
\pgfsetroundjoin%
\definecolor{currentfill}{rgb}{0.487026,0.823929,0.312321}%
\pgfsetfillcolor{currentfill}%
\pgfsetlinewidth{0.000000pt}%
\definecolor{currentstroke}{rgb}{0.000000,0.000000,0.000000}%
\pgfsetstrokecolor{currentstroke}%
\pgfsetdash{}{0pt}%
\pgfpathmoveto{\pgfqpoint{1.828674in}{1.810663in}}%
\pgfpathlineto{\pgfqpoint{2.197156in}{1.653149in}}%
\pgfpathlineto{\pgfqpoint{1.965856in}{1.994265in}}%
\pgfpathlineto{\pgfqpoint{1.828674in}{1.810663in}}%
\pgfpathclose%
\pgfusepath{fill}%
\end{pgfscope}%
\begin{pgfscope}%
\pgfpathrectangle{\pgfqpoint{0.100000in}{0.285925in}}{\pgfqpoint{2.700000in}{2.700000in}}%
\pgfusepath{clip}%
\pgfsetbuttcap%
\pgfsetroundjoin%
\definecolor{currentfill}{rgb}{0.626579,0.854645,0.223353}%
\pgfsetfillcolor{currentfill}%
\pgfsetlinewidth{0.000000pt}%
\definecolor{currentstroke}{rgb}{0.000000,0.000000,0.000000}%
\pgfsetstrokecolor{currentstroke}%
\pgfsetdash{}{0pt}%
\pgfpathmoveto{\pgfqpoint{1.600004in}{1.896108in}}%
\pgfpathlineto{\pgfqpoint{1.443802in}{1.682409in}}%
\pgfpathlineto{\pgfqpoint{1.828674in}{1.810663in}}%
\pgfpathlineto{\pgfqpoint{1.600004in}{1.896108in}}%
\pgfpathclose%
\pgfusepath{fill}%
\end{pgfscope}%
\begin{pgfscope}%
\pgfpathrectangle{\pgfqpoint{0.100000in}{0.285925in}}{\pgfqpoint{2.700000in}{2.700000in}}%
\pgfusepath{clip}%
\pgfsetbuttcap%
\pgfsetroundjoin%
\definecolor{currentfill}{rgb}{0.845561,0.887322,0.099702}%
\pgfsetfillcolor{currentfill}%
\pgfsetlinewidth{0.000000pt}%
\definecolor{currentstroke}{rgb}{0.000000,0.000000,0.000000}%
\pgfsetstrokecolor{currentstroke}%
\pgfsetdash{}{0pt}%
\pgfpathmoveto{\pgfqpoint{1.443802in}{1.682409in}}%
\pgfpathlineto{\pgfqpoint{0.920970in}{1.870121in}}%
\pgfpathlineto{\pgfqpoint{0.769174in}{1.743335in}}%
\pgfpathlineto{\pgfqpoint{1.443802in}{1.682409in}}%
\pgfpathclose%
\pgfusepath{fill}%
\end{pgfscope}%
\begin{pgfscope}%
\pgfpathrectangle{\pgfqpoint{0.100000in}{0.285925in}}{\pgfqpoint{2.700000in}{2.700000in}}%
\pgfusepath{clip}%
\pgfsetbuttcap%
\pgfsetroundjoin%
\definecolor{currentfill}{rgb}{0.246070,0.738910,0.452024}%
\pgfsetfillcolor{currentfill}%
\pgfsetlinewidth{0.000000pt}%
\definecolor{currentstroke}{rgb}{0.000000,0.000000,0.000000}%
\pgfsetstrokecolor{currentstroke}%
\pgfsetdash{}{0pt}%
\pgfpathmoveto{\pgfqpoint{2.508608in}{1.889540in}}%
\pgfpathlineto{\pgfqpoint{2.332683in}{1.883264in}}%
\pgfpathlineto{\pgfqpoint{2.197156in}{1.653149in}}%
\pgfpathlineto{\pgfqpoint{2.508608in}{1.889540in}}%
\pgfpathclose%
\pgfusepath{fill}%
\end{pgfscope}%
\begin{pgfscope}%
\pgfpathrectangle{\pgfqpoint{0.100000in}{0.285925in}}{\pgfqpoint{2.700000in}{2.700000in}}%
\pgfusepath{clip}%
\pgfsetbuttcap%
\pgfsetroundjoin%
\definecolor{currentfill}{rgb}{0.906311,0.894855,0.098125}%
\pgfsetfillcolor{currentfill}%
\pgfsetlinewidth{0.000000pt}%
\definecolor{currentstroke}{rgb}{0.000000,0.000000,0.000000}%
\pgfsetstrokecolor{currentstroke}%
\pgfsetdash{}{0pt}%
\pgfpathmoveto{\pgfqpoint{1.828674in}{1.810663in}}%
\pgfpathlineto{\pgfqpoint{1.443802in}{1.682409in}}%
\pgfpathlineto{\pgfqpoint{1.674452in}{1.605055in}}%
\pgfpathlineto{\pgfqpoint{1.828674in}{1.810663in}}%
\pgfpathclose%
\pgfusepath{fill}%
\end{pgfscope}%
\begin{pgfscope}%
\pgfpathrectangle{\pgfqpoint{0.100000in}{0.285925in}}{\pgfqpoint{2.700000in}{2.700000in}}%
\pgfusepath{clip}%
\pgfsetbuttcap%
\pgfsetroundjoin%
\definecolor{currentfill}{rgb}{0.886271,0.892374,0.095374}%
\pgfsetfillcolor{currentfill}%
\pgfsetlinewidth{0.000000pt}%
\definecolor{currentstroke}{rgb}{0.000000,0.000000,0.000000}%
\pgfsetstrokecolor{currentstroke}%
\pgfsetdash{}{0pt}%
\pgfpathmoveto{\pgfqpoint{2.197156in}{1.653149in}}%
\pgfpathlineto{\pgfqpoint{1.828674in}{1.810663in}}%
\pgfpathlineto{\pgfqpoint{1.674452in}{1.605055in}}%
\pgfpathlineto{\pgfqpoint{2.197156in}{1.653149in}}%
\pgfpathclose%
\pgfusepath{fill}%
\end{pgfscope}%
\begin{pgfscope}%
\pgfpathrectangle{\pgfqpoint{0.100000in}{0.285925in}}{\pgfqpoint{2.700000in}{2.700000in}}%
\pgfusepath{clip}%
\pgfsetbuttcap%
\pgfsetroundjoin%
\definecolor{currentfill}{rgb}{0.964894,0.902323,0.123941}%
\pgfsetfillcolor{currentfill}%
\pgfsetlinewidth{0.000000pt}%
\definecolor{currentstroke}{rgb}{0.000000,0.000000,0.000000}%
\pgfsetstrokecolor{currentstroke}%
\pgfsetdash{}{0pt}%
\pgfpathmoveto{\pgfqpoint{0.769174in}{1.743335in}}%
\pgfpathlineto{\pgfqpoint{1.297158in}{1.550691in}}%
\pgfpathlineto{\pgfqpoint{1.443802in}{1.682409in}}%
\pgfpathlineto{\pgfqpoint{0.769174in}{1.743335in}}%
\pgfpathclose%
\pgfusepath{fill}%
\end{pgfscope}%
\begin{pgfscope}%
\pgfpathrectangle{\pgfqpoint{0.100000in}{0.285925in}}{\pgfqpoint{2.700000in}{2.700000in}}%
\pgfusepath{clip}%
\pgfsetbuttcap%
\pgfsetroundjoin%
\definecolor{currentfill}{rgb}{0.412913,0.803041,0.357269}%
\pgfsetfillcolor{currentfill}%
\pgfsetlinewidth{0.000000pt}%
\definecolor{currentstroke}{rgb}{0.000000,0.000000,0.000000}%
\pgfsetstrokecolor{currentstroke}%
\pgfsetdash{}{0pt}%
\pgfpathmoveto{\pgfqpoint{2.197156in}{1.653149in}}%
\pgfpathlineto{\pgfqpoint{2.373310in}{1.641462in}}%
\pgfpathlineto{\pgfqpoint{2.508608in}{1.889540in}}%
\pgfpathlineto{\pgfqpoint{2.197156in}{1.653149in}}%
\pgfpathclose%
\pgfusepath{fill}%
\end{pgfscope}%
\begin{pgfscope}%
\pgfpathrectangle{\pgfqpoint{0.100000in}{0.285925in}}{\pgfqpoint{2.700000in}{2.700000in}}%
\pgfusepath{clip}%
\pgfsetbuttcap%
\pgfsetroundjoin%
\definecolor{currentfill}{rgb}{0.993248,0.906157,0.143936}%
\pgfsetfillcolor{currentfill}%
\pgfsetlinewidth{0.000000pt}%
\definecolor{currentstroke}{rgb}{0.000000,0.000000,0.000000}%
\pgfsetstrokecolor{currentstroke}%
\pgfsetdash{}{0pt}%
\pgfpathmoveto{\pgfqpoint{1.443802in}{1.682409in}}%
\pgfpathlineto{\pgfqpoint{1.530207in}{1.514568in}}%
\pgfpathlineto{\pgfqpoint{1.674452in}{1.605055in}}%
\pgfpathlineto{\pgfqpoint{1.443802in}{1.682409in}}%
\pgfpathclose%
\pgfusepath{fill}%
\end{pgfscope}%
\begin{pgfscope}%
\pgfpathrectangle{\pgfqpoint{0.100000in}{0.285925in}}{\pgfqpoint{2.700000in}{2.700000in}}%
\pgfusepath{clip}%
\pgfsetbuttcap%
\pgfsetroundjoin%
\definecolor{currentfill}{rgb}{0.983868,0.904867,0.136897}%
\pgfsetfillcolor{currentfill}%
\pgfsetlinewidth{0.000000pt}%
\definecolor{currentstroke}{rgb}{0.000000,0.000000,0.000000}%
\pgfsetstrokecolor{currentstroke}%
\pgfsetdash{}{0pt}%
\pgfpathmoveto{\pgfqpoint{1.297158in}{1.550691in}}%
\pgfpathlineto{\pgfqpoint{1.530207in}{1.514568in}}%
\pgfpathlineto{\pgfqpoint{1.443802in}{1.682409in}}%
\pgfpathlineto{\pgfqpoint{1.297158in}{1.550691in}}%
\pgfpathclose%
\pgfusepath{fill}%
\end{pgfscope}%
\begin{pgfscope}%
\pgfpathrectangle{\pgfqpoint{0.100000in}{0.285925in}}{\pgfqpoint{2.700000in}{2.700000in}}%
\pgfusepath{clip}%
\pgfsetbuttcap%
\pgfsetroundjoin%
\definecolor{currentfill}{rgb}{0.855810,0.888601,0.097452}%
\pgfsetfillcolor{currentfill}%
\pgfsetlinewidth{0.000000pt}%
\definecolor{currentstroke}{rgb}{0.000000,0.000000,0.000000}%
\pgfsetstrokecolor{currentstroke}%
\pgfsetdash{}{0pt}%
\pgfpathmoveto{\pgfqpoint{0.769174in}{1.743335in}}%
\pgfpathlineto{\pgfqpoint{0.435541in}{1.482141in}}%
\pgfpathlineto{\pgfqpoint{1.297158in}{1.550691in}}%
\pgfpathlineto{\pgfqpoint{0.769174in}{1.743335in}}%
\pgfpathclose%
\pgfusepath{fill}%
\end{pgfscope}%
\begin{pgfscope}%
\pgfpathrectangle{\pgfqpoint{0.100000in}{0.285925in}}{\pgfqpoint{2.700000in}{2.700000in}}%
\pgfusepath{clip}%
\pgfsetbuttcap%
\pgfsetroundjoin%
\definecolor{currentfill}{rgb}{0.130067,0.651384,0.521608}%
\pgfsetfillcolor{currentfill}%
\pgfsetlinewidth{0.000000pt}%
\definecolor{currentstroke}{rgb}{0.000000,0.000000,0.000000}%
\pgfsetstrokecolor{currentstroke}%
\pgfsetdash{}{0pt}%
\pgfpathmoveto{\pgfqpoint{1.674452in}{1.605055in}}%
\pgfpathlineto{\pgfqpoint{2.235996in}{1.843149in}}%
\pgfpathlineto{\pgfqpoint{2.197156in}{1.653149in}}%
\pgfpathlineto{\pgfqpoint{1.674452in}{1.605055in}}%
\pgfpathclose%
\pgfusepath{fill}%
\end{pgfscope}%
\begin{pgfscope}%
\pgfpathrectangle{\pgfqpoint{0.100000in}{0.285925in}}{\pgfqpoint{2.700000in}{2.700000in}}%
\pgfusepath{clip}%
\pgfsetbuttcap%
\pgfsetroundjoin%
\definecolor{currentfill}{rgb}{0.121148,0.592739,0.544641}%
\pgfsetfillcolor{currentfill}%
\pgfsetlinewidth{0.000000pt}%
\definecolor{currentstroke}{rgb}{0.000000,0.000000,0.000000}%
\pgfsetstrokecolor{currentstroke}%
\pgfsetdash{}{0pt}%
\pgfpathmoveto{\pgfqpoint{2.235996in}{1.843149in}}%
\pgfpathlineto{\pgfqpoint{2.373310in}{1.641462in}}%
\pgfpathlineto{\pgfqpoint{2.197156in}{1.653149in}}%
\pgfpathlineto{\pgfqpoint{2.235996in}{1.843149in}}%
\pgfpathclose%
\pgfusepath{fill}%
\end{pgfscope}%
\begin{pgfscope}%
\pgfpathrectangle{\pgfqpoint{0.100000in}{0.285925in}}{\pgfqpoint{2.700000in}{2.700000in}}%
\pgfusepath{clip}%
\pgfsetbuttcap%
\pgfsetroundjoin%
\definecolor{currentfill}{rgb}{0.124780,0.640461,0.527068}%
\pgfsetfillcolor{currentfill}%
\pgfsetlinewidth{0.000000pt}%
\definecolor{currentstroke}{rgb}{0.000000,0.000000,0.000000}%
\pgfsetstrokecolor{currentstroke}%
\pgfsetdash{}{0pt}%
\pgfpathmoveto{\pgfqpoint{1.674452in}{1.605055in}}%
\pgfpathlineto{\pgfqpoint{1.530207in}{1.514568in}}%
\pgfpathlineto{\pgfqpoint{1.913651in}{1.753680in}}%
\pgfpathlineto{\pgfqpoint{1.674452in}{1.605055in}}%
\pgfpathclose%
\pgfusepath{fill}%
\end{pgfscope}%
\begin{pgfscope}%
\pgfpathrectangle{\pgfqpoint{0.100000in}{0.285925in}}{\pgfqpoint{2.700000in}{2.700000in}}%
\pgfusepath{clip}%
\pgfsetbuttcap%
\pgfsetroundjoin%
\definecolor{currentfill}{rgb}{0.906311,0.894855,0.098125}%
\pgfsetfillcolor{currentfill}%
\pgfsetlinewidth{0.000000pt}%
\definecolor{currentstroke}{rgb}{0.000000,0.000000,0.000000}%
\pgfsetstrokecolor{currentstroke}%
\pgfsetdash{}{0pt}%
\pgfpathmoveto{\pgfqpoint{0.435541in}{1.482141in}}%
\pgfpathlineto{\pgfqpoint{0.974592in}{1.295139in}}%
\pgfpathlineto{\pgfqpoint{1.297158in}{1.550691in}}%
\pgfpathlineto{\pgfqpoint{0.435541in}{1.482141in}}%
\pgfpathclose%
\pgfusepath{fill}%
\end{pgfscope}%
\begin{pgfscope}%
\pgfpathrectangle{\pgfqpoint{0.100000in}{0.285925in}}{\pgfqpoint{2.700000in}{2.700000in}}%
\pgfusepath{clip}%
\pgfsetbuttcap%
\pgfsetroundjoin%
\definecolor{currentfill}{rgb}{0.263663,0.237631,0.518762}%
\pgfsetfillcolor{currentfill}%
\pgfsetlinewidth{0.000000pt}%
\definecolor{currentstroke}{rgb}{0.000000,0.000000,0.000000}%
\pgfsetstrokecolor{currentstroke}%
\pgfsetdash{}{0pt}%
\pgfpathmoveto{\pgfqpoint{1.913651in}{1.753680in}}%
\pgfpathlineto{\pgfqpoint{2.235996in}{1.843149in}}%
\pgfpathlineto{\pgfqpoint{1.674452in}{1.605055in}}%
\pgfpathlineto{\pgfqpoint{1.913651in}{1.753680in}}%
\pgfpathclose%
\pgfusepath{fill}%
\end{pgfscope}%
\begin{pgfscope}%
\pgfpathrectangle{\pgfqpoint{0.100000in}{0.285925in}}{\pgfqpoint{2.700000in}{2.700000in}}%
\pgfusepath{clip}%
\pgfsetbuttcap%
\pgfsetroundjoin%
\definecolor{currentfill}{rgb}{0.487026,0.823929,0.312321}%
\pgfsetfillcolor{currentfill}%
\pgfsetlinewidth{0.000000pt}%
\definecolor{currentstroke}{rgb}{0.000000,0.000000,0.000000}%
\pgfsetstrokecolor{currentstroke}%
\pgfsetdash{}{0pt}%
\pgfpathmoveto{\pgfqpoint{1.211027in}{1.378207in}}%
\pgfpathlineto{\pgfqpoint{1.530207in}{1.514568in}}%
\pgfpathlineto{\pgfqpoint{1.297158in}{1.550691in}}%
\pgfpathlineto{\pgfqpoint{1.211027in}{1.378207in}}%
\pgfpathclose%
\pgfusepath{fill}%
\end{pgfscope}%
\begin{pgfscope}%
\pgfpathrectangle{\pgfqpoint{0.100000in}{0.285925in}}{\pgfqpoint{2.700000in}{2.700000in}}%
\pgfusepath{clip}%
\pgfsetbuttcap%
\pgfsetroundjoin%
\definecolor{currentfill}{rgb}{0.515992,0.831158,0.294279}%
\pgfsetfillcolor{currentfill}%
\pgfsetlinewidth{0.000000pt}%
\definecolor{currentstroke}{rgb}{0.000000,0.000000,0.000000}%
\pgfsetstrokecolor{currentstroke}%
\pgfsetdash{}{0pt}%
\pgfpathmoveto{\pgfqpoint{1.211027in}{1.378207in}}%
\pgfpathlineto{\pgfqpoint{1.297158in}{1.550691in}}%
\pgfpathlineto{\pgfqpoint{0.974592in}{1.295139in}}%
\pgfpathlineto{\pgfqpoint{1.211027in}{1.378207in}}%
\pgfpathclose%
\pgfusepath{fill}%
\end{pgfscope}%
\begin{pgfscope}%
\pgfpathrectangle{\pgfqpoint{0.100000in}{0.285925in}}{\pgfqpoint{2.700000in}{2.700000in}}%
\pgfusepath{clip}%
\pgfsetbuttcap%
\pgfsetroundjoin%
\definecolor{currentfill}{rgb}{0.278012,0.180367,0.486697}%
\pgfsetfillcolor{currentfill}%
\pgfsetlinewidth{0.000000pt}%
\definecolor{currentstroke}{rgb}{0.000000,0.000000,0.000000}%
\pgfsetstrokecolor{currentstroke}%
\pgfsetdash{}{0pt}%
\pgfpathmoveto{\pgfqpoint{2.086779in}{1.356675in}}%
\pgfpathlineto{\pgfqpoint{2.235996in}{1.843149in}}%
\pgfpathlineto{\pgfqpoint{1.913651in}{1.753680in}}%
\pgfpathlineto{\pgfqpoint{2.086779in}{1.356675in}}%
\pgfpathclose%
\pgfusepath{fill}%
\end{pgfscope}%
\begin{pgfscope}%
\pgfpathrectangle{\pgfqpoint{0.100000in}{0.285925in}}{\pgfqpoint{2.700000in}{2.700000in}}%
\pgfusepath{clip}%
\pgfsetbuttcap%
\pgfsetroundjoin%
\definecolor{currentfill}{rgb}{0.172719,0.448791,0.557885}%
\pgfsetfillcolor{currentfill}%
\pgfsetlinewidth{0.000000pt}%
\definecolor{currentstroke}{rgb}{0.000000,0.000000,0.000000}%
\pgfsetstrokecolor{currentstroke}%
\pgfsetdash{}{0pt}%
\pgfpathmoveto{\pgfqpoint{1.211027in}{1.378207in}}%
\pgfpathlineto{\pgfqpoint{1.913651in}{1.753680in}}%
\pgfpathlineto{\pgfqpoint{1.530207in}{1.514568in}}%
\pgfpathlineto{\pgfqpoint{1.211027in}{1.378207in}}%
\pgfpathclose%
\pgfusepath{fill}%
\end{pgfscope}%
\begin{pgfscope}%
\pgfpathrectangle{\pgfqpoint{0.100000in}{0.285925in}}{\pgfqpoint{2.700000in}{2.700000in}}%
\pgfusepath{clip}%
\pgfsetbuttcap%
\pgfsetroundjoin%
\definecolor{currentfill}{rgb}{0.271828,0.209303,0.504434}%
\pgfsetfillcolor{currentfill}%
\pgfsetlinewidth{0.000000pt}%
\definecolor{currentstroke}{rgb}{0.000000,0.000000,0.000000}%
\pgfsetstrokecolor{currentstroke}%
\pgfsetdash{}{0pt}%
\pgfpathmoveto{\pgfqpoint{1.913651in}{1.753680in}}%
\pgfpathlineto{\pgfqpoint{1.600049in}{1.460897in}}%
\pgfpathlineto{\pgfqpoint{2.086779in}{1.356675in}}%
\pgfpathlineto{\pgfqpoint{1.913651in}{1.753680in}}%
\pgfpathclose%
\pgfusepath{fill}%
\end{pgfscope}%
\begin{pgfscope}%
\pgfpathrectangle{\pgfqpoint{0.100000in}{0.285925in}}{\pgfqpoint{2.700000in}{2.700000in}}%
\pgfusepath{clip}%
\pgfsetbuttcap%
\pgfsetroundjoin%
\definecolor{currentfill}{rgb}{0.267004,0.004874,0.329415}%
\pgfsetfillcolor{currentfill}%
\pgfsetlinewidth{0.000000pt}%
\definecolor{currentstroke}{rgb}{0.000000,0.000000,0.000000}%
\pgfsetstrokecolor{currentstroke}%
\pgfsetdash{}{0pt}%
\pgfpathmoveto{\pgfqpoint{1.600049in}{1.460897in}}%
\pgfpathlineto{\pgfqpoint{1.913651in}{1.753680in}}%
\pgfpathlineto{\pgfqpoint{1.211027in}{1.378207in}}%
\pgfpathlineto{\pgfqpoint{1.600049in}{1.460897in}}%
\pgfpathclose%
\pgfusepath{fill}%
\end{pgfscope}%
\end{pgfpicture}%
\makeatother%
\endgroup%

%% file: images/ablation/thresold_s2n.pgf
\begingroup%
\makeatletter%
\begin{pgfpicture}%
\pgfpathrectangle{\pgfpointorigin}{\pgfqpoint{2.908450in}{2.900000in}}%
\pgfusepath{use as bounding box, clip}%
\begin{pgfscope}%
\pgfsetbuttcap%
\pgfsetmiterjoin%
\definecolor{currentfill}{rgb}{1.000000,1.000000,1.000000}%
\pgfsetfillcolor{currentfill}%
\pgfsetlinewidth{0.000000pt}%
\definecolor{currentstroke}{rgb}{1.000000,1.000000,1.000000}%
\pgfsetstrokecolor{currentstroke}%
\pgfsetdash{}{0pt}%
\pgfpathmoveto{\pgfqpoint{0.000000in}{0.000000in}}%
\pgfpathlineto{\pgfqpoint{2.908450in}{0.000000in}}%
\pgfpathlineto{\pgfqpoint{2.908450in}{2.900000in}}%
\pgfpathlineto{\pgfqpoint{0.000000in}{2.900000in}}%
\pgfpathlineto{\pgfqpoint{0.000000in}{0.000000in}}%
\pgfpathclose%
\pgfusepath{fill}%
\end{pgfscope}%
\begin{pgfscope}%
\pgfsetbuttcap%
\pgfsetmiterjoin%
\definecolor{currentfill}{rgb}{1.000000,1.000000,1.000000}%
\pgfsetfillcolor{currentfill}%
\pgfsetlinewidth{0.000000pt}%
\definecolor{currentstroke}{rgb}{0.000000,0.000000,0.000000}%
\pgfsetstrokecolor{currentstroke}%
\pgfsetstrokeopacity{0.000000}%
\pgfsetdash{}{0pt}%
\pgfpathmoveto{\pgfqpoint{0.634105in}{0.521603in}}%
\pgfpathlineto{\pgfqpoint{2.726152in}{0.521603in}}%
\pgfpathlineto{\pgfqpoint{2.726152in}{2.800000in}}%
\pgfpathlineto{\pgfqpoint{0.634105in}{2.800000in}}%
\pgfpathlineto{\pgfqpoint{0.634105in}{0.521603in}}%
\pgfpathclose%
\pgfusepath{fill}%
\end{pgfscope}%
\begin{pgfscope}%
\pgfpathrectangle{\pgfqpoint{0.634105in}{0.521603in}}{\pgfqpoint{2.092048in}{2.278397in}}%
\pgfusepath{clip}%
\pgfsetrectcap%
\pgfsetroundjoin%
\pgfsetlinewidth{0.803000pt}%
\definecolor{currentstroke}{rgb}{0.690196,0.690196,0.690196}%
\pgfsetstrokecolor{currentstroke}%
\pgfsetdash{}{0pt}%
\pgfpathmoveto{\pgfqpoint{0.729198in}{0.521603in}}%
\pgfpathlineto{\pgfqpoint{0.729198in}{2.800000in}}%
\pgfusepath{stroke}%
\end{pgfscope}%
\begin{pgfscope}%
\pgfsetbuttcap%
\pgfsetroundjoin%
\definecolor{currentfill}{rgb}{0.000000,0.000000,0.000000}%
\pgfsetfillcolor{currentfill}%
\pgfsetlinewidth{0.803000pt}%
\definecolor{currentstroke}{rgb}{0.000000,0.000000,0.000000}%
\pgfsetstrokecolor{currentstroke}%
\pgfsetdash{}{0pt}%
\pgfsys@defobject{currentmarker}{\pgfqpoint{0.000000in}{-0.048611in}}{\pgfqpoint{0.000000in}{0.000000in}}{%
\pgfpathmoveto{\pgfqpoint{0.000000in}{0.000000in}}%
\pgfpathlineto{\pgfqpoint{0.000000in}{-0.048611in}}%
\pgfusepath{stroke,fill}%
}%
\begin{pgfscope}%
\pgfsys@transformshift{0.729198in}{0.521603in}%
\pgfsys@useobject{currentmarker}{}%
\end{pgfscope}%
\end{pgfscope}%
\begin{pgfscope}%
\definecolor{textcolor}{rgb}{0.000000,0.000000,0.000000}%
\pgfsetstrokecolor{textcolor}%
\pgfsetfillcolor{textcolor}%
\pgftext[x=0.729198in,y=0.424381in,,top]{\color{textcolor}\sffamily\fontsize{10.000000}{12.000000}\selectfont \(\displaystyle {0.900}\)}%
\end{pgfscope}%
\begin{pgfscope}%
\pgfpathrectangle{\pgfqpoint{0.634105in}{0.521603in}}{\pgfqpoint{2.092048in}{2.278397in}}%
\pgfusepath{clip}%
\pgfsetrectcap%
\pgfsetroundjoin%
\pgfsetlinewidth{0.803000pt}%
\definecolor{currentstroke}{rgb}{0.690196,0.690196,0.690196}%
\pgfsetstrokecolor{currentstroke}%
\pgfsetdash{}{0pt}%
\pgfpathmoveto{\pgfqpoint{1.209466in}{0.521603in}}%
\pgfpathlineto{\pgfqpoint{1.209466in}{2.800000in}}%
\pgfusepath{stroke}%
\end{pgfscope}%
\begin{pgfscope}%
\pgfsetbuttcap%
\pgfsetroundjoin%
\definecolor{currentfill}{rgb}{0.000000,0.000000,0.000000}%
\pgfsetfillcolor{currentfill}%
\pgfsetlinewidth{0.803000pt}%
\definecolor{currentstroke}{rgb}{0.000000,0.000000,0.000000}%
\pgfsetstrokecolor{currentstroke}%
\pgfsetdash{}{0pt}%
\pgfsys@defobject{currentmarker}{\pgfqpoint{0.000000in}{-0.048611in}}{\pgfqpoint{0.000000in}{0.000000in}}{%
\pgfpathmoveto{\pgfqpoint{0.000000in}{0.000000in}}%
\pgfpathlineto{\pgfqpoint{0.000000in}{-0.048611in}}%
\pgfusepath{stroke,fill}%
}%
\begin{pgfscope}%
\pgfsys@transformshift{1.209466in}{0.521603in}%
\pgfsys@useobject{currentmarker}{}%
\end{pgfscope}%
\end{pgfscope}%
\begin{pgfscope}%
\definecolor{textcolor}{rgb}{0.000000,0.000000,0.000000}%
\pgfsetstrokecolor{textcolor}%
\pgfsetfillcolor{textcolor}%
\pgftext[x=1.209466in,y=0.424381in,,top]{\color{textcolor}\sffamily\fontsize{10.000000}{12.000000}\selectfont \(\displaystyle {0.925}\)}%
\end{pgfscope}%
\begin{pgfscope}%
\pgfpathrectangle{\pgfqpoint{0.634105in}{0.521603in}}{\pgfqpoint{2.092048in}{2.278397in}}%
\pgfusepath{clip}%
\pgfsetrectcap%
\pgfsetroundjoin%
\pgfsetlinewidth{0.803000pt}%
\definecolor{currentstroke}{rgb}{0.690196,0.690196,0.690196}%
\pgfsetstrokecolor{currentstroke}%
\pgfsetdash{}{0pt}%
\pgfpathmoveto{\pgfqpoint{1.689734in}{0.521603in}}%
\pgfpathlineto{\pgfqpoint{1.689734in}{2.800000in}}%
\pgfusepath{stroke}%
\end{pgfscope}%
\begin{pgfscope}%
\pgfsetbuttcap%
\pgfsetroundjoin%
\definecolor{currentfill}{rgb}{0.000000,0.000000,0.000000}%
\pgfsetfillcolor{currentfill}%
\pgfsetlinewidth{0.803000pt}%
\definecolor{currentstroke}{rgb}{0.000000,0.000000,0.000000}%
\pgfsetstrokecolor{currentstroke}%
\pgfsetdash{}{0pt}%
\pgfsys@defobject{currentmarker}{\pgfqpoint{0.000000in}{-0.048611in}}{\pgfqpoint{0.000000in}{0.000000in}}{%
\pgfpathmoveto{\pgfqpoint{0.000000in}{0.000000in}}%
\pgfpathlineto{\pgfqpoint{0.000000in}{-0.048611in}}%
\pgfusepath{stroke,fill}%
}%
\begin{pgfscope}%
\pgfsys@transformshift{1.689734in}{0.521603in}%
\pgfsys@useobject{currentmarker}{}%
\end{pgfscope}%
\end{pgfscope}%
\begin{pgfscope}%
\definecolor{textcolor}{rgb}{0.000000,0.000000,0.000000}%
\pgfsetstrokecolor{textcolor}%
\pgfsetfillcolor{textcolor}%
\pgftext[x=1.689734in,y=0.424381in,,top]{\color{textcolor}\sffamily\fontsize{10.000000}{12.000000}\selectfont \(\displaystyle {0.950}\)}%
\end{pgfscope}%
\begin{pgfscope}%
\pgfpathrectangle{\pgfqpoint{0.634105in}{0.521603in}}{\pgfqpoint{2.092048in}{2.278397in}}%
\pgfusepath{clip}%
\pgfsetrectcap%
\pgfsetroundjoin%
\pgfsetlinewidth{0.803000pt}%
\definecolor{currentstroke}{rgb}{0.690196,0.690196,0.690196}%
\pgfsetstrokecolor{currentstroke}%
\pgfsetdash{}{0pt}%
\pgfpathmoveto{\pgfqpoint{2.170002in}{0.521603in}}%
\pgfpathlineto{\pgfqpoint{2.170002in}{2.800000in}}%
\pgfusepath{stroke}%
\end{pgfscope}%
\begin{pgfscope}%
\pgfsetbuttcap%
\pgfsetroundjoin%
\definecolor{currentfill}{rgb}{0.000000,0.000000,0.000000}%
\pgfsetfillcolor{currentfill}%
\pgfsetlinewidth{0.803000pt}%
\definecolor{currentstroke}{rgb}{0.000000,0.000000,0.000000}%
\pgfsetstrokecolor{currentstroke}%
\pgfsetdash{}{0pt}%
\pgfsys@defobject{currentmarker}{\pgfqpoint{0.000000in}{-0.048611in}}{\pgfqpoint{0.000000in}{0.000000in}}{%
\pgfpathmoveto{\pgfqpoint{0.000000in}{0.000000in}}%
\pgfpathlineto{\pgfqpoint{0.000000in}{-0.048611in}}%
\pgfusepath{stroke,fill}%
}%
\begin{pgfscope}%
\pgfsys@transformshift{2.170002in}{0.521603in}%
\pgfsys@useobject{currentmarker}{}%
\end{pgfscope}%
\end{pgfscope}%
\begin{pgfscope}%
\definecolor{textcolor}{rgb}{0.000000,0.000000,0.000000}%
\pgfsetstrokecolor{textcolor}%
\pgfsetfillcolor{textcolor}%
\pgftext[x=2.170002in,y=0.424381in,,top]{\color{textcolor}\sffamily\fontsize{10.000000}{12.000000}\selectfont \(\displaystyle {0.975}\)}%
\end{pgfscope}%
\begin{pgfscope}%
\pgfpathrectangle{\pgfqpoint{0.634105in}{0.521603in}}{\pgfqpoint{2.092048in}{2.278397in}}%
\pgfusepath{clip}%
\pgfsetrectcap%
\pgfsetroundjoin%
\pgfsetlinewidth{0.803000pt}%
\definecolor{currentstroke}{rgb}{0.690196,0.690196,0.690196}%
\pgfsetstrokecolor{currentstroke}%
\pgfsetdash{}{0pt}%
\pgfpathmoveto{\pgfqpoint{2.650270in}{0.521603in}}%
\pgfpathlineto{\pgfqpoint{2.650270in}{2.800000in}}%
\pgfusepath{stroke}%
\end{pgfscope}%
\begin{pgfscope}%
\pgfsetbuttcap%
\pgfsetroundjoin%
\definecolor{currentfill}{rgb}{0.000000,0.000000,0.000000}%
\pgfsetfillcolor{currentfill}%
\pgfsetlinewidth{0.803000pt}%
\definecolor{currentstroke}{rgb}{0.000000,0.000000,0.000000}%
\pgfsetstrokecolor{currentstroke}%
\pgfsetdash{}{0pt}%
\pgfsys@defobject{currentmarker}{\pgfqpoint{0.000000in}{-0.048611in}}{\pgfqpoint{0.000000in}{0.000000in}}{%
\pgfpathmoveto{\pgfqpoint{0.000000in}{0.000000in}}%
\pgfpathlineto{\pgfqpoint{0.000000in}{-0.048611in}}%
\pgfusepath{stroke,fill}%
}%
\begin{pgfscope}%
\pgfsys@transformshift{2.650270in}{0.521603in}%
\pgfsys@useobject{currentmarker}{}%
\end{pgfscope}%
\end{pgfscope}%
\begin{pgfscope}%
\definecolor{textcolor}{rgb}{0.000000,0.000000,0.000000}%
\pgfsetstrokecolor{textcolor}%
\pgfsetfillcolor{textcolor}%
\pgftext[x=2.650270in,y=0.424381in,,top]{\color{textcolor}\sffamily\fontsize{10.000000}{12.000000}\selectfont \(\displaystyle {1.000}\)}%
\end{pgfscope}%
\begin{pgfscope}%
\definecolor{textcolor}{rgb}{0.000000,0.000000,0.000000}%
\pgfsetstrokecolor{textcolor}%
\pgfsetfillcolor{textcolor}%
\pgftext[x=1.680129in,y=0.234413in,,top]{\color{textcolor}\sffamily\fontsize{10.000000}{12.000000}\selectfont Threshold}%
\end{pgfscope}%
\begin{pgfscope}%
\pgfpathrectangle{\pgfqpoint{0.634105in}{0.521603in}}{\pgfqpoint{2.092048in}{2.278397in}}%
\pgfusepath{clip}%
\pgfsetrectcap%
\pgfsetroundjoin%
\pgfsetlinewidth{0.803000pt}%
\definecolor{currentstroke}{rgb}{0.690196,0.690196,0.690196}%
\pgfsetstrokecolor{currentstroke}%
\pgfsetdash{}{0pt}%
\pgfpathmoveto{\pgfqpoint{0.634105in}{0.521603in}}%
\pgfpathlineto{\pgfqpoint{2.726152in}{0.521603in}}%
\pgfusepath{stroke}%
\end{pgfscope}%
\begin{pgfscope}%
\pgfsetbuttcap%
\pgfsetroundjoin%
\definecolor{currentfill}{rgb}{0.000000,0.000000,0.000000}%
\pgfsetfillcolor{currentfill}%
\pgfsetlinewidth{0.803000pt}%
\definecolor{currentstroke}{rgb}{0.000000,0.000000,0.000000}%
\pgfsetstrokecolor{currentstroke}%
\pgfsetdash{}{0pt}%
\pgfsys@defobject{currentmarker}{\pgfqpoint{-0.048611in}{0.000000in}}{\pgfqpoint{-0.000000in}{0.000000in}}{%
\pgfpathmoveto{\pgfqpoint{-0.000000in}{0.000000in}}%
\pgfpathlineto{\pgfqpoint{-0.048611in}{0.000000in}}%
\pgfusepath{stroke,fill}%
}%
\begin{pgfscope}%
\pgfsys@transformshift{0.634105in}{0.521603in}%
\pgfsys@useobject{currentmarker}{}%
\end{pgfscope}%
\end{pgfscope}%
\begin{pgfscope}%
\definecolor{textcolor}{rgb}{0.000000,0.000000,0.000000}%
\pgfsetstrokecolor{textcolor}%
\pgfsetfillcolor{textcolor}%
\pgftext[x=0.289968in, y=0.468842in, left, base]{\color{textcolor}\sffamily\fontsize{10.000000}{12.000000}\selectfont \(\displaystyle {0.00}\)}%
\end{pgfscope}%
\begin{pgfscope}%
\pgfpathrectangle{\pgfqpoint{0.634105in}{0.521603in}}{\pgfqpoint{2.092048in}{2.278397in}}%
\pgfusepath{clip}%
\pgfsetrectcap%
\pgfsetroundjoin%
\pgfsetlinewidth{0.803000pt}%
\definecolor{currentstroke}{rgb}{0.690196,0.690196,0.690196}%
\pgfsetstrokecolor{currentstroke}%
\pgfsetdash{}{0pt}%
\pgfpathmoveto{\pgfqpoint{0.634105in}{0.825390in}}%
\pgfpathlineto{\pgfqpoint{2.726152in}{0.825390in}}%
\pgfusepath{stroke}%
\end{pgfscope}%
\begin{pgfscope}%
\pgfsetbuttcap%
\pgfsetroundjoin%
\definecolor{currentfill}{rgb}{0.000000,0.000000,0.000000}%
\pgfsetfillcolor{currentfill}%
\pgfsetlinewidth{0.803000pt}%
\definecolor{currentstroke}{rgb}{0.000000,0.000000,0.000000}%
\pgfsetstrokecolor{currentstroke}%
\pgfsetdash{}{0pt}%
\pgfsys@defobject{currentmarker}{\pgfqpoint{-0.048611in}{0.000000in}}{\pgfqpoint{-0.000000in}{0.000000in}}{%
\pgfpathmoveto{\pgfqpoint{-0.000000in}{0.000000in}}%
\pgfpathlineto{\pgfqpoint{-0.048611in}{0.000000in}}%
\pgfusepath{stroke,fill}%
}%
\begin{pgfscope}%
\pgfsys@transformshift{0.634105in}{0.825390in}%
\pgfsys@useobject{currentmarker}{}%
\end{pgfscope}%
\end{pgfscope}%
\begin{pgfscope}%
\definecolor{textcolor}{rgb}{0.000000,0.000000,0.000000}%
\pgfsetstrokecolor{textcolor}%
\pgfsetfillcolor{textcolor}%
\pgftext[x=0.289968in, y=0.772628in, left, base]{\color{textcolor}\sffamily\fontsize{10.000000}{12.000000}\selectfont \(\displaystyle {0.02}\)}%
\end{pgfscope}%
\begin{pgfscope}%
\pgfpathrectangle{\pgfqpoint{0.634105in}{0.521603in}}{\pgfqpoint{2.092048in}{2.278397in}}%
\pgfusepath{clip}%
\pgfsetrectcap%
\pgfsetroundjoin%
\pgfsetlinewidth{0.803000pt}%
\definecolor{currentstroke}{rgb}{0.690196,0.690196,0.690196}%
\pgfsetstrokecolor{currentstroke}%
\pgfsetdash{}{0pt}%
\pgfpathmoveto{\pgfqpoint{0.634105in}{1.129176in}}%
\pgfpathlineto{\pgfqpoint{2.726152in}{1.129176in}}%
\pgfusepath{stroke}%
\end{pgfscope}%
\begin{pgfscope}%
\pgfsetbuttcap%
\pgfsetroundjoin%
\definecolor{currentfill}{rgb}{0.000000,0.000000,0.000000}%
\pgfsetfillcolor{currentfill}%
\pgfsetlinewidth{0.803000pt}%
\definecolor{currentstroke}{rgb}{0.000000,0.000000,0.000000}%
\pgfsetstrokecolor{currentstroke}%
\pgfsetdash{}{0pt}%
\pgfsys@defobject{currentmarker}{\pgfqpoint{-0.048611in}{0.000000in}}{\pgfqpoint{-0.000000in}{0.000000in}}{%
\pgfpathmoveto{\pgfqpoint{-0.000000in}{0.000000in}}%
\pgfpathlineto{\pgfqpoint{-0.048611in}{0.000000in}}%
\pgfusepath{stroke,fill}%
}%
\begin{pgfscope}%
\pgfsys@transformshift{0.634105in}{1.129176in}%
\pgfsys@useobject{currentmarker}{}%
\end{pgfscope}%
\end{pgfscope}%
\begin{pgfscope}%
\definecolor{textcolor}{rgb}{0.000000,0.000000,0.000000}%
\pgfsetstrokecolor{textcolor}%
\pgfsetfillcolor{textcolor}%
\pgftext[x=0.289968in, y=1.076414in, left, base]{\color{textcolor}\sffamily\fontsize{10.000000}{12.000000}\selectfont \(\displaystyle {0.04}\)}%
\end{pgfscope}%
\begin{pgfscope}%
\pgfpathrectangle{\pgfqpoint{0.634105in}{0.521603in}}{\pgfqpoint{2.092048in}{2.278397in}}%
\pgfusepath{clip}%
\pgfsetrectcap%
\pgfsetroundjoin%
\pgfsetlinewidth{0.803000pt}%
\definecolor{currentstroke}{rgb}{0.690196,0.690196,0.690196}%
\pgfsetstrokecolor{currentstroke}%
\pgfsetdash{}{0pt}%
\pgfpathmoveto{\pgfqpoint{0.634105in}{1.432962in}}%
\pgfpathlineto{\pgfqpoint{2.726152in}{1.432962in}}%
\pgfusepath{stroke}%
\end{pgfscope}%
\begin{pgfscope}%
\pgfsetbuttcap%
\pgfsetroundjoin%
\definecolor{currentfill}{rgb}{0.000000,0.000000,0.000000}%
\pgfsetfillcolor{currentfill}%
\pgfsetlinewidth{0.803000pt}%
\definecolor{currentstroke}{rgb}{0.000000,0.000000,0.000000}%
\pgfsetstrokecolor{currentstroke}%
\pgfsetdash{}{0pt}%
\pgfsys@defobject{currentmarker}{\pgfqpoint{-0.048611in}{0.000000in}}{\pgfqpoint{-0.000000in}{0.000000in}}{%
\pgfpathmoveto{\pgfqpoint{-0.000000in}{0.000000in}}%
\pgfpathlineto{\pgfqpoint{-0.048611in}{0.000000in}}%
\pgfusepath{stroke,fill}%
}%
\begin{pgfscope}%
\pgfsys@transformshift{0.634105in}{1.432962in}%
\pgfsys@useobject{currentmarker}{}%
\end{pgfscope}%
\end{pgfscope}%
\begin{pgfscope}%
\definecolor{textcolor}{rgb}{0.000000,0.000000,0.000000}%
\pgfsetstrokecolor{textcolor}%
\pgfsetfillcolor{textcolor}%
\pgftext[x=0.289968in, y=1.380200in, left, base]{\color{textcolor}\sffamily\fontsize{10.000000}{12.000000}\selectfont \(\displaystyle {0.06}\)}%
\end{pgfscope}%
\begin{pgfscope}%
\pgfpathrectangle{\pgfqpoint{0.634105in}{0.521603in}}{\pgfqpoint{2.092048in}{2.278397in}}%
\pgfusepath{clip}%
\pgfsetrectcap%
\pgfsetroundjoin%
\pgfsetlinewidth{0.803000pt}%
\definecolor{currentstroke}{rgb}{0.690196,0.690196,0.690196}%
\pgfsetstrokecolor{currentstroke}%
\pgfsetdash{}{0pt}%
\pgfpathmoveto{\pgfqpoint{0.634105in}{1.736748in}}%
\pgfpathlineto{\pgfqpoint{2.726152in}{1.736748in}}%
\pgfusepath{stroke}%
\end{pgfscope}%
\begin{pgfscope}%
\pgfsetbuttcap%
\pgfsetroundjoin%
\definecolor{currentfill}{rgb}{0.000000,0.000000,0.000000}%
\pgfsetfillcolor{currentfill}%
\pgfsetlinewidth{0.803000pt}%
\definecolor{currentstroke}{rgb}{0.000000,0.000000,0.000000}%
\pgfsetstrokecolor{currentstroke}%
\pgfsetdash{}{0pt}%
\pgfsys@defobject{currentmarker}{\pgfqpoint{-0.048611in}{0.000000in}}{\pgfqpoint{-0.000000in}{0.000000in}}{%
\pgfpathmoveto{\pgfqpoint{-0.000000in}{0.000000in}}%
\pgfpathlineto{\pgfqpoint{-0.048611in}{0.000000in}}%
\pgfusepath{stroke,fill}%
}%
\begin{pgfscope}%
\pgfsys@transformshift{0.634105in}{1.736748in}%
\pgfsys@useobject{currentmarker}{}%
\end{pgfscope}%
\end{pgfscope}%
\begin{pgfscope}%
\definecolor{textcolor}{rgb}{0.000000,0.000000,0.000000}%
\pgfsetstrokecolor{textcolor}%
\pgfsetfillcolor{textcolor}%
\pgftext[x=0.289968in, y=1.683987in, left, base]{\color{textcolor}\sffamily\fontsize{10.000000}{12.000000}\selectfont \(\displaystyle {0.08}\)}%
\end{pgfscope}%
\begin{pgfscope}%
\pgfpathrectangle{\pgfqpoint{0.634105in}{0.521603in}}{\pgfqpoint{2.092048in}{2.278397in}}%
\pgfusepath{clip}%
\pgfsetrectcap%
\pgfsetroundjoin%
\pgfsetlinewidth{0.803000pt}%
\definecolor{currentstroke}{rgb}{0.690196,0.690196,0.690196}%
\pgfsetstrokecolor{currentstroke}%
\pgfsetdash{}{0pt}%
\pgfpathmoveto{\pgfqpoint{0.634105in}{2.040534in}}%
\pgfpathlineto{\pgfqpoint{2.726152in}{2.040534in}}%
\pgfusepath{stroke}%
\end{pgfscope}%
\begin{pgfscope}%
\pgfsetbuttcap%
\pgfsetroundjoin%
\definecolor{currentfill}{rgb}{0.000000,0.000000,0.000000}%
\pgfsetfillcolor{currentfill}%
\pgfsetlinewidth{0.803000pt}%
\definecolor{currentstroke}{rgb}{0.000000,0.000000,0.000000}%
\pgfsetstrokecolor{currentstroke}%
\pgfsetdash{}{0pt}%
\pgfsys@defobject{currentmarker}{\pgfqpoint{-0.048611in}{0.000000in}}{\pgfqpoint{-0.000000in}{0.000000in}}{%
\pgfpathmoveto{\pgfqpoint{-0.000000in}{0.000000in}}%
\pgfpathlineto{\pgfqpoint{-0.048611in}{0.000000in}}%
\pgfusepath{stroke,fill}%
}%
\begin{pgfscope}%
\pgfsys@transformshift{0.634105in}{2.040534in}%
\pgfsys@useobject{currentmarker}{}%
\end{pgfscope}%
\end{pgfscope}%
\begin{pgfscope}%
\definecolor{textcolor}{rgb}{0.000000,0.000000,0.000000}%
\pgfsetstrokecolor{textcolor}%
\pgfsetfillcolor{textcolor}%
\pgftext[x=0.289968in, y=1.987773in, left, base]{\color{textcolor}\sffamily\fontsize{10.000000}{12.000000}\selectfont \(\displaystyle {0.10}\)}%
\end{pgfscope}%
\begin{pgfscope}%
\pgfpathrectangle{\pgfqpoint{0.634105in}{0.521603in}}{\pgfqpoint{2.092048in}{2.278397in}}%
\pgfusepath{clip}%
\pgfsetrectcap%
\pgfsetroundjoin%
\pgfsetlinewidth{0.803000pt}%
\definecolor{currentstroke}{rgb}{0.690196,0.690196,0.690196}%
\pgfsetstrokecolor{currentstroke}%
\pgfsetdash{}{0pt}%
\pgfpathmoveto{\pgfqpoint{0.634105in}{2.344321in}}%
\pgfpathlineto{\pgfqpoint{2.726152in}{2.344321in}}%
\pgfusepath{stroke}%
\end{pgfscope}%
\begin{pgfscope}%
\pgfsetbuttcap%
\pgfsetroundjoin%
\definecolor{currentfill}{rgb}{0.000000,0.000000,0.000000}%
\pgfsetfillcolor{currentfill}%
\pgfsetlinewidth{0.803000pt}%
\definecolor{currentstroke}{rgb}{0.000000,0.000000,0.000000}%
\pgfsetstrokecolor{currentstroke}%
\pgfsetdash{}{0pt}%
\pgfsys@defobject{currentmarker}{\pgfqpoint{-0.048611in}{0.000000in}}{\pgfqpoint{-0.000000in}{0.000000in}}{%
\pgfpathmoveto{\pgfqpoint{-0.000000in}{0.000000in}}%
\pgfpathlineto{\pgfqpoint{-0.048611in}{0.000000in}}%
\pgfusepath{stroke,fill}%
}%
\begin{pgfscope}%
\pgfsys@transformshift{0.634105in}{2.344321in}%
\pgfsys@useobject{currentmarker}{}%
\end{pgfscope}%
\end{pgfscope}%
\begin{pgfscope}%
\definecolor{textcolor}{rgb}{0.000000,0.000000,0.000000}%
\pgfsetstrokecolor{textcolor}%
\pgfsetfillcolor{textcolor}%
\pgftext[x=0.289968in, y=2.291559in, left, base]{\color{textcolor}\sffamily\fontsize{10.000000}{12.000000}\selectfont \(\displaystyle {0.12}\)}%
\end{pgfscope}%
\begin{pgfscope}%
\pgfpathrectangle{\pgfqpoint{0.634105in}{0.521603in}}{\pgfqpoint{2.092048in}{2.278397in}}%
\pgfusepath{clip}%
\pgfsetrectcap%
\pgfsetroundjoin%
\pgfsetlinewidth{0.803000pt}%
\definecolor{currentstroke}{rgb}{0.690196,0.690196,0.690196}%
\pgfsetstrokecolor{currentstroke}%
\pgfsetdash{}{0pt}%
\pgfpathmoveto{\pgfqpoint{0.634105in}{2.648107in}}%
\pgfpathlineto{\pgfqpoint{2.726152in}{2.648107in}}%
\pgfusepath{stroke}%
\end{pgfscope}%
\begin{pgfscope}%
\pgfsetbuttcap%
\pgfsetroundjoin%
\definecolor{currentfill}{rgb}{0.000000,0.000000,0.000000}%
\pgfsetfillcolor{currentfill}%
\pgfsetlinewidth{0.803000pt}%
\definecolor{currentstroke}{rgb}{0.000000,0.000000,0.000000}%
\pgfsetstrokecolor{currentstroke}%
\pgfsetdash{}{0pt}%
\pgfsys@defobject{currentmarker}{\pgfqpoint{-0.048611in}{0.000000in}}{\pgfqpoint{-0.000000in}{0.000000in}}{%
\pgfpathmoveto{\pgfqpoint{-0.000000in}{0.000000in}}%
\pgfpathlineto{\pgfqpoint{-0.048611in}{0.000000in}}%
\pgfusepath{stroke,fill}%
}%
\begin{pgfscope}%
\pgfsys@transformshift{0.634105in}{2.648107in}%
\pgfsys@useobject{currentmarker}{}%
\end{pgfscope}%
\end{pgfscope}%
\begin{pgfscope}%
\definecolor{textcolor}{rgb}{0.000000,0.000000,0.000000}%
\pgfsetstrokecolor{textcolor}%
\pgfsetfillcolor{textcolor}%
\pgftext[x=0.289968in, y=2.595345in, left, base]{\color{textcolor}\sffamily\fontsize{10.000000}{12.000000}\selectfont \(\displaystyle {0.14}\)}%
\end{pgfscope}%
\begin{pgfscope}%
\definecolor{textcolor}{rgb}{0.000000,0.000000,0.000000}%
\pgfsetstrokecolor{textcolor}%
\pgfsetfillcolor{textcolor}%
\pgftext[x=0.234413in,y=1.660802in,,bottom,rotate=90.000000]{\color{textcolor}\sffamily\fontsize{10.000000}{12.000000}\selectfont MAE}%
\end{pgfscope}%
\begin{pgfscope}%
\pgfpathrectangle{\pgfqpoint{0.634105in}{0.521603in}}{\pgfqpoint{2.092048in}{2.278397in}}%
\pgfusepath{clip}%
\pgfsetrectcap%
\pgfsetroundjoin%
\pgfsetlinewidth{1.505625pt}%
\definecolor{currentstroke}{rgb}{1.000000,0.000000,0.000000}%
\pgfsetstrokecolor{currentstroke}%
\pgfsetdash{}{0pt}%
\pgfpathmoveto{\pgfqpoint{0.729198in}{1.675991in}}%
\pgfpathlineto{\pgfqpoint{1.689734in}{1.736748in}}%
\pgfpathlineto{\pgfqpoint{1.881841in}{1.843073in}}%
\pgfpathlineto{\pgfqpoint{2.073948in}{1.645612in}}%
\pgfpathlineto{\pgfqpoint{2.266056in}{1.554476in}}%
\pgfpathlineto{\pgfqpoint{2.458163in}{1.615234in}}%
\pgfpathlineto{\pgfqpoint{2.631059in}{1.554476in}}%
\pgfusepath{stroke}%
\end{pgfscope}%
\begin{pgfscope}%
\pgfpathrectangle{\pgfqpoint{0.634105in}{0.521603in}}{\pgfqpoint{2.092048in}{2.278397in}}%
\pgfusepath{clip}%
\pgfsetbuttcap%
\pgfsetroundjoin%
\definecolor{currentfill}{rgb}{1.000000,0.000000,0.000000}%
\pgfsetfillcolor{currentfill}%
\pgfsetlinewidth{1.003750pt}%
\definecolor{currentstroke}{rgb}{1.000000,0.000000,0.000000}%
\pgfsetstrokecolor{currentstroke}%
\pgfsetdash{}{0pt}%
\pgfsys@defobject{currentmarker}{\pgfqpoint{-0.041667in}{-0.041667in}}{\pgfqpoint{0.041667in}{0.041667in}}{%
\pgfpathmoveto{\pgfqpoint{0.000000in}{-0.041667in}}%
\pgfpathcurveto{\pgfqpoint{0.011050in}{-0.041667in}}{\pgfqpoint{0.021649in}{-0.037276in}}{\pgfqpoint{0.029463in}{-0.029463in}}%
\pgfpathcurveto{\pgfqpoint{0.037276in}{-0.021649in}}{\pgfqpoint{0.041667in}{-0.011050in}}{\pgfqpoint{0.041667in}{0.000000in}}%
\pgfpathcurveto{\pgfqpoint{0.041667in}{0.011050in}}{\pgfqpoint{0.037276in}{0.021649in}}{\pgfqpoint{0.029463in}{0.029463in}}%
\pgfpathcurveto{\pgfqpoint{0.021649in}{0.037276in}}{\pgfqpoint{0.011050in}{0.041667in}}{\pgfqpoint{0.000000in}{0.041667in}}%
\pgfpathcurveto{\pgfqpoint{-0.011050in}{0.041667in}}{\pgfqpoint{-0.021649in}{0.037276in}}{\pgfqpoint{-0.029463in}{0.029463in}}%
\pgfpathcurveto{\pgfqpoint{-0.037276in}{0.021649in}}{\pgfqpoint{-0.041667in}{0.011050in}}{\pgfqpoint{-0.041667in}{0.000000in}}%
\pgfpathcurveto{\pgfqpoint{-0.041667in}{-0.011050in}}{\pgfqpoint{-0.037276in}{-0.021649in}}{\pgfqpoint{-0.029463in}{-0.029463in}}%
\pgfpathcurveto{\pgfqpoint{-0.021649in}{-0.037276in}}{\pgfqpoint{-0.011050in}{-0.041667in}}{\pgfqpoint{0.000000in}{-0.041667in}}%
\pgfpathlineto{\pgfqpoint{0.000000in}{-0.041667in}}%
\pgfpathclose%
\pgfusepath{stroke,fill}%
}%
\begin{pgfscope}%
\pgfsys@transformshift{0.729198in}{1.675991in}%
\pgfsys@useobject{currentmarker}{}%
\end{pgfscope}%
\begin{pgfscope}%
\pgfsys@transformshift{1.689734in}{1.736748in}%
\pgfsys@useobject{currentmarker}{}%
\end{pgfscope}%
\begin{pgfscope}%
\pgfsys@transformshift{1.881841in}{1.843073in}%
\pgfsys@useobject{currentmarker}{}%
\end{pgfscope}%
\begin{pgfscope}%
\pgfsys@transformshift{2.073948in}{1.645612in}%
\pgfsys@useobject{currentmarker}{}%
\end{pgfscope}%
\begin{pgfscope}%
\pgfsys@transformshift{2.266056in}{1.554476in}%
\pgfsys@useobject{currentmarker}{}%
\end{pgfscope}%
\begin{pgfscope}%
\pgfsys@transformshift{2.458163in}{1.615234in}%
\pgfsys@useobject{currentmarker}{}%
\end{pgfscope}%
\begin{pgfscope}%
\pgfsys@transformshift{2.631059in}{1.554476in}%
\pgfsys@useobject{currentmarker}{}%
\end{pgfscope}%
\end{pgfscope}%
\begin{pgfscope}%
\pgfsetrectcap%
\pgfsetmiterjoin%
\pgfsetlinewidth{0.803000pt}%
\definecolor{currentstroke}{rgb}{0.000000,0.000000,0.000000}%
\pgfsetstrokecolor{currentstroke}%
\pgfsetdash{}{0pt}%
\pgfpathmoveto{\pgfqpoint{0.634105in}{0.521603in}}%
\pgfpathlineto{\pgfqpoint{0.634105in}{2.800000in}}%
\pgfusepath{stroke}%
\end{pgfscope}%
\begin{pgfscope}%
\pgfsetrectcap%
\pgfsetmiterjoin%
\pgfsetlinewidth{0.803000pt}%
\definecolor{currentstroke}{rgb}{0.000000,0.000000,0.000000}%
\pgfsetstrokecolor{currentstroke}%
\pgfsetdash{}{0pt}%
\pgfpathmoveto{\pgfqpoint{2.726152in}{0.521603in}}%
\pgfpathlineto{\pgfqpoint{2.726152in}{2.800000in}}%
\pgfusepath{stroke}%
\end{pgfscope}%
\begin{pgfscope}%
\pgfsetrectcap%
\pgfsetmiterjoin%
\pgfsetlinewidth{0.803000pt}%
\definecolor{currentstroke}{rgb}{0.000000,0.000000,0.000000}%
\pgfsetstrokecolor{currentstroke}%
\pgfsetdash{}{0pt}%
\pgfpathmoveto{\pgfqpoint{0.634105in}{0.521603in}}%
\pgfpathlineto{\pgfqpoint{2.726152in}{0.521603in}}%
\pgfusepath{stroke}%
\end{pgfscope}%
\begin{pgfscope}%
\pgfsetrectcap%
\pgfsetmiterjoin%
\pgfsetlinewidth{0.803000pt}%
\definecolor{currentstroke}{rgb}{0.000000,0.000000,0.000000}%
\pgfsetstrokecolor{currentstroke}%
\pgfsetdash{}{0pt}%
\pgfpathmoveto{\pgfqpoint{0.634105in}{2.800000in}}%
\pgfpathlineto{\pgfqpoint{2.726152in}{2.800000in}}%
\pgfusepath{stroke}%
\end{pgfscope}%
\end{pgfpicture}%
\makeatother%
\endgroup%